\documentclass[10pt,twocolumn,letterpaper]{article}
\usepackage{acronym}
\PassOptionsToPackage{table,xcdraw}{xcolor}
\PassOptionsToPackage{nameinlink}{cleveref}

\usepackage[pagenumbers]{cvpr}
\usepackage{tabularx}
\usepackage{makecell}
\usepackage{multirow}
\usepackage{array}
\usepackage{threeparttable}
\usepackage{adjustbox}
\usepackage{pgfplots}
\usepgfplotslibrary{groupplots, fillbetween, colormaps}
\pgfplotsset{compat=1.18}
\usepackage{tikz}
\usetikzlibrary{positioning,arrows.meta,shapes.geometric,fit,backgrounds,calc,patterns}
\usepackage{soul}

\definecolor{cvprblue}{rgb}{0.21,0.49,0.74}
\usepackage[pagebackref,breaklinks,colorlinks,allcolors=cvprblue]{hyperref}

\AtEndPreamble{%
  \Crefname{figure}{Fig.}{Figs.}
  \crefname{figure}{Fig.}{Figs.}
  \Crefname{table}{Table}{Tables}
  \crefname{table}{Table}{Tables}
  \Crefname{section}{Sec.}{Secs.}
  \crefname{section}{Sec.}{Secs.}
}

\makeatletter
\newcommand{\traversal}[1]{\texttt{\trav@split#1-\trav@end-}}
\def\trav@split#1-#2\trav@end-{%
  #1\trav@more#2\trav@end-}
\def\trav@more#1\trav@end-{%
  \ifx\relax#1\relax\else\discretionary{-}{}{-}\trav@split#1\trav@end-\fi}
\makeatother

\newacro{R1}[R@1]{Recall@1}
\newacro{RK}[R@$K$]{Recall@$K$}
\newacro{DAR}{Distractor-Augmented Recall}
\newacro{DARx}{DAR excluded}
\newacro{StLuciaMToD}{St.~Lucia Multiple Times of Day}

\newacro{pp}{percentage points}

\newacro{NLP}{natural language processing}
\newacro{STD}{Standardization of Descriptors}
\newacro{VPR}{Visual Place Recognition}

\definecolor{bestcolor}{RGB}{130,190,130}
\definecolor{secondcolor}{RGB}{200,255,200}
\definecolor{worstcolor}{RGB}{190,130,130}

\def\confName{CVPR}
\def\confYear{2027}

\title{
What Should Visual Place Recognition Be Invariant To?\\
Rethinking Condition Invariance in Visual Place Recognition
}
 
\hypersetup{
  pdftitle={What Should Visual Place Recognition Be Invariant To? Rethinking Condition Invariance in Visual Place Recognition},
  pdfauthor={Beomsu Kim, Minwoo Jung, and Giseop Kim}
}

\author{Beomsu Kim, Minwoo Jung, and Giseop Kim$^{*}$%
\thanks{$^{*}$Corresponding author.}%
\thanks{B. Kim and G. Kim are with the Department of Robotics and
Mechatronics Engineering, DGIST, Daegu, Republic of Korea,
{\tt\small \{beomsu.kim, gsk\}@dgist.ac.kr}.}%
\thanks{M. Jung is with the Department of Mechanical Engineering,
Seoul National University, Seoul, Republic of Korea,
{\tt\small moonshot@snu.ac.kr}.}%
}

\begin{document}
\begingroup
\makeatletter
\renewcommand{\@fnsymbol}[1]{}
\makeatother
\maketitle
\endgroup

\begin{abstract}
    
Condition invariance in \ac{VPR} is typically evaluated by matching queries
from one condition against a database from another condition.
This tests the ability to recognize the same place
when the query's condition differs from the database's condition.
However, to map large environments, a database can be built
from images collected by multiple robots or across multiple
traversals under different conditions. In this setting,
condition blindness---the ability to remain invariant to condition differences
among database images---becomes important, but standard cross-condition
evaluation does not provide a controlled assessment of this ability.
Therefore, we introduce \ac{DAR} to
measure condition blindness in a controlled manner.
Across eleven methods and six datasets, we find that
differences in condition blindness can be larger
than Recall@1 alone would suggest. To improve
the condition blindness that \ac{DAR} reveals, we compare
three methods that make \ac{VPR} descriptors more invariant to
condition changes through post-processing. Among the three methods,
we demonstrate INLP on a database built from multiple
traversals under different conditions, showing that it can
improve the performance of existing \ac{VPR} methods.

\end{abstract}
\section{Introduction}
\begin{figure}[t!]
\centering
\begin{tikzpicture}[
    font=\scriptsize\sffamily,
    node distance=3mm,
    img/.style={
        draw, line width=0.3pt,
        inner sep=0.15pt, outer sep=1.05pt,
        minimum width=22.5mm, minimum height=16.875mm
    },
    match/.style={
        draw=green!55!black, line width=2.4pt,
        inner sep=1.2pt, outer sep=0pt,
        minimum width=22.5mm, minimum height=16.875mm
    },
    wrongmatch/.style={
        draw=red!70!black, line width=2.4pt,
        inner sep=1.2pt, outer sep=0pt,
        minimum width=22.5mm, minimum height=16.875mm
    },
    nodist/.style={
        draw=gray, dashed, line width=0.6pt,
        pattern=north east lines, pattern color=gray!45,
        inner sep=0.15pt, outer sep=1.05pt,
        minimum width=22.5mm, minimum height=16.875mm,
        align=center, text=gray
    },
    lbl/.style={font=\footnotesize\sffamily, inner sep=1pt},
    caselbl/.style={font=\footnotesize\sffamily\bfseries, inner sep=1pt},
    score/.style={font=\scriptsize\sffamily, inner sep=0.5pt},
    tagday/.style={font=\scriptsize\sffamily, inner sep=0.5pt,
        fill=yellow!60!orange, text=black},
    tagnight/.style={font=\scriptsize\sffamily, inner sep=0.5pt,
        fill=blue!40!black, text=white},
    arr/.style={-latex, line width=0.4pt},
    arrmatch/.style={-{Latex[length=3mm,width=3mm]}, line width=2.4pt, draw=green!55!black},
    arrwrong/.style={-{Latex[length=3mm,width=3mm]}, line width=2.4pt, draw=red!70!black}
]

\node[img] (queryA)
    {\includegraphics[width=22.5mm,height=16.875mm]{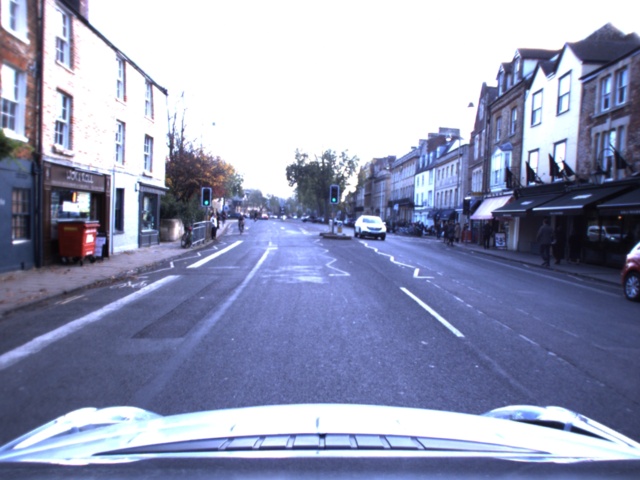}};
\node[tagday, below=0.3mm of queryA] {Day};

\node[match, right=1.5mm of queryA] (dbA1)
    {\includegraphics[width=22.5mm,height=16.875mm]{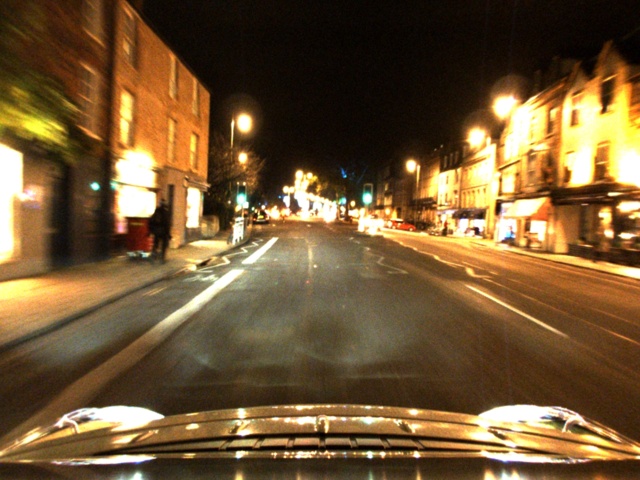}};
\node[img, right=1.5mm of dbA1] (dbA2)
    {\includegraphics[width=22.5mm,height=16.875mm]{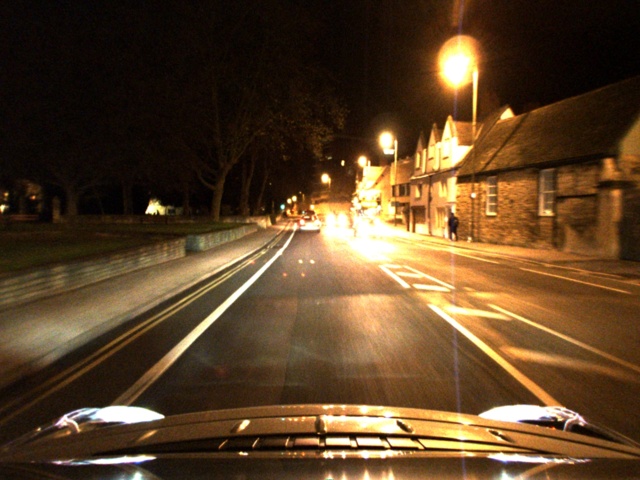}};
\node[img, below=5mm of queryA] (dbA3)
    {\includegraphics[width=22.5mm,height=16.875mm]{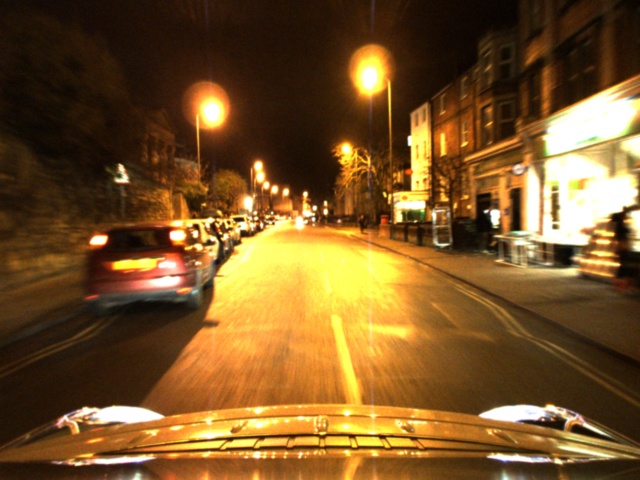}};
\node[img, below=5mm of dbA1] (dbA4)
    {\includegraphics[width=22.5mm,height=16.875mm]{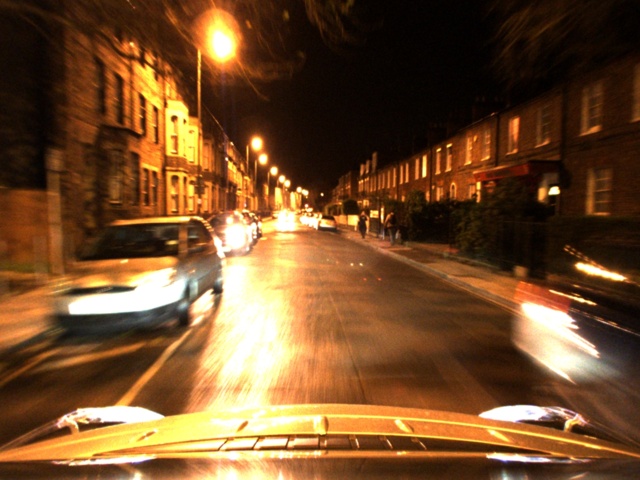}};
\node[img, below=5mm of dbA2] (dbA5)
    {\includegraphics[width=22.5mm,height=16.875mm]{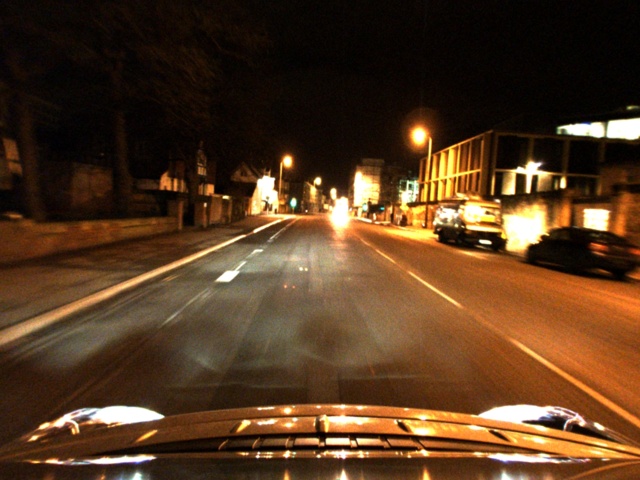}};

\node[anchor=north, font=\scriptsize\sffamily, inner sep=0.5pt]
    at ($(dbA1.south)+(0,-0.4mm)$)
    {\fboxsep=0.5pt\colorbox{blue!40!black}{\textcolor{white}{Night}}\,\textcolor{green!55!black}{\textbf{sim=0.59\,\checkmark}}};

\node[anchor=north, font=\scriptsize\sffamily, inner sep=0.5pt]
    at ($(dbA2.south)+(0,-0.4mm)$)
    {\fboxsep=0.5pt\colorbox{blue!40!black}{\textcolor{white}{Night}}\,\textcolor{red!70!black}{sim=0.31\,$\boldsymbol\times$}};

\node[anchor=north, font=\scriptsize\sffamily, inner sep=0.5pt]
    at ($(dbA3.south)+(0,-0.4mm)$)
    {\fboxsep=0.5pt\colorbox{blue!40!black}{\textcolor{white}{Night}}\,\textcolor{red!70!black}{sim=0.50\,$\boldsymbol\times$}};

\node[anchor=north, font=\scriptsize\sffamily, inner sep=0.5pt]
    at ($(dbA4.south)+(0,-0.4mm)$)
    {\fboxsep=0.5pt\colorbox{blue!40!black}{\textcolor{white}{Night}}\,\textcolor{red!70!black}{sim=0.43\,$\boldsymbol\times$}};

\node[anchor=north, font=\scriptsize\sffamily, inner sep=0.5pt]
    at ($(dbA5.south)+(0,-0.4mm)$)
    {\fboxsep=0.5pt\colorbox{blue!40!black}{\textcolor{white}{Night}}\,\textcolor{red!70!black}{sim=0.30\,$\boldsymbol\times$}};

\begin{scope}[on background layer]
    \coordinate (Axl) at ($(dbA3.west)+(-1.5mm,0)$);
    \coordinate (Axr) at ($(dbA5.east)+(1.5mm,0)$);
    \coordinate (Ayt) at ($(dbA2.north)+(0,1.5mm)$);
    \coordinate (Ayb) at ($(dbA4.south)+(0,-4.5mm)$);
    \coordinate (Agreenedge) at ($(dbA1.west)+(-1.2pt,0)$);
    \coordinate (Axn) at ($(queryA.east)!0.5!(Agreenedge)$);
    \coordinate (Ayn) at ($(dbA4.north)+(0,1.5mm)$);
    \draw[draw, dashed, line width=0.8pt, rounded corners=2pt]
        (Axn |- Ayt) -- (Axr |- Ayt) -- (Axr |- Ayb) --
        (Axl |- Ayb) -- (Axl |- Ayn) -- (Axn |- Ayn) -- cycle;
    \node[lbl, anchor=south] at ($(Axn |- Ayt)!0.5!(Axr |- Ayt)+(0,1mm)$)
        {Single-condition DB};
    \coordinate (Aboxtop) at (Axr |- Ayt);
\end{scope}
\node[lbl, anchor=south] at ($(queryA.north)+(0,1mm)$) {Query};

\coordinate (AarrS) at ($(queryA.north west)!0.75!(queryA.north east)$);
\coordinate (AarrE) at ($(dbA1.north west)!0.25!(dbA1.north east)$);
\draw[arrmatch] (AarrS) to[out=90,in=135,looseness=0.75] (AarrE);

\node[caselbl, text=green!55!black, anchor=west]
    at ($(queryA.west |- Aboxtop)+(-0.5,7mm)$)
    {(a) Place recognition across conditions};  
\node[img, below=40mm of queryA] (queryB)
    {\includegraphics[width=22.5mm,height=16.875mm]{figures/distractor_illustration/query.jpg}};
\node[tagday, below=0.3mm of queryB] {Day};

\node[img, right=1.5mm of queryB] (dbB1)
    {\includegraphics[width=22.5mm,height=16.875mm]{figures/distractor_illustration/correct.jpg}};
\node[img, right=1.5mm of dbB1] (dbB2)
    {\includegraphics[width=22.5mm,height=16.875mm]{figures/distractor_illustration/wrong1.jpg}};
\node[wrongmatch, below=5mm of queryB] (dbB3)
    {\includegraphics[width=22.5mm,height=16.875mm]{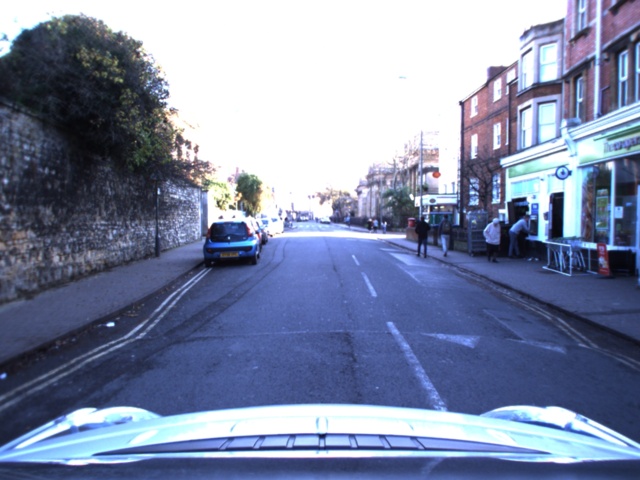}};
\node[img, below=5mm of dbB1] (dbB4)
    {\includegraphics[width=22.5mm,height=16.875mm]{figures/distractor_illustration/wrong2.jpg}};
\node[img, below=5mm of dbB2] (dbB5)
    {\includegraphics[width=22.5mm,height=16.875mm]{figures/distractor_illustration/wrong3.jpg}};

\node[anchor=north, font=\scriptsize\sffamily, inner sep=0.5pt]
    at ($(dbB1.south)+(0,-0.4mm)$)
    {\fboxsep=0.5pt\colorbox{blue!40!black}{\textcolor{white}{Night}}\,\textcolor{green!55!black}{sim=0.59\,\checkmark}};

\node[anchor=north, font=\scriptsize\sffamily, inner sep=0.5pt]
    at ($(dbB2.south)+(0,-0.4mm)$)
    {\fboxsep=0.5pt\colorbox{blue!40!black}{\textcolor{white}{Night}}\,\textcolor{red!70!black}{sim=0.31\,$\boldsymbol\times$}};

\node[anchor=north, font=\scriptsize\sffamily, inner sep=0.5pt]
    at ($(dbB3.south)+(0,-0.4mm)$)
    {\fboxsep=0.5pt\colorbox{yellow!60!orange}{Day}\,\textcolor{red!70!black}{\textbf{sim=0.62\,$\boldsymbol\times$}}};

\node[anchor=north, font=\scriptsize\sffamily, inner sep=0.5pt]
    at ($(dbB4.south)+(0,-0.4mm)$)
    {\fboxsep=0.5pt\colorbox{blue!40!black}{\textcolor{white}{Night}}\,\textcolor{red!70!black}{sim=0.43\,$\boldsymbol\times$}};

\node[anchor=north, font=\scriptsize\sffamily, inner sep=0.5pt]
    at ($(dbB5.south)+(0,-0.4mm)$)
    {\fboxsep=0.5pt\colorbox{blue!40!black}{\textcolor{white}{Night}}\,\textcolor{red!70!black}{sim=0.30\,$\boldsymbol\times$}};

\begin{scope}[on background layer]
    \coordinate (Bxl) at ($(dbB3.west)+(-1.5mm,0)$);
    \coordinate (Bxr) at ($(dbB5.east)+(1.5mm,0)$);
    \coordinate (Byt) at ($(dbB2.north)+(0,1.5mm)$);
    \coordinate (Byb) at ($(dbB4.south)+(0,-4.5mm)$);
    \coordinate (Bxn) at ($(dbB4.west)+(-0.75mm,0)$);
    \coordinate (Byn) at ($(dbB4.north)+(0,1.5mm)$);
    \draw[draw, dashed, line width=0.8pt, rounded corners=2pt]
        (Bxn |- Byt) -- (Bxr |- Byt) -- (Bxr |- Byb) --
        (Bxl |- Byb) -- (Bxl |- Byn) -- (Bxn |- Byn) -- cycle;
    \node[lbl, anchor=south] at ($(Bxn |- Byt)!0.5!(Bxr |- Byt)+(0,1mm)$)
        {Mixed-condition DB};
    \coordinate (Bboxtop) at (Bxr |- Byt);
\end{scope}
\node[lbl, anchor=south] at ($(queryB.north)+(0,1mm)$) {Query};

\coordinate (BarrS) at ($(queryB.north west)!0.75!(queryB.south west)$);
\coordinate (BarrE) at ($(dbB3.north west)!0.25!(dbB3.south west)$);
\draw[arrwrong] (BarrS) to[out=180,in=135,looseness=0.75] (BarrE);

\node[caselbl, text=red!70!black, anchor=west]
    at ($(queryB.west |- Bboxtop)+(-0.5,6mm)$)
    {(b) Failure in mixed-condition databases};

\end{tikzpicture}
\caption{\textbf{(a)} The method can recognize the same place under different conditions.
\textbf{(b)} The method fails to ignore the condition and is distracted by an image from a condition similar to the query's.}
\label{fig:distractor}

\vspace{-5mm}
\end{figure}
\newcommand{\rcornerlabel}[2]{%
  \begin{tikzpicture}[inner sep=0pt, outer sep=0pt]
    \node[draw, line width=\fboxrule, minimum width=\linewidth,
          minimum height=\linewidth, inner sep=0pt] (img) {%
      \adjincludegraphics[width=\dimexpr 0.95\linewidth-2\fboxrule\relax,
                          height=\dimexpr 0.95\linewidth-2\fboxrule\relax,
                          keepaspectratio]{#1}};
    \node[anchor=north east, fill=white, fill opacity=0.85, text opacity=1,
          inner sep=1.5pt, font=\scriptsize]
      at ([xshift=-2pt,yshift=-2pt]img.north east) {R@1: #2\%};
  \end{tikzpicture}%
}
\begin{figure}[t!]
\centering
\setlength{\tabcolsep}{2pt}
\renewcommand{\arraystretch}{1.0}
\setlength{\fboxsep}{0pt}
\begin{tabular}{@{}m{0.5cm} m{0.435\columnwidth} m{0.435\columnwidth}@{}}
 & \centering\small Single-condition DB & \centering\small Mixed-condition DB \tabularnewline
\centering\rotatebox{90}{\small AnyLoc} &
\rcornerlabel{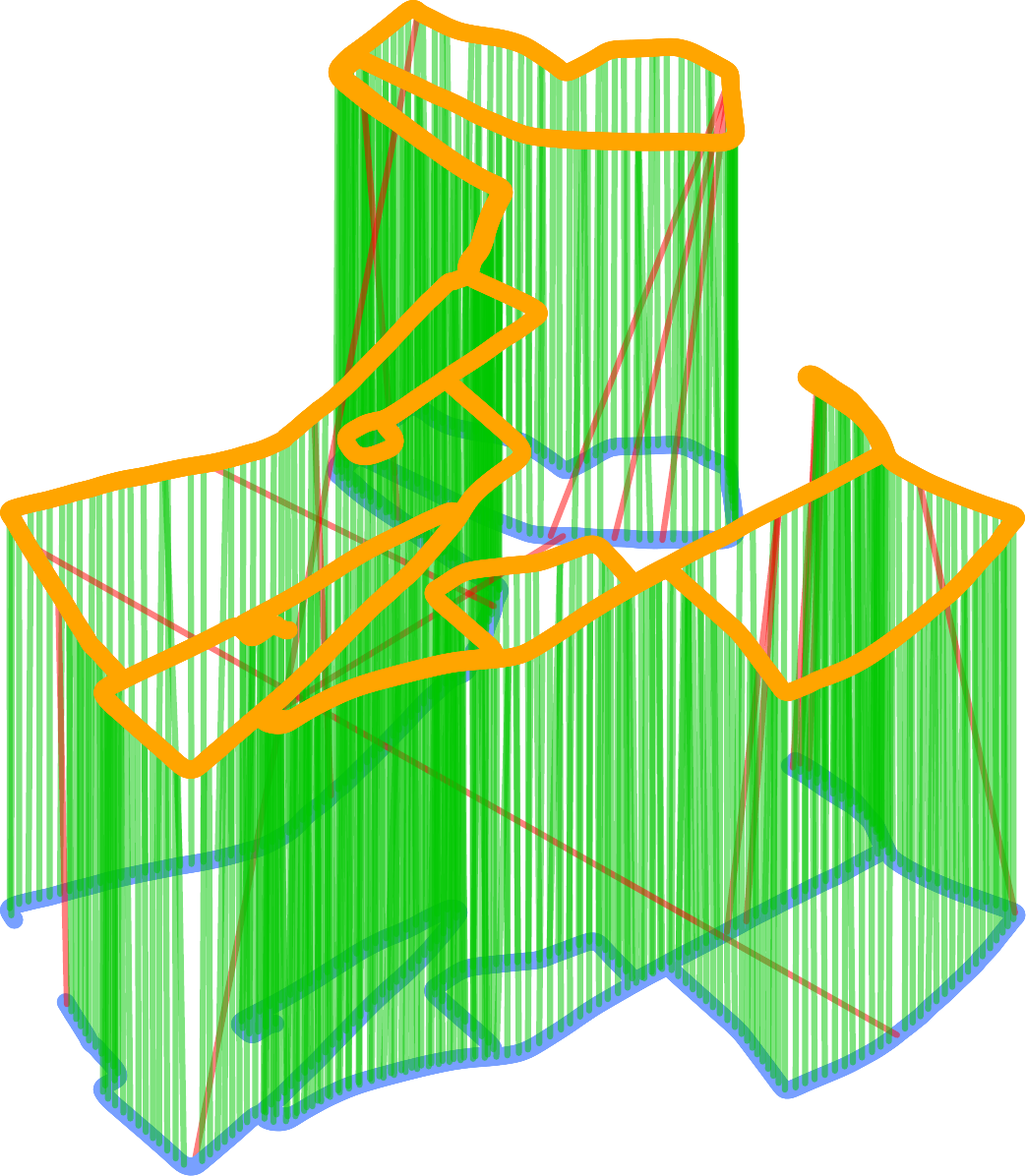}{96.1} &
\rcornerlabel{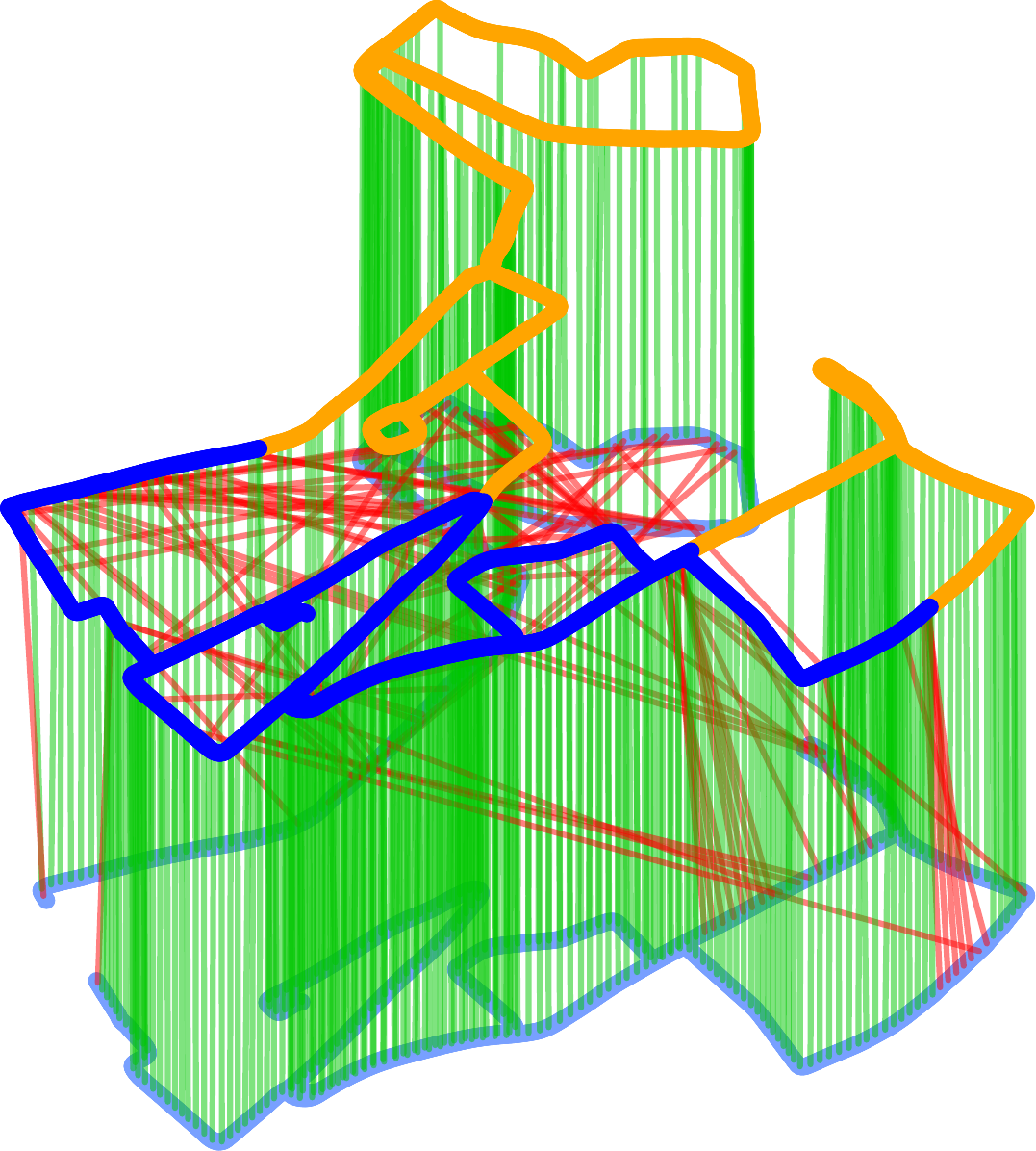}{80.4} \tabularnewline
\centering\rotatebox{90}{\small EigenPlaces} &
\rcornerlabel{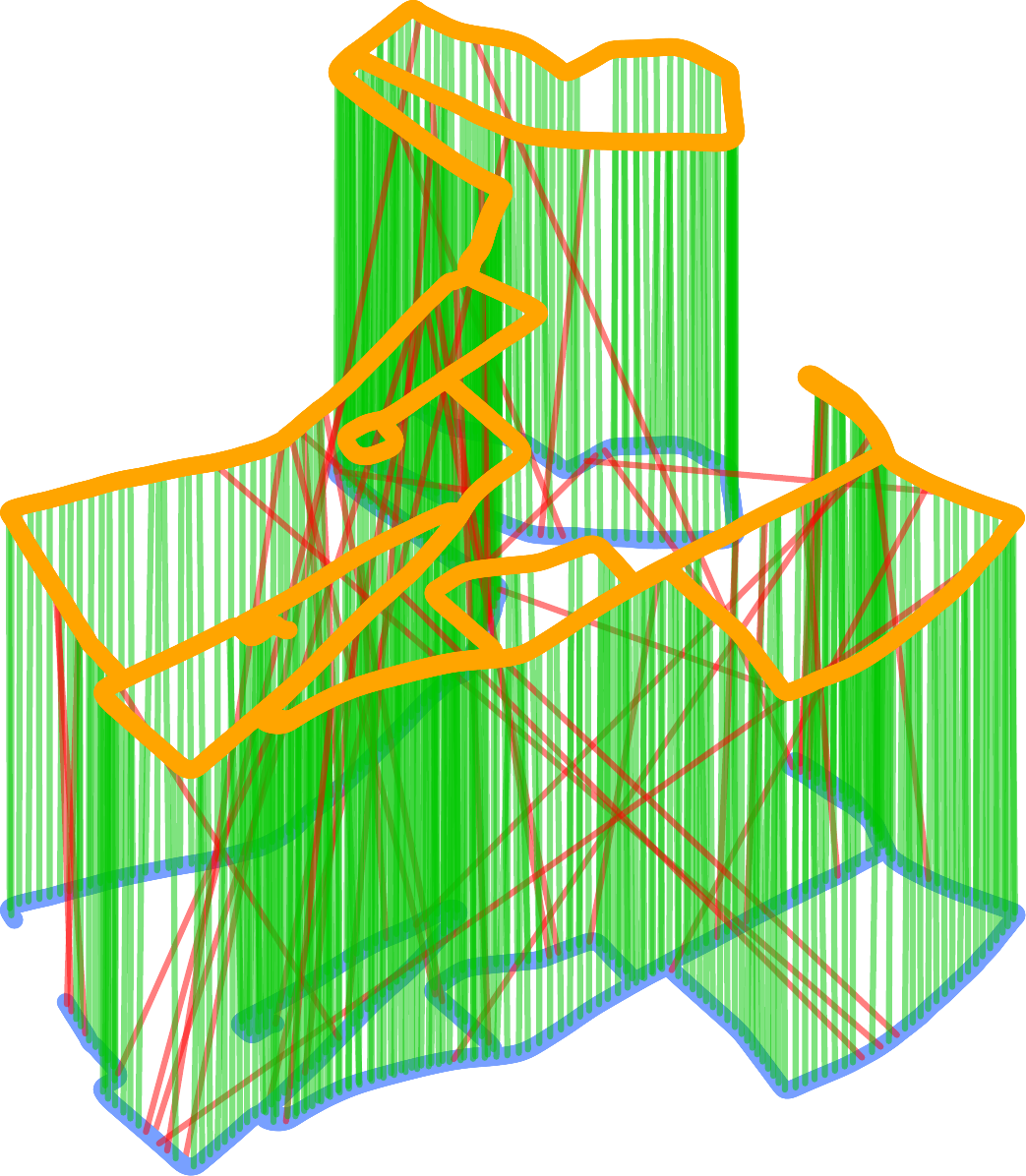}{91.7} &
\rcornerlabel{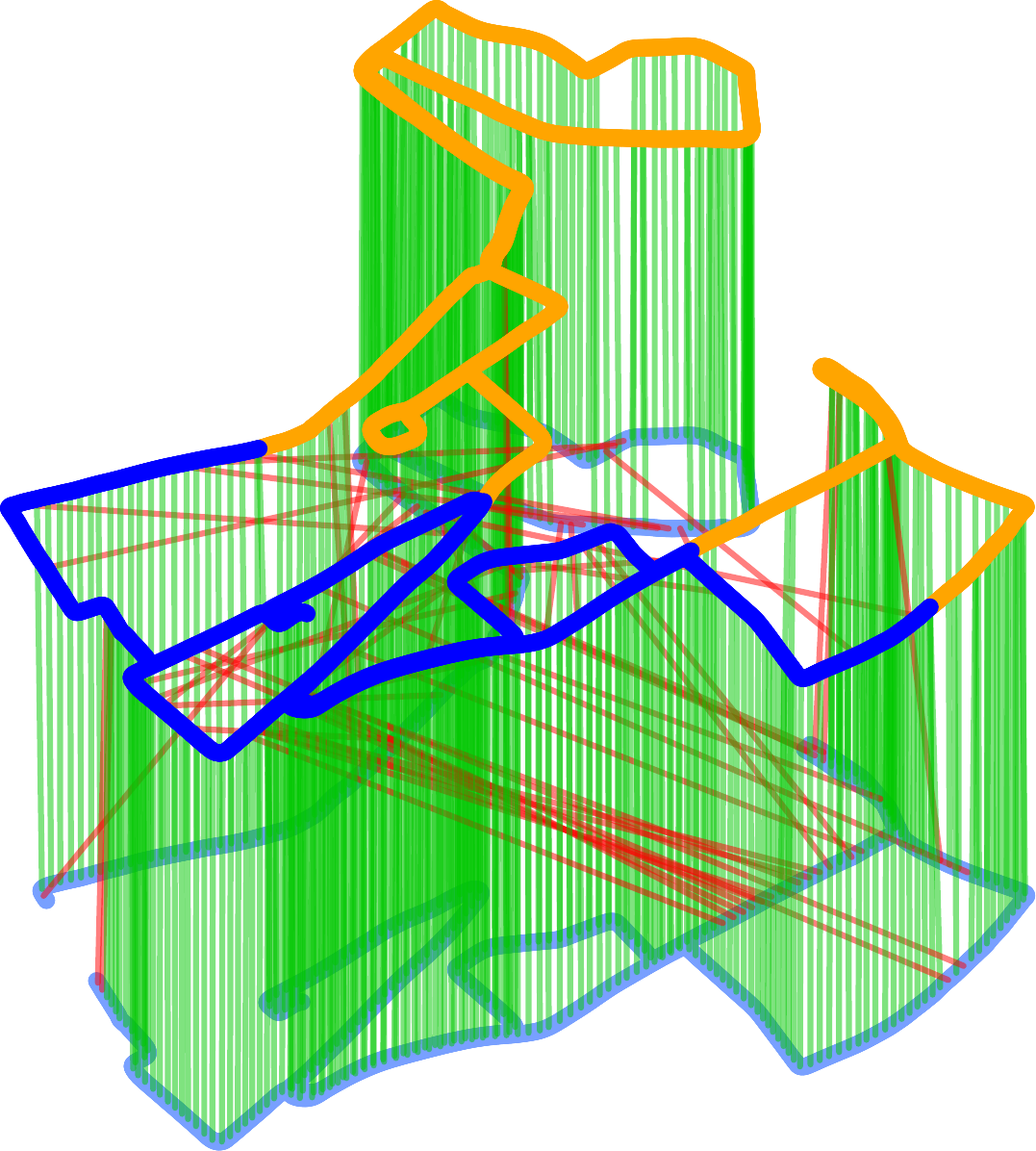}{88.9} \tabularnewline
\end{tabular}

\vspace{2pt}
\begingroup
\definecolor{querycolor}{RGB}{120,160,255}
\definecolor{samedbcolor}{RGB}{0,0,255}
\definecolor{crossdbcolor}{RGB}{255,165,0}
\small
\fcolorbox{black}{querycolor}{\rule{0pt}{6pt}\rule{6pt}{0pt}}~Day query\quad
\fcolorbox{black}{crossdbcolor}{\rule{0pt}{6pt}\rule{6pt}{0pt}}~Night DB\quad
\fcolorbox{black}{samedbcolor}{\rule{0pt}{6pt}\rule{6pt}{0pt}}~Day DB
\endgroup

\caption{Top-1 retrieval on RobotCar,
with a nighttime database (left) or an equal mix of daytime
and nighttime images (right). Lines connect queries to
retrieved images: green within 25\,m, red otherwise.}
\label{fig:test3_grid}

\vspace{-6mm}
\end{figure}

A central challenge in \acf{VPR} is
to recognize the same place across changes
in condition~\cite{nie2023training, kim2026class}.
This condition invariance is typically evaluated
using a query set captured under one condition and
a database set captured under another~\cite{sunderhauf2013we,
keetha2023anyloc, arren2020day, glover2010fab, maddern20171}.
This tests whether the method can choose the closest match in a
different condition than the query~(\Cref{fig:distractor}(a)).

However, in large environments, databases can combine images
collected by multiple robots or across multiple traversals
under different conditions~\cite{athukorala2025multi, burgard2005coordinated}.
In this setting, a different kind of condition invariance is needed.
The \ac{VPR} method should not only learn to recognize the same
place under a different condition but also ignore capture conditions
when ranking database images. In this paper, we refer to this
ideal property of ignoring capture conditions as \emph{condition blindness}.
When a method is not condition-blind, it may assign
an incorrectly high similarity score to an image
with a similar condition to the query's, even though it depicts a
different place~(\Cref{fig:distractor}(b)).

Condition blindness can affect performance in real-world scenarios.
Consider a database built from two traversals:
one traversal captured a section at night, and the other captured
a different section during the day, while the queries
are in daytime condition. In \Cref{fig:test3_grid},
performance degrades for both methods
in the mixed-condition database. However, EigenPlaces ignores
the capture conditions of database images more effectively
than AnyLoc, reversing AnyLoc's advantage in the
single-condition database.

Although Schubert~et~al.~\cite{schubert2020unsupervised}
have studied condition blindness in \ac{VPR}, they primarily frame
the problem as in-sequence condition changes where the condition
change happens within a single traversal. Instead,
we propose a new evaluation protocol, \acf{DAR},
to measure condition blindness in a more controlled manner.
\ac{DAR} holds the query-database pair constant
and adds a controlled distractor set.
This lets us isolate the effect of distractors
from changes in the query--database pair.

The same work ~\cite{schubert2020unsupervised} also uses
\ac{STD} to improve condition blindness under discrete
condition changes. In \Cref{sec:condition_suppression},
we compare \ac{STD} with two established linear concept
erasure methods, INLP and LEACE. Concept erasure is a
framework for removing information about a particular concept,
first studied in \ac{NLP} to remove gender bias in word
embeddings~\cite{bolukbasi2016man}. Although \ac{STD}
was not originally proposed as a linear
concept erasure method, it satisfies the definition of linear
concept erasure~\cite{belrose2023leace}, allowing us to compare
it with INLP and LEACE within the same framework.

\vspace{1mm}
\noindent Our contributions are:
\begin{itemize}[leftmargin=*]
    \item We construct a new evaluation protocol, \acf{DAR},
    to measure the effect of condition blindness.
    \item We study different aspects of successful distractors in
    modern \ac{VPR} methods by separating the effects of
    view overlap and condition similarity.
    \item We reinterpret \ac{STD} under the definition
    of linear concept erasure and show that established linear concept
    erasure methods can generalize better to unseen places than \ac{STD}.
\end{itemize}
\section{Related Work}

\subsection{Making VPR Condition-Invariant}
Numerous methods have been proposed to improve
the robustness of \ac{VPR} to condition changes.
Early \ac{VPR} methods relied on handcrafted representations
to handle condition changes~\cite{cummins2008fab, milford2012seqslam}.
With the advent of deep learning, NetVLAD~\cite{arandjelovic2016netvlad}
uses a triplet ranking loss to align images of the same place
under different conditions, while many recent \ac{VPR}
methods~\cite{ali2024boq, lu2024cricavpr, berton2025megaloc,
ali2023mixvpr, izquierdo2024optimal, izquierdo2024close,
lu2025selavpr++} use multi-similarity loss~\cite{wang2019multi}
for the same purpose. A parallel line formulates \ac{VPR} as a classification task,
assigning same-place images to the same class~\cite{berton2022rethinking,
berton2023eigenplaces}. Other methods directly transform images,
with earlier work relying on physics-based transformations~\cite{corke2013dealing,
maddern2014illumination} and later work on learned models~\cite{porav2018adversarial,
anoosheh2019night, clement2018train}. Meanwhile,
Schubert~et~al.~\cite{schubert2020unsupervised}
standardize NetVLAD~\cite{arandjelovic2016netvlad}
and AlexNet~\cite{krizhevsky2012imagenet} descriptors across
conditions. Since this approach directly transforms descriptors,
we compare it with other linear concept erasure methods
on recent \ac{VPR} methods.

\subsection{Evaluating Condition Invariance in VPR}
Despite numerous efforts to make \ac{VPR} methods robust to condition
changes, condition invariance is still primarily evaluated
by matching a query set from one condition to a database set
under a different condition~\cite{sunderhauf2013we, arren2020day,
keetha2023anyloc, glover2010fab, maddern20171}. This evaluates the ability
to retrieve images of the same place under a different condition,
but does not test condition blindness. Other benchmarks construct
mixed-condition databases containing images captured under diverse
conditions~\cite{warburg2020mapillary, chen2018learning, jegou2008hamming}.
However, this setting confounds condition blindness with
standard cross-condition performance by measuring both
with a single metric under an uncontrolled mix of conditions.
Schubert~et~al.~\cite{schubert2020unsupervised}
take a different approach by constructing databases with
controlled combinations of conditions. However, the query set
changes along with the database, confounding the effect of
condition blindness with changes in the query--database pair.
We take this a step further with an evaluation protocol
that keeps the query fixed and constructs a retrieval pool
with a controlled mix of conditions.

\subsection{Linear Concept Erasure in NLP}
Linear concept erasure was originally proposed
in \ac{NLP} to erase gender biases from word embeddings.
Bolukbasi~et~al.~\cite{bolukbasi2016man} identified a gender
subspace in \texttt{word2vec} embeddings and debiased them
by removing their projections onto this subspace.
Gonen~et~al.~\cite{gonen2019lipstick} subsequently showed that
this approach still leaves words clustered by gender, motivating
more rigorous methods such as INLP~\cite{ravfogel2020null} and
LEACE~\cite{belrose2023leace}. We use linear concept erasure to
suppress condition-dependent variation and improve the place-based
alignment of \ac{VPR} descriptors. %
\section{Measuring and Improving Condition Blindness in Visual Place Recognition}
\label{sec:distractor_augmented_evaluation}

\subsection{Measuring Condition Blindness with Distractor-Augmented Recall (DAR)}
In a typical cross-condition \ac{RK} evaluation
(\Cref{fig:retrieval_diagram}(a)), a single-condition
query set is matched against a single-condition database
from a different condition. This tests if the \ac{VPR} method
can correctly identify the same place under a different condition,
but it does not test whether the method prefers one condition over another.
We extend this setting by constructing,
for each query, an augmented database that contains the
cross-condition database and distractors
(\Cref{fig:retrieval_diagram}(b)). The distractors
are images captured under the same or similar
conditions as the query. This tests whether the method can
ignore condition similarity while retrieving the same place.

Because \ac{DAR}@$K$ adds a controlled distractor set to
standard \ac{RK} evaluation, comparing the two lets us isolate
the effect of distractors on retrieval performance.
Adding distractors increases the retrieval pool size, but
a variant that replaces database images with distractors
to keep the size constant yields nearly identical results
(supplementary material, Sec.~S3). We therefore use addition
as the default for \ac{DAR} because it keeps the
original database fixed, allowing us to isolate the effect
of distractors.
\definecolor{querycolor}{RGB}{120,160,255}
\definecolor{samedbcolor}{RGB}{0,0,255}
\definecolor{crossdbcolor}{RGB}{255,165,0}
\newcommand{\retrievaldiagram}[1]{%
\begin{tikzpicture}[
    >={Stealth[length=2.2mm]},
    font=\sffamily,
]

\fill[green!8] (0,0) circle (2.8);
\draw[green!55!black, dashed, thick] (0,0) circle (2.8);

\ifnum\pdfstrcmp{#1}{b}=0
    \foreach \x/\y in {4.0/1.8, 3.6/-2.4, -2.7/-3.3}{
        \begin{scope}[shift={(\x,\y)}]
            \draw[fill=samedbcolor, draw=black, thin] (-0.3,-0.225) rectangle (0.3,0.225);
        \end{scope}
    }
\fi
\foreach \x/\y in {-3.9/2.3, 1.5/-3.7, 4.4/0.3, -4.2/-1.1}{
    \begin{scope}[shift={(\x,\y)}]
        \draw[fill=crossdbcolor, draw=black, thin] (-0.3,-0.225) rectangle (0.3,0.225);
    \end{scope}
}

\def\innerfill{crossdbcolor}
\foreach \x/\y in {1.5/1.6, -1.7/1.3, -0.9/-1.9}{
    \begin{scope}[shift={(\x,\y)}]
        \draw[fill=\innerfill, draw=black, thin] (-0.33,-0.24) rectangle (0.33,0.24);
    \end{scope}
}

\begin{scope}[shift={(0,0)}]
    \draw[fill=blue!25, draw=blue!70!black, thick] (-0.56,-0.44) rectangle (0.56,0.36);
    \draw[fill=blue!38, draw=blue!70!black, thick] (-0.36,0.36) rectangle (0.36,0.72);
    \fill[blue!70!black] (-0.16,0.54) circle (0.056);
    \fill[blue!70!black] ( 0.16,0.54) circle (0.056);
    \draw[blue!70!black, thick] (0,0.72) -- (0,0.98);
    \fill[blue!70!black] (0,0.98) circle (0.0907);
    \fill[black] (-0.4,-0.56) circle (0.12);
    \fill[black] ( 0.4,-0.56) circle (0.12);
\end{scope}
\end{tikzpicture}%
}

\newcommand{\retrievallegend}{%
\begin{tikzpicture}[baseline=-0.6ex]
    \draw[fill=crossdbcolor, draw=black, thin] (0,0) rectangle (0.22,0.22);
    \node[anchor=west, font=\small] at (0.3,0.11) {Cross-condition DB};

    \draw[fill=samedbcolor, draw=black, thin] (3.7,0) rectangle (3.92,0.22);
    \node[anchor=west, font=\small] at (4.0,0.11) {Distractors};
\end{tikzpicture}%
}

\begin{figure}[t!]
\centering
\newcommand{\panelframe}[1]{%
    \tikz[baseline]{\node[draw, rounded corners=2pt, line width=0.4pt,
        inner sep=2pt]{#1};}%
}
\begin{tabular}{@{}c@{\hspace{1mm}}c@{}}
    {\small (a) Recall} & {\small (b) DAR} \\
    \panelframe{\resizebox{0.46\columnwidth}{!}{\retrievaldiagram{a}}} &
    \panelframe{\resizebox{0.46\columnwidth}{!}{\retrievaldiagram{b}}} \\
\end{tabular}

\vspace{2pt}
\retrievallegend
\caption{The dashed circle marks the positive-match
threshold around the robot pose corresponding to the query.
\textbf{(a)} Recall uses a cross-condition database.
\textbf{(b)} \ac{DAR} adds distractors among the negatives
outside the circle.}
\label{fig:retrieval_diagram}

\vspace{-6mm}
\end{figure}

\subsection{Formalization of DAR}
Let $\mathcal{Q}=\{q_i\}_{i=1}^{N_Q}$,
$\mathcal{C}=\{c_j\}_{j=1}^{N_C}$, and
$\mathcal{D}=\{d_l\}_{l=1}^{N_D}$ 
denote, respectively, the query, the cross-condition database, and
the distractor image set, with associated positions
$\mathcal{Y}^{\mathcal{Q}}=\{y_i^{\mathcal{Q}}\}_{i=1}^{N_Q}$,
$\mathcal{Y}^{\mathcal{C}}=\{y_j^{\mathcal{C}}\}_{j=1}^{N_C}$, and
$\mathcal{Y}^{\mathcal{D}}=\{y_l^{\mathcal{D}}\}_{l=1}^{N_D}$,
where all positions lie in $\mathbb{R}^n$.

Using distance threshold $d_{\mathrm{thr}}$, the cross-condition positive set
and the per-query distractor set for $q_i$ are defined as
\begin{equation}
\label{eq:positive_and_distractor_sets}
\begin{aligned}
    \mathcal{P}_i
    &= \left\{ c_j \in \mathcal{C} \;\middle|\;
    \lVert y_i^{\mathcal{Q}} - y_j^{\mathcal{C}} \rVert_2
    \leq d_{\mathrm{thr}}
    \right\}, \\
    \mathcal{N}_i
    &= \left\{ d_l \in \mathcal{D} \;\middle|\;
    \lVert y_i^{\mathcal{Q}} - y_l^{\mathcal{D}} \rVert_2
    > d_{\mathrm{thr}}
    \right\}.
\end{aligned}
\end{equation}
The augmented database set for query $q_i$ is the union of the entire
cross-condition database and the distractor set:
\[
    \mathcal{A}_i = \mathcal{C} \cup \mathcal{N}_i.
\]

For a query $q_i$, the method ranks the database images $\mathcal{A}_i$ by their
similarity to $q_i$. Writing $\mathrm{rank}_{q_i}(a) \in \{1, \dots, |\mathcal{A}_i|\}$
for the position of $a$ in this ranking (rank $1$ being the top match), the
top-$K$ retrieved set is
\[
    \mathcal{R}_i^{K} = \{\, a \in \mathcal{A}_i :
    \mathrm{rank}_{q_i}(a) \le K \,\}.
\]
The method must rank a positive from $\mathcal{P}_i$ above the rest of
$\mathcal{A}_i$, particularly above the distractors in $\mathcal{N}_i$, whose
condition is similar to the query's and which are thus visually easier to match.

\ac{DAR}@$K$ is the fraction of queries for which the top-$K$ retrievals contain
at least one cross-condition positive:
\[
    \mathrm{DAR@}K =
    \frac{1}{N_Q}
    \sum_{i=1}^{N_Q}
    \left[\,\mathcal{R}_i^{K} \cap \mathcal{P}_i \neq \emptyset\,\right].
\]
A method's DAR@K is upper-bounded by its R@K,
\(\text{DAR@K} \leq \text{R@K}\), since adding distractors can only
displace cross-condition positives from the top-$K$.

\subsection{Improving Condition Blindness with Linear Concept Erasure}
\newcommand{\umapcornerlabel}[1]{%
  \begin{tikzpicture}[inner sep=0pt, outer sep=0pt]
    \node[draw, line width=\fboxrule, inner sep=0pt] (img) {%
      \adjincludegraphics[width=0.36\columnwidth, keepaspectratio,
                          trim={0 0 0 0}, clip]{#1}};
  \end{tikzpicture}%
}
\newcommand{\suppressionarrow}{%
  \begin{tikzpicture}[inner sep=0pt, outer sep=0pt]
    \draw[-{Triangle[length=3mm, width=3mm]}, line width=3pt] (0,0) -- (0.5,0);
  \end{tikzpicture}%
}
\begin{figure}[t!]
\centering
\setlength{\tabcolsep}{2pt}
\renewcommand{\arraystretch}{1.0}
\setlength{\fboxsep}{0pt}
\begin{tabular}{@{}m{0.36\columnwidth} c m{0.36\columnwidth}@{}}
\centering\small Original & & \centering\small Erased \tabularnewline
\umapcornerlabel{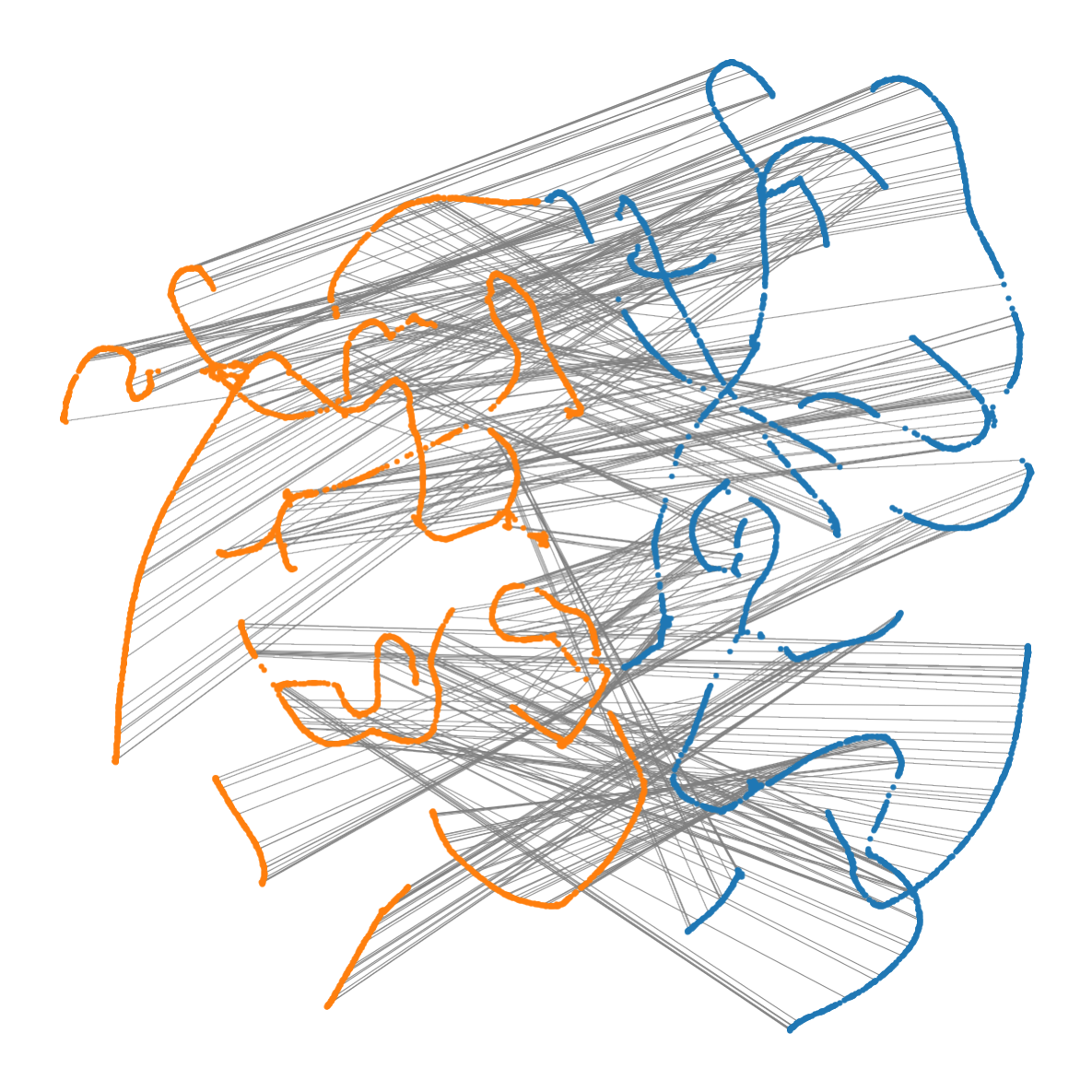} &
\suppressionarrow &
\umapcornerlabel{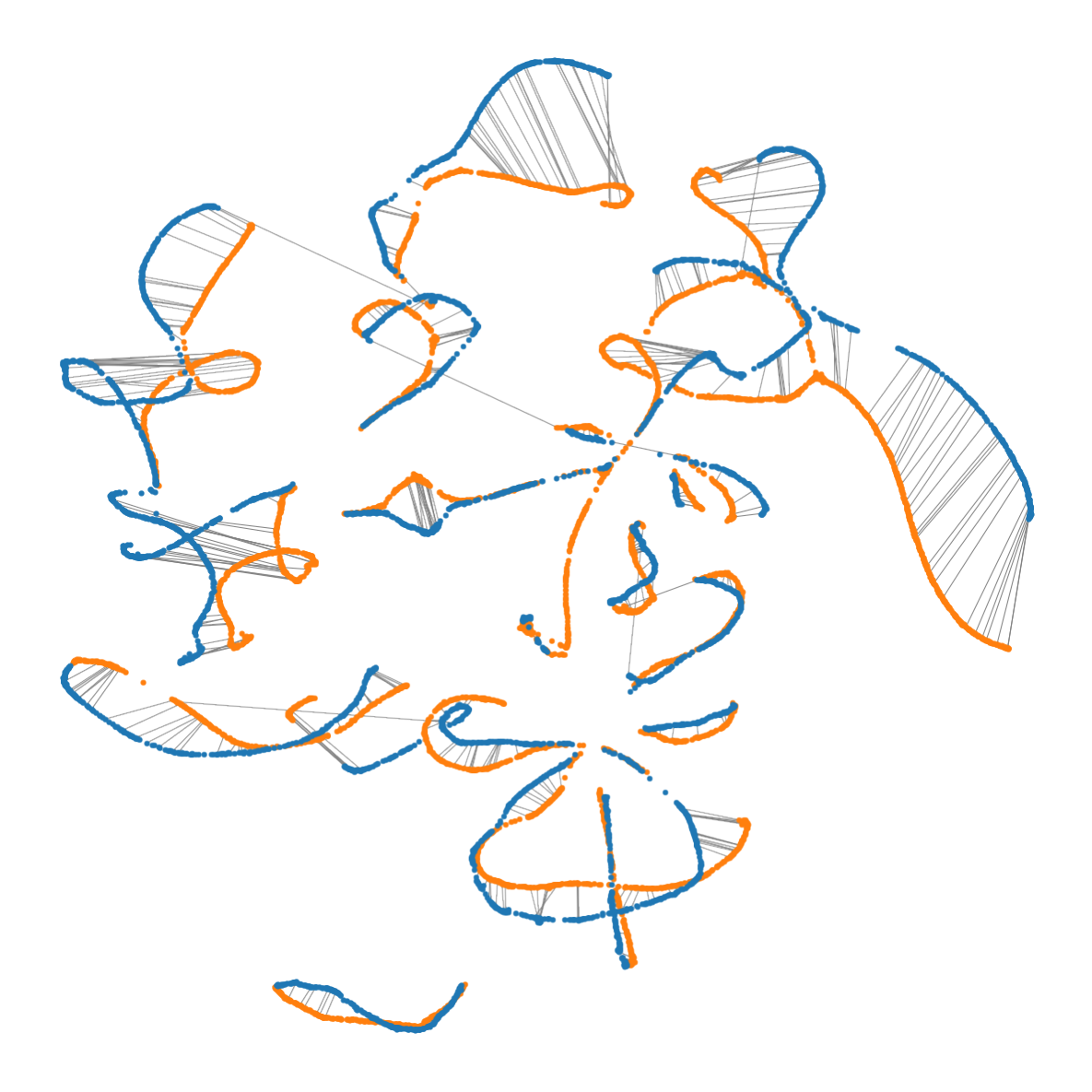} \tabularnewline
\end{tabular}

\vspace{2pt}
\begingroup
\definecolor{querycolor}{RGB}{31,119,180}
\definecolor{dbcolor}{RGB}{255,127,14}
\small
\fcolorbox{black}{querycolor}{\rule{0pt}{6pt}\rule{6pt}{0pt}}~Day\quad
\fcolorbox{black}{dbcolor}{\rule{0pt}{6pt}\rule{6pt}{0pt}}~Night
\endgroup

\caption{UMAP of AnyLoc descriptors on RobotCar before and after INLP.
Gray lines connect spatial mutual nearest-neighbor pairs.
INLP reduces condition-based separation and improves
place-based alignment.}
\label{fig:umap_grid}
\vspace{-5mm}
\end{figure}
 Linear concept erasure methods aim to remove information
about a particular concept from a representation. Linear
concept erasure methods do so by hiding the concept labels
from linear classifiers, a property that Belrose~et~al.~\cite{belrose2023leace}
formalized as \emph{linear guardedness}. To achieve linear guardedness,
linear concept erasure methods suppress variation associated with
the target concept. In \ac{VPR}, we can use linear
concept erasure methods to realign the descriptors by place
and reduce condition-specific variation (\Cref{fig:umap_grid}).

\begin{table*}[t!]
\centering
\begingroup
\setlength{\tabcolsep}{6pt}
\renewcommand{\arraystretch}{1.05}
\caption{
Positive thresholds define correct retrievals; exclusion
thresholds define the radius within which distractors
are removed for \acs{DARx}@1. Both are measured in meters
or frames for frame-aligned datasets.
XB3=Bumblebee XB3 camera; GSV=Google Street View.
}
\label{tab:datasets}
\footnotesize
\begin{tabular}{l l l r r l l l}
\toprule
Dataset                                & Query                        & Database                     & \#Query  & \#Database & \makecell[l]{Positive\\threshold} & \makecell[l]{Exclusion\\threshold} & Condition shift \\
\midrule
AmsterTime~\cite{yildiz2022amstertime} & \texttt{new}                 & \texttt{old}                 & 1{,}231  & 1{,}231    & exact match   & --            & modern$\rightarrow$historical \\
Gardens Point~\cite{arren2020day}      & \texttt{day\_right}          & \texttt{night\_right}        & 200      & 200        & $\pm$2 frames & 10 frames     & day/RGB$\rightarrow$night/B\&W \\
Nordland~\cite{sunderhauf2013we}       & \texttt{Summer}              & \texttt{Winter}              & 27{,}592 & 27{,}592   & $\pm$10 frames & 1{,}000 frames & sum$\rightarrow$win \\
RobotCar~\cite{maddern20171}           & \texttt{2014-11-18} & \texttt{2014-12-16} & 6{,}021  & 7{,}513    & 25\,m         & 100\,m        & day$\rightarrow$night \\
StLuciaMToD~\cite{glover2010fab}       & \texttt{100909\_0845}        & \texttt{180809\_1545}        & 1{,}619  & 1{,}615    & 25\,m         & 100\,m        & morn.$\rightarrow$aftn. \\
SVOX~\cite{berton2021adaptive}         & \texttt{queries\_overcast}   & \texttt{gallery}             & 872      & 17{,}166   & 25\,m         & 100\,m        & XB3$\rightarrow$GSV \\
\bottomrule
\end{tabular}
\endgroup

\end{table*}
\begin{table*}[t!]
\centering
\begingroup
\setlength{\tabcolsep}{3pt}
\renewcommand{\arraystretch}{0.95}
\caption{R@1, DAR@1, and \acs{DARx}@1 (\%) across six datasets; higher
is better. \acs{DARx}@1 excludes distractors within each query's
exclusion threshold (\Cref{tab:datasets}) and is omitted for
AmsterTime, which lacks positions. \textbf{Bold}: best per column.}
\label{tab:recall_combined}
\resizebox{\textwidth}{!}{\footnotesize%
\begin{tabular}{l cc ccc ccc ccc ccc ccc}
\toprule
 & \multicolumn{2}{c}{AmsterTime}
 & \multicolumn{3}{c}{Gardens Point}
 & \multicolumn{3}{c}{Nordland}
 & \multicolumn{3}{c}{RobotCar}
 & \multicolumn{3}{c}{StLuciaMToD}
 & \multicolumn{3}{c}{SVOX} \\
\cmidrule(lr){2-3} \cmidrule(lr){4-6} \cmidrule(lr){7-9} \cmidrule(lr){10-12} \cmidrule(lr){13-15} \cmidrule(lr){16-18}
\textbf{Method}
 & R@1 & DAR@1
 & R@1 & DAR@1 & \acs{DARx}@1
 & R@1 & DAR@1 & \acs{DARx}@1
 & R@1 & DAR@1 & \acs{DARx}@1
 & R@1 & DAR@1 & \acs{DARx}@1
 & R@1 & DAR@1 & \acs{DARx}@1 \\
\midrule
AnyLoc & 44.5 & 10.7 & 93.5 & 15.0 & 81.0 & 22.7 & 2.23 & 3.50 & 95.9 & 4.21 & 20.2 & 79.9 & 44.2 & 60.6 & 89.3 & 1.38 & 5.62 \\
BoQ & 68.6 & 49.4 & \textbf{99.0} & \textbf{37.5} & \textbf{99.0} & 88.5 & 44.4 & 54.4 & \textbf{99.5} & 85.3 & \textbf{99.4} & 99.9 & 99.6 & 99.9 & 98.4 & 86.9 & 97.1 \\
CosPlace & 55.5 & 40.4 & 77.0 & 18.5 & 63.5 & 47.8 & 15.1 & 18.7 & 91.5 & 38.1 & 60.7 & 99.6 & 98.9 & 99.6 & 89.7 & 76.6 & 87.4 \\
CricaVPR & 68.1 & 41.9 & 95.5 & 29.5 & 95.0 & 83.4 & 19.9 & 28.1 & 98.8 & 51.4 & 94.4 & \textbf{100} & 97.7 & 99.7 & 97.0 & 56.2 & 90.2 \\
EigenPlaces & 54.8 & 37.2 & 80.0 & 16.0 & 71.5 & 53.1 & 15.3 & 18.7 & 91.5 & 37.0 & 60.0 & 99.3 & 98.4 & 99.0 & 91.6 & 77.1 & 88.3 \\
MegaLoc & \textbf{68.7} & \textbf{50.3} & 93.0 & 12.5 & 90.5 & 92.9 & 73.1 & 79.5 & 98.8 & 79.7 & 98.3 & 99.9 & 99.6 & 99.9 & 97.7 & 90.1 & 97.6 \\
MixVPR & 43.8 & 25.3 & 85.5 & 9.00 & 67.0 & 59.9 & 9.75 & 12.4 & 89.9 & 32.5 & 65.1 & 99.5 & 97.7 & 99.2 & 96.1 & 71.2 & 91.9 \\
SALAD & 63.8 & 42.2 & 98.0 & 34.0 & 97.0 & 84.1 & 32.1 & 43.3 & 99.2 & 83.1 & 99.2 & 99.9 & 99.1 & 99.9 & 98.3 & 83.1 & 97.7 \\
SALAD-CM & 62.4 & 45.6 & 93.0 & 26.0 & 93.0 & \textbf{96.0} & \textbf{87.7} & \textbf{91.1} & 99.3 & \textbf{93.6} & 99.3 & \textbf{100} & \textbf{99.9} & 99.9 & \textbf{98.5} & 91.5 & 98.0 \\
SelaVPR & 57.0 & 38.5 & 95.5 & 32.0 & 94.0 & 77.5 & 18.0 & 26.9 & 98.2 & 71.3 & 94.7 & 99.5 & 98.2 & 99.4 & 95.8 & 71.9 & 91.6 \\
SelaVPR++ & 66.3 & 45.7 & 97.0 & 30.5 & 97.0 & \textbf{96.0} & 70.4 & 80.4 & 99.4 & 92.0 & \textbf{99.4} & \textbf{100} & \textbf{99.9} & \textbf{100} & \textbf{98.5} & \textbf{92.3} & \textbf{98.5} \\
\bottomrule
\end{tabular}%
}
\endgroup
\vspace{-3mm}
\end{table*}
 \subsection{STD is Secretly a Linear Concept Erasure Method}
\label{sec:condition_suppression}
Schubert~et~al.~\cite{schubert2020unsupervised} used
\ac{STD} to improve condition blindness in discrete condition
changes. Although this approach was developed independently of
linear concept erasure methods in
\ac{NLP}~\cite{bolukbasi2016man,ravfogel2020null,belrose2023leace},
it can be viewed as pursuing the same goal:
hiding condition labels from linear classifiers.
\ac{STD} equalizes the condition
means to the zero vector, which Belrose~et~al.~\cite{belrose2023leace}
proved equivalent to linear guardedness.
Under this definition, \ac{STD} qualifies as a linear concept
erasure method, along with established linear concept erasure
methods such as INLP~\cite{ravfogel2020null} and LEACE~\cite{belrose2023leace}.
Therefore, we compare \ac{STD}, INLP, and LEACE as linear
concept erasure methods for suppressing condition-dependent
variation in \ac{VPR} descriptors.

\subsection{Concept Erasure Methods}
We compare three transforms
($r:\mathbb{R}^{d}\to\mathbb{R}^{d}$) applied to each
descriptor ($x$). INLP~\cite{ravfogel2020null} and
LEACE~\cite{belrose2023leace} fit transformations
directly from descriptors $X$ and condition labels
$Z$, whereas \ac{STD}~\cite{schubert2020unsupervised}
uses $Z$ only to partition $X$ and estimates descriptor
statistics separately for each condition.

\paragraph{INLP.}
INLP iteratively fits a linear classifier to predict
the condition and obtains a projection $P_i$ onto
its null space at each iteration $i$. After $m$ iterations,
\begin{equation*}
r_{\mathrm{INLP}}(x) = P_m \cdots P_1 x.
\end{equation*}

\paragraph{LEACE.}
LEACE defines the transformation using the whitening matrix
$W=(\Sigma_{XX}^{1/2})^{+}$ and the orthogonal projection
$P_{W\Sigma_{XZ}}=(W\Sigma_{XZ})(W\Sigma_{XZ})^{+}$:
\begin{equation*}
r_{\mathrm{LEACE}}(x)
= x - W^{+} P_{W\Sigma_{XZ}} W\bigl(x-\mathbb{E}[X]\bigr).
\end{equation*}

\paragraph{STD.}
\ac{STD} standardizes a descriptor $x_c$ from condition $c$
using the condition-specific mean $\mu_c$ and standard
deviation $\sigma_c$:
\begin{equation*}
r_{\mathrm{STD}}(x_c)
= \frac{x_c-\mu_c}{\sigma_c}.
\end{equation*}
\section{Experiments}
\begin{figure*}[t!]
\centering
\begingroup
\definecolor{nbpart}{HTML}{2A78D6}
\definecolor{cdpart}{HTML}{EB6834}
\pgfplotsset{
  dumbbell/.style={
    scale only axis, width=0.289\textwidth, height=104pt,
    xmin=0, xmax=101, ymin=-12.4, ymax=0.6,
    xtick={0,50,100}, tick label style={font=\scriptsize},
    title style={font=\scriptsize, yshift=-0.5ex},
    xmajorgrids, major grid style={line width=.1pt,draw=gray!25},
    ytick style={draw=none}, axis x line*=bottom, axis y line*=left, clip=false,
  },
}
\begin{tikzpicture}
\begin{groupplot}[
  group style={group size=3 by 1, horizontal sep=0.25cm, yticklabels at=edge left},
  dumbbell,
  ytick={0,-1,-2,-3,-4.6,-5.6,-7.2,-8.2,-9.8,-10.8,-11.8},
  yticklabels={{AnyLoc},{CosPlace},{EigenPlaces},{MixVPR},{CricaVPR},{SelaVPR},{BoQ},{SALAD},{MegaLoc},{SALAD-CM},{SelaVPR++}},
]
\nextgroupplot[title={Gardens Point\strut},
  legend to name=leg:dar_decomposition, legend columns=-1,
  legend style={font=\scriptsize, draw=none, /tikz/every even column/.append style={column sep=4pt}}]
  \draw[gray!60, dashed, line width=0.3pt] (axis cs:0,-3.8) -- (axis cs:101,-3.8);
  \draw[gray!60, dashed, line width=0.3pt] (axis cs:0,-6.4) -- (axis cs:101,-6.4);
  \draw[gray!60, dashed, line width=0.3pt] (axis cs:0,-9) -- (axis cs:101,-9);
  \draw[nbpart, line width=1pt, line cap=round] (axis cs:15.0,0) -- (axis cs:81.0,0); \draw[cdpart, line width=1pt, line cap=round] (axis cs:81.0,0) -- (axis cs:93.5,0); %
  \draw[nbpart, line width=1pt, line cap=round] (axis cs:18.5,-1) -- (axis cs:63.5,-1); \draw[cdpart, line width=1pt, line cap=round] (axis cs:63.5,-1) -- (axis cs:77.0,-1); %
  \draw[nbpart, line width=1pt, line cap=round] (axis cs:16.0,-2) -- (axis cs:71.5,-2); \draw[cdpart, line width=1pt, line cap=round] (axis cs:71.5,-2) -- (axis cs:80.0,-2); %
  \draw[nbpart, line width=1pt, line cap=round] (axis cs:9.00,-3) -- (axis cs:67.0,-3); \draw[cdpart, line width=1pt, line cap=round] (axis cs:67.0,-3) -- (axis cs:85.5,-3); %
  \draw[nbpart, line width=1pt, line cap=round] (axis cs:29.5,-4.6) -- (axis cs:95.0,-4.6); \draw[cdpart, line width=1pt, line cap=round] (axis cs:95.0,-4.6) -- (axis cs:95.5,-4.6); %
  \draw[nbpart, line width=1pt, line cap=round] (axis cs:32.0,-5.6) -- (axis cs:94.0,-5.6); \draw[cdpart, line width=1pt, line cap=round] (axis cs:94.0,-5.6) -- (axis cs:95.5,-5.6); %
  \draw[nbpart, line width=1pt, line cap=round] (axis cs:37.5,-7.2) -- (axis cs:99.0,-7.2); \draw[cdpart, line width=1pt, line cap=round] (axis cs:99.0,-7.2) -- (axis cs:99.0,-7.2); %
  \draw[nbpart, line width=1pt, line cap=round] (axis cs:34.0,-8.2) -- (axis cs:97.0,-8.2); \draw[cdpart, line width=1pt, line cap=round] (axis cs:97.0,-8.2) -- (axis cs:98.0,-8.2); %
  \draw[nbpart, line width=1pt, line cap=round] (axis cs:12.5,-9.8) -- (axis cs:90.5,-9.8); \draw[cdpart, line width=1pt, line cap=round] (axis cs:90.5,-9.8) -- (axis cs:93.0,-9.8); %
  \draw[nbpart, line width=1pt, line cap=round] (axis cs:26.0,-10.8) -- (axis cs:93.0,-10.8); \draw[cdpart, line width=1pt, line cap=round] (axis cs:93.0,-10.8) -- (axis cs:93.0,-10.8); %
  \draw[nbpart, line width=1pt, line cap=round] (axis cs:30.5,-11.8) -- (axis cs:97.0,-11.8); \draw[cdpart, line width=1pt, line cap=round] (axis cs:97.0,-11.8) -- (axis cs:97.0,-11.8); %
  \addplot[only marks, mark=o, mark size=1.1pt, mark options={black, fill=white, line width=0.4pt}] coordinates {(15.0,0) (18.5,-1) (16.0,-2) (9.00,-3) (29.5,-4.6) (32.0,-5.6) (37.5,-7.2) (34.0,-8.2) (12.5,-9.8) (26.0,-10.8) (30.5,-11.8)};
  \addplot[only marks, mark=diamond*, mark size=1.3pt, black] coordinates {(81.0,0) (63.5,-1) (71.5,-2) (67.0,-3) (95.0,-4.6) (94.0,-5.6) (99.0,-7.2) (97.0,-8.2) (90.5,-9.8) (93.0,-10.8) (97.0,-11.8)};
  \addplot[only marks, mark=|, mark size=1.8pt, black, line width=0.7pt] coordinates {(93.5,0) (77.0,-1) (80.0,-2) (85.5,-3) (95.5,-4.6) (95.5,-5.6) (99.0,-7.2) (98.0,-8.2) (93.0,-9.8) (93.0,-10.8) (97.0,-11.8)};
  \addlegendentry{DAR@1}\addlegendentry{\acs{DARx}@1}\addlegendentry{R@1}
  \addlegendimage{nbpart, line width=1pt}\addlegendentry{Closed by exclusion}
  \addlegendimage{cdpart, line width=1pt}\addlegendentry{Remaining gap}
\nextgroupplot[title={RobotCar\strut}, xtick={50,100}]
  \draw[gray!60, dashed, line width=0.3pt] (axis cs:0,-3.8) -- (axis cs:101,-3.8);
  \draw[gray!60, dashed, line width=0.3pt] (axis cs:0,-6.4) -- (axis cs:101,-6.4);
  \draw[gray!60, dashed, line width=0.3pt] (axis cs:0,-9) -- (axis cs:101,-9);
  \draw[nbpart, line width=1pt, line cap=round] (axis cs:4.21,0) -- (axis cs:20.2,0); \draw[cdpart, line width=1pt, line cap=round] (axis cs:20.2,0) -- (axis cs:95.9,0); %
  \draw[nbpart, line width=1pt, line cap=round] (axis cs:38.1,-1) -- (axis cs:60.7,-1); \draw[cdpart, line width=1pt, line cap=round] (axis cs:60.7,-1) -- (axis cs:91.5,-1); %
  \draw[nbpart, line width=1pt, line cap=round] (axis cs:37.0,-2) -- (axis cs:60.0,-2); \draw[cdpart, line width=1pt, line cap=round] (axis cs:60.0,-2) -- (axis cs:91.5,-2); %
  \draw[nbpart, line width=1pt, line cap=round] (axis cs:32.5,-3) -- (axis cs:65.1,-3); \draw[cdpart, line width=1pt, line cap=round] (axis cs:65.1,-3) -- (axis cs:89.9,-3); %
  \draw[nbpart, line width=1pt, line cap=round] (axis cs:51.4,-4.6) -- (axis cs:94.4,-4.6); \draw[cdpart, line width=1pt, line cap=round] (axis cs:94.4,-4.6) -- (axis cs:98.8,-4.6); %
  \draw[nbpart, line width=1pt, line cap=round] (axis cs:71.3,-5.6) -- (axis cs:94.7,-5.6); \draw[cdpart, line width=1pt, line cap=round] (axis cs:94.7,-5.6) -- (axis cs:98.2,-5.6); %
  \draw[nbpart, line width=1pt, line cap=round] (axis cs:85.3,-7.2) -- (axis cs:99.4,-7.2); \draw[cdpart, line width=1pt, line cap=round] (axis cs:99.4,-7.2) -- (axis cs:99.5,-7.2); %
  \draw[nbpart, line width=1pt, line cap=round] (axis cs:83.1,-8.2) -- (axis cs:99.2,-8.2); \draw[cdpart, line width=1pt, line cap=round] (axis cs:99.2,-8.2) -- (axis cs:99.2,-8.2); %
  \draw[nbpart, line width=1pt, line cap=round] (axis cs:79.7,-9.8) -- (axis cs:98.3,-9.8); \draw[cdpart, line width=1pt, line cap=round] (axis cs:98.3,-9.8) -- (axis cs:98.8,-9.8); %
  \draw[nbpart, line width=1pt, line cap=round] (axis cs:93.6,-10.8) -- (axis cs:99.3,-10.8); \draw[cdpart, line width=1pt, line cap=round] (axis cs:99.3,-10.8) -- (axis cs:99.3,-10.8); %
  \draw[nbpart, line width=1pt, line cap=round] (axis cs:92.0,-11.8) -- (axis cs:99.4,-11.8); \draw[cdpart, line width=1pt, line cap=round] (axis cs:99.4,-11.8) -- (axis cs:99.4,-11.8); %
  \addplot[only marks, mark=o, mark size=1.1pt, mark options={black, fill=white, line width=0.4pt}] coordinates {(4.21,0) (38.1,-1) (37.0,-2) (32.5,-3) (51.4,-4.6) (71.3,-5.6) (85.3,-7.2) (83.1,-8.2) (79.7,-9.8) (93.6,-10.8) (92.0,-11.8)};
  \addplot[only marks, mark=diamond*, mark size=1.3pt, black] coordinates {(20.2,0) (60.7,-1) (60.0,-2) (65.1,-3) (94.4,-4.6) (94.7,-5.6) (99.4,-7.2) (99.2,-8.2) (98.3,-9.8) (99.3,-10.8) (99.4,-11.8)};
  \addplot[only marks, mark=|, mark size=1.8pt, black, line width=0.7pt] coordinates {(95.9,0) (91.5,-1) (91.5,-2) (89.9,-3) (98.8,-4.6) (98.2,-5.6) (99.5,-7.2) (99.2,-8.2) (98.8,-9.8) (99.3,-10.8) (99.4,-11.8)};
\nextgroupplot[title={Nordland\strut}, xtick={50,100}]
  \draw[gray!60, dashed, line width=0.3pt] (axis cs:0,-3.8) -- (axis cs:101,-3.8);
  \draw[gray!60, dashed, line width=0.3pt] (axis cs:0,-6.4) -- (axis cs:101,-6.4);
  \draw[gray!60, dashed, line width=0.3pt] (axis cs:0,-9) -- (axis cs:101,-9);
  \draw[nbpart, line width=1pt, line cap=round] (axis cs:2.23,0) -- (axis cs:3.50,0); \draw[cdpart, line width=1pt, line cap=round] (axis cs:3.50,0) -- (axis cs:22.7,0); %
  \draw[nbpart, line width=1pt, line cap=round] (axis cs:15.1,-1) -- (axis cs:18.7,-1); \draw[cdpart, line width=1pt, line cap=round] (axis cs:18.7,-1) -- (axis cs:47.8,-1); %
  \draw[nbpart, line width=1pt, line cap=round] (axis cs:15.3,-2) -- (axis cs:18.7,-2); \draw[cdpart, line width=1pt, line cap=round] (axis cs:18.7,-2) -- (axis cs:53.1,-2); %
  \draw[nbpart, line width=1pt, line cap=round] (axis cs:9.75,-3) -- (axis cs:12.4,-3); \draw[cdpart, line width=1pt, line cap=round] (axis cs:12.4,-3) -- (axis cs:59.9,-3); %
  \draw[nbpart, line width=1pt, line cap=round] (axis cs:19.9,-4.6) -- (axis cs:28.1,-4.6); \draw[cdpart, line width=1pt, line cap=round] (axis cs:28.1,-4.6) -- (axis cs:83.4,-4.6); %
  \draw[nbpart, line width=1pt, line cap=round] (axis cs:18.0,-5.6) -- (axis cs:26.9,-5.6); \draw[cdpart, line width=1pt, line cap=round] (axis cs:26.9,-5.6) -- (axis cs:77.5,-5.6); %
  \draw[nbpart, line width=1pt, line cap=round] (axis cs:44.4,-7.2) -- (axis cs:54.4,-7.2); \draw[cdpart, line width=1pt, line cap=round] (axis cs:54.4,-7.2) -- (axis cs:88.5,-7.2); %
  \draw[nbpart, line width=1pt, line cap=round] (axis cs:32.1,-8.2) -- (axis cs:43.3,-8.2); \draw[cdpart, line width=1pt, line cap=round] (axis cs:43.3,-8.2) -- (axis cs:84.1,-8.2); %
  \draw[nbpart, line width=1pt, line cap=round] (axis cs:73.1,-9.8) -- (axis cs:79.5,-9.8); \draw[cdpart, line width=1pt, line cap=round] (axis cs:79.5,-9.8) -- (axis cs:92.9,-9.8); %
  \draw[nbpart, line width=1pt, line cap=round] (axis cs:87.7,-10.8) -- (axis cs:91.1,-10.8); \draw[cdpart, line width=1pt, line cap=round] (axis cs:91.1,-10.8) -- (axis cs:96.0,-10.8); %
  \draw[nbpart, line width=1pt, line cap=round] (axis cs:70.4,-11.8) -- (axis cs:80.4,-11.8); \draw[cdpart, line width=1pt, line cap=round] (axis cs:80.4,-11.8) -- (axis cs:96.0,-11.8); %
  \addplot[only marks, mark=o, mark size=1.1pt, mark options={black, fill=white, line width=0.4pt}] coordinates {(2.23,0) (15.1,-1) (15.3,-2) (9.75,-3) (19.9,-4.6) (18.0,-5.6) (44.4,-7.2) (32.1,-8.2) (73.1,-9.8) (87.7,-10.8) (70.4,-11.8)};
  \addplot[only marks, mark=diamond*, mark size=1.3pt, black] coordinates {(3.50,0) (18.7,-1) (18.7,-2) (12.4,-3) (28.1,-4.6) (26.9,-5.6) (54.4,-7.2) (43.3,-8.2) (79.5,-9.8) (91.1,-10.8) (80.4,-11.8)};
  \addplot[only marks, mark=|, mark size=1.8pt, black, line width=0.7pt] coordinates {(22.7,0) (47.8,-1) (53.1,-2) (59.9,-3) (83.4,-4.6) (77.5,-5.6) (88.5,-7.2) (84.1,-8.2) (92.9,-9.8) (96.0,-10.8) (96.0,-11.8)};
\end{groupplot}
\node[font=\scriptsize, anchor=north] at ($(group c2r1.south)+(0,-0.45cm)$) {Recall (\%)};
\end{tikzpicture}\\[-1mm]
\pgfplotslegendfromname{leg:dar_decomposition}
\endgroup
\caption{Decomposition of the \ac{R1}--\ac{DAR}@1 gap.
Blue shows the gap closed by excluding nearby distractors
(\ac{DAR}@1 to \acs{DARx}@1); orange shows the remainder.
Methods are ordered by the groups in \Cref{subsec:dar_decomposition}.
SVOX and \acs{StLuciaMToD} are shown in the supplementary material (Sec.~S1).}
\label{fig:dar_decomposition}
\vspace{-3mm}
\end{figure*}
 \subsection{Setup}
\label{subsec:setup}

\paragraph{VPR Methods and Datasets.}
\Cref{tab:datasets} summarizes the datasets.
We evaluate eleven \ac{VPR} methods with publicly released
weights. Dataset preparation details and \ac{VPR} method configurations
are provided in the supplementary material (Secs.~S6 and~S7).
Following Schubert~et~al.~\cite{schubert2020unsupervised},
we use the query set as the distractor pool for the main
results (\Cref{subsec:main_results}) to obtain distractors
that match the query's condition. However, with
the query set as the pool, the hardest distractors
for a query are often the images captured
near it (\Cref{tab:best_distractor_dist}).
Although these images meet the definition of
successful distractors, their visual overlap
with the true positive can additionally contribute
to their effectiveness as distractors. To separate this effect,
we also report \ac{DARx}@1, computed on a reduced distractor
set that excludes distractors within the exclusion threshold
of the query. The thresholds are listed in \Cref{tab:datasets}.

\begin{table}[t!]
\centering
\begingroup
\setlength{\tabcolsep}{4pt}
\renewcommand{\arraystretch}{1.05}
\caption{Median query-to-distractor distance when
a distractor ranks first. Values are averaged across
eleven methods. AmsterTime is omitted because it
lacks positions.}
\label{tab:best_distractor_dist}
\footnotesize
\begin{tabular}{l rr}
\toprule
Dataset       & Positive threshold & Median \\
\midrule
Gardens Point & 2\,fr              & 3.0\,frames  \\
Nordland      & 10\,fr             & 385\,frames  \\
RobotCar      & 25\,m              & 26.8\,m  \\
StLuciaMToD   & 25\,m              & 47.5\,m  \\
SVOX          & 25\,m              & 29.1\,m  \\
\bottomrule
\end{tabular}
\endgroup

\vspace{-0.5mm}
\end{table}
 \paragraph{Concept Erasure.}
We fit transforms to global descriptors, assigning one condition
label to queries/distractors and another to database images.
Following the evaluation convention of linear concept erasure
methods~\cite{ravfogel2020null,belrose2023leace}, we evaluate
all three methods inductively by fitting on a small subset
of the data and evaluating on the held-out set.
We run INLP for 35 iterations, following the main experiment
of Ravfogel~et~al.~\cite{ravfogel2020null}.
We exclude SelaVPR~\cite{lu2024towards} and SelaVPR++,
whose quantized hashing and local
re-ranking fall outside global descriptor transforms.
We use a spatially contiguous region around the map center
as the fitting split (supplementary material, Sec.~S4), except in
\Cref{fig:nordland_inlp_grid}, where INLP is fit on two
database traversals where they overlap.

\subsection{DAR Distinguishes Methods with Similar Recall}
\label{subsec:main_results}
In \Cref{tab:recall_combined}, methods with higher \ac{R1}
generally also achieve higher \ac{DAR}@1 and \ac{DARx}@1,
suggesting that stronger standard recall tends to accompany
greater condition blindness. However,
both \ac{DAR}@1 and \ac{DARx}@1 can reverse method
rankings relative to \ac{R1}, as seen with CricaVPR and
EigenPlaces on SVOX (\ac{R1}: 97.0\% vs.\ 91.6\%;
\ac{DAR}@1: 56.2\% vs.\ 77.1\%), and EigenPlaces and
MixVPR on RobotCar (\ac{R1}: 91.5\% vs.\ 89.9\%;
\ac{DARx}@1: 60.0\% vs.\ 65.1\%). Moreover, similar
\ac{R1} can conceal large differences in
\ac{DAR}@1 or \ac{DARx}@1, as seen on Nordland between
SALAD-CM and SelaVPR++ (\ac{R1}: 96.0\% for both;
\ac{DAR}@1: 87.7\% vs.\ 70.4\%) and between SALAD
and CricaVPR (\ac{R1}: 84.1\% vs.\ 83.4\%;
\ac{DARx}@1: 43.3\% vs.\ 28.1\%). These findings support
using \ac{DAR} alongside standard recall to reveal
differences in condition blindness.

\begin{figure}[t!]
\centering
\begingroup
\definecolor{gOne}   {HTML}{E34948}
\definecolor{gTwo}   {HTML}{1BAF7A}
\definecolor{gThreeA}{HTML}{2A78D6}
\definecolor{gThreeB}{HTML}{EDA100}
\begin{tikzpicture}[font=\scriptsize,
  box/.style={rounded corners=2pt, align=center, inner sep=2.5pt,
              minimum height=1.05cm},
  q/.style={box, draw=gray, fill=gray!8, line width=0.4pt,
            text width=2.7cm},
  leaf/.style={box, draw=#1, fill=#1!12, line width=0.6pt,
               text width=2.4cm},
  ed/.style={-{Stealth[length=3pt]}, gray, line width=0.5pt},
  yn/.style={font=\tiny, text=gray, inner sep=1.5pt}]
  \node[q] (q1) at (1.4,0) {Does excluding nearby\\distractors close the gap\\on Gardens Point?};
  \node[q, below=0.25cm of q1] (q2) {Does it close the gap\\on RobotCar?};
  \node[q, below=0.25cm of q2] (q3) {Is the remaining gap\\on Nordland small?};
  \node[leaf=gOne, left=1cm of q1] (l1)
    {(1) AnyLoc, CosPlace,\\EigenPlaces, MixVPR};
  \node[leaf=gTwo] (l2) at (l1.center |- q2.center)
    {(2) CricaVPR, SelaVPR};
  \node[leaf=gThreeA] (l3a) at (l1.center |- q3.center) {(3) BoQ, SALAD};
  \node[leaf=gThreeB, below=0.25cm of l3a] (l3b) {(4) MegaLoc, SALAD-CM,\\SelaVPR++};
  \draw[ed] (q1.west) -- node[yn, above] {no} (l1.east);
  \draw[ed] (q1.south) -- node[yn, right] {yes} (q2.north);
  \draw[ed] (q2.west) -- node[yn, above] {no} (l2.east);
  \draw[ed] (q2.south) -- node[yn, right] {yes} (q3.north);
  \draw[ed] (q3.west) -- node[yn, above] {no} (l3a.east);
  \draw[ed] (q3.south) |- node[yn, pos=0.25, right] {yes} (l3b.east);
\end{tikzpicture}
\endgroup
\caption{Grouping of methods by the remaining gap after excluding nearby distractors
(\ac{R1} $-$ \acs{DARx}@1). Condition blindness increases from top to bottom.}
\label{fig:dar_group_tree}
\vspace{-3mm}
\end{figure}
 \subsection{Excluding Nearby Distractors Reveals Four Failure Profiles}
\label{subsec:dar_decomposition}

Nearby distractors can be particularly effective and
excluding them can close the performance gap between
\ac{R1} and \ac{DAR}@1 in some cases (\Cref{fig:dar_decomposition}).
Based on their behavior on Gardens Point, RobotCar, and
Nordland, we divide the methods into four groups, ordered
by increasing condition blindness
(\Cref{fig:dar_group_tree}).
\begin{enumerate}[label=(\arabic*)]
\item \textit{AnyLoc, CosPlace, EigenPlaces, MixVPR:}
They keep a large gap after exclusion on RobotCar
and Gardens Point. Their low \ac{DAR}@1 stems from the
condition gap rather than nearby distractors on both datasets.
\item \textit{CricaVPR, SelaVPR:}
Exclusion closes the gap on Gardens Point but leaves
a small gap on RobotCar and a large gap on Nordland.
\item \textit{BoQ, SALAD:}
Exclusion largely closes the gap on RobotCar, Gardens Point,
and SVOX, but leaves a large residual gap on Nordland.
\item \textit{MegaLoc, SALAD-CM, SelaVPR++:}
Exclusion likewise closes most of the gap on RobotCar,
Gardens Point, and SVOX. On Nordland, these methods achieve
the highest \ac{R1} among the four groups, but a moderate
residual gap remains after exclusion.
\end{enumerate}
SVOX and \ac{StLuciaMToD} leave only a small gap after exclusion
for all methods except AnyLoc. Their gap decompositions
are provided in the supplementary material (Sec.~S1).

\begin{figure}[t!]
\centering
\begingroup
\definecolor{mAnyLoc}     {HTML}{1f77b4}
\definecolor{mBoQ}        {HTML}{ff7f0e}
\definecolor{mCosPlace}   {HTML}{2ca02c}
\definecolor{mCricaVPR}   {HTML}{d62728}
\definecolor{mEigenPlaces}{HTML}{9467bd}
\definecolor{mMegaLoc}    {HTML}{8c564b}
\definecolor{mMixVPR}     {HTML}{e377c2}
\definecolor{mSALAD}      {HTML}{7f7f7f}
\definecolor{mSALADCM}    {HTML}{bcbd22}
\definecolor{mSelaVPR}    {HTML}{17becf}
\definecolor{mSelaVPRpp}  {HTML}{000000}
\begin{tikzpicture}
\begin{groupplot}[
  group style={group size=2 by 1, horizontal sep=3mm},
  width=0.6\columnwidth, height=0.55\columnwidth,
  ymin=0, ymax=100,
  ytick={0,25,50,75,100},
  xtick={0,1,2,3,4},
  xticklabels={%
    {\strut sun,clouds\\\strut(query)},
    {\strut overcast},
    {\strut rain},
    {\strut night},
    {\strut none}},
  xticklabel style={align=center, font=\tiny, anchor=north,
                    inner ysep=2pt},
  tick label style={font=\scriptsize},
  label style={font=\scriptsize},
  title style={font=\small, yshift=-1ex},
  grid=both,
  major grid style={line width=.1pt,draw=gray!25},
  minor grid style={line width=.1pt,draw=gray!10},
  axis on top,
  enlarge x limits=0.08,
]
\nextgroupplot[title={(a) DAR@1\strut}, ylabel={DAR@1 (\%)},
  legend to name=leg:distractors_per_distractor_line,
  legend columns=4, legend cell align=left,
  legend style={font=\scriptsize, draw=none, fill=none,
                /tikz/every even column/.append style={column sep=5pt}}]
\addplot[mAnyLoc, mark=*, mark size=1.6pt] coordinates {(0,4.21) (1,12.78) (2,21.76) (3,94.54) (4,95.92)};
\addlegendentry{AnyLoc}
\addplot[mBoQ, mark=square*, mark size=1.6pt] coordinates {(0,85.33) (1,94.64) (2,96.44) (3,99.35) (4,99.45)};
\addlegendentry{BoQ}
\addplot[mCosPlace, mark=triangle*, mark size=2pt] coordinates {(0,38.1) (1,44.35) (2,46.25) (3,91) (4,91.46)};
\addlegendentry{CosPlace}
\addplot[mCricaVPR, mark=diamond*, mark size=2pt] coordinates {(0,51.36) (1,73.34) (2,75.15) (3,98.53) (4,98.76)};
\addlegendentry{CricaVPR}
\addplot[mEigenPlaces, mark=pentagon*, mark size=2pt] coordinates {(0,37.01) (1,42.91) (2,45.82) (3,91.08) (4,91.51)};
\addlegendentry{EigenPlaces}
\addplot[mMegaLoc, mark=*, mark size=1.6pt] coordinates {(0,79.71) (1,88.58) (2,90.16) (3,98.41) (4,98.81)};
\addlegendentry{MegaLoc}
\addplot[mMixVPR, mark=square*, mark size=1.6pt] coordinates {(0,32.47) (1,46.85) (2,50.68) (3,88.78) (4,89.94)};
\addlegendentry{MixVPR}
\addplot[mSALAD, mark=triangle*, mark size=2pt] coordinates {(0,83.11) (1,92.98) (2,94.17) (3,99.01) (4,99.2)};
\addlegendentry{SALAD}
\addplot[mSALADCM, mark=diamond*, mark size=2pt] coordinates {(0,93.6) (1,97.38) (2,97.33) (3,99.15) (4,99.35)};
\addlegendentry{SALAD-CM}
\addplot[mSelaVPR, mark=pentagon*, mark size=2pt] coordinates {(0,71.27) (1,83.81) (2,86.75) (3,97.36) (4,98.18)};
\addlegendentry{SelaVPR}
\addplot[mSelaVPRpp, mark=*, mark size=1.6pt] coordinates {(0,91.95) (1,97.19) (2,97.53) (3,99.18) (4,99.4)};
\addlegendentry{SelaVPR++}
\nextgroupplot[title={(b) \acs{DARx}@1\strut}, yticklabels={}]
\addplot[mAnyLoc, mark=*, mark size=1.6pt] coordinates {(0,20.23) (1,36.08) (2,46.3) (3,95.27) (4,95.92)};
\addplot[mBoQ, mark=square*, mark size=1.6pt] coordinates {(0,99.38) (1,99.43) (2,99.45) (3,99.43) (4,99.45)};
\addplot[mCosPlace, mark=triangle*, mark size=2pt] coordinates {(0,60.68) (1,62.57) (2,63.2) (3,91.26) (4,91.46)};
\addplot[mCricaVPR, mark=diamond*, mark size=2pt] coordinates {(0,94.37) (1,96.07) (2,95.44) (3,98.66) (4,98.76)};
\addplot[mEigenPlaces, mark=pentagon*, mark size=2pt] coordinates {(0,59.98) (1,60.16) (2,63.04) (3,91.16) (4,91.51)};
\addplot[mMegaLoc, mark=*, mark size=1.6pt] coordinates {(0,98.3) (1,98.53) (2,98.46) (3,98.71) (4,98.81)};
\addplot[mMixVPR, mark=square*, mark size=1.6pt] coordinates {(0,65.07) (1,70.74) (2,72.91) (3,89.16) (4,89.94)};
\addplot[mSALAD, mark=triangle*, mark size=2pt] coordinates {(0,99.15) (1,99.06) (2,99.15) (3,99.15) (4,99.2)};
\addplot[mSALADCM, mark=diamond*, mark size=2pt] coordinates {(0,99.32) (1,99.32) (2,99.21) (3,99.3) (4,99.35)};
\addplot[mSelaVPR, mark=pentagon*, mark size=2pt] coordinates {(0,94.74) (1,96.83) (2,96.81) (3,97.93) (4,98.18)};
\addplot[mSelaVPRpp, mark=*, mark size=1.6pt] coordinates {(0,99.37) (1,99.35) (2,99.32) (3,99.38) (4,99.4)};
\end{groupplot}
\node[anchor=north, font=\scriptsize] at ($(group c1r1.south east)!0.5!(group c2r1.south west)+(0,-0.55cm)$) {Distractor set};
\end{tikzpicture}\\[-1mm]
\pgfplotslegendfromname{leg:distractors_per_distractor_line}
\endgroup
\caption{\textbf{(a)} \ac{DAR}@1 and \textbf{(b)} \acs{DARx}@1 with
RobotCar distractor sets from different conditions,
ordered by mean \ac{DAR}@1. In (b), distractors within 100\,m
of each query are removed. \texttt{sun,clouds} is the query set
(main results); \texttt{none} gives \ac{R1} without distractors.}
\label{fig:distractors_per_distractor_line}

\vspace{-4mm}
\end{figure}
 \subsection{DAR Allows Distractor Condition Ablation}
\label{subsec:ablation}

\ac{DAR} lets us isolate the effect of distractor conditions
by varying the distractor set while holding the query and
database fixed. On RobotCar (\Cref{fig:distractors_per_distractor_line}),
the groups from \Cref{subsec:dar_decomposition} show three
patterns after excluding nearby distractors. Daytime distractors
still cause large losses for AnyLoc, CosPlace, EigenPlaces,
and MixVPR (group~1), modest losses for CricaVPR and SelaVPR
(group~2), and almost none for the remaining methods
(groups~3 and~4). Nighttime distractors
have little effect both before and after exclusion.
\begin{figure}[t!]
\centering
\begingroup
\pgfplotsset{
  colormap={gainloss}{
    rgb255(0cm)=(178,24,43);
    rgb255(0.5cm)=(247,247,247);
    rgb255(1cm)=(0,104,55);
  },
  lcheat/.style={
    scale only axis, width=51pt, height=100pt,
    colormap name=gainloss,
    xmin=-0.5, xmax=5.5, ymin=-0.5, ymax=8.5,
    enlargelimits=false, axis on top,
    xtick={0,1,2,3,4,5},
    xticklabels={Nordland,RobotCar,AmsterTime,SVOX,StLucia,Gardens Point},
    x tick label style={font=\fontsize{6}{7}\selectfont, rotate=45, anchor=north east, inner sep=1pt},
    ytick={8,7,6,5,4,3,2,1,0},
    yticklabels={AnyLoc,EigenPlaces,CricaVPR,CosPlace,MixVPR,SALAD,MegaLoc,BoQ,SALAD-CM},
    yticklabel pos=left,
    y tick label style={font=\scriptsize},
    tick style={draw=none},
    title style={font=\small, yshift=-0.5ex},
    every node near coord/.append style={font=\fontsize{5.2}{5.6}\selectfont, /pgf/number format/assume math mode=true},
  },
  scorescale/.style={ point meta min=-1, point meta max=1 },
  darscale/.style={ point meta min=-1, point meta max=1 },
  lcmatrix/.style={
    matrix plot*, mesh/cols=6, point meta=explicit, table/col sep=space,
    visualization depends on=\thisrow{lbl}\as\mylabel,
    nodes near coords={%
      \pgfmathparse{abs(\mylabel)>=9.95?0:1}\edef\myprec{\pgfmathresult}%
      \pgfmathparse{\myprec==0?round(\mylabel):round(\mylabel*10)/10}\edef\myround{\pgfmathresult}%
      \ifdim\myround pt<-1.5pt\color{white}\fi
      \pgfmathparse{abs(\myround)}\edef\mymag{\pgfmathresult}%
      \shortstack[c]{%
        {\ifdim\myround pt<0pt$-$\else\ifdim\myround pt>0pt$+$\else\vphantom{$+$}\fi\fi}\\[-1.1ex]%
        \pgfmathprintnumber[fixed,fixed zerofill,precision=\myprec]{\mymag}%
      }%
      },
    every node near coord/.append style={anchor=center},
  },
}
\begin{tikzpicture}
\begin{axis}[lcheat, darscale, name=s00, title={INLP\strut}, xticklabels={}, ylabel near ticks, ylabel={$\Delta$DAR@1 (pp)}, ylabel style={font=\small}]
\addplot[lcmatrix] table[meta=c] {
x y c lbl
0 8 0.4150 5.24
1 8 0.8225 36.66
2 8 0.5980 12.99
3 8 0.7645 28.16
4 8 0.6475 16.40
5 8 0.2651 2.22
0 7 0.6140 14.01
1 7 0.6736 18.53
2 7 0.3014 2.78
3 7 0.2787 2.42
4 7 -0.0357 -0.14
5 7 0.2651 2.22
0 6 0.7286 23.89
1 6 0.4417 6.02
2 6 0.4490 6.25
3 6 0.4095 5.09
4 6 0.0779 0.41
5 6 -0.1211 -0.56
0 5 0.6165 14.18
1 5 0.6272 14.91
2 5 0.2692 2.28
3 5 0.2699 2.29
4 5 0.0153 0.07
5 5 0.2651 2.22
0 4 0.5362 9.65
1 4 0.5047 8.27
2 4 0.3588 3.87
3 4 0.3901 4.59
4 4 0.0153 0.07
5 4 0.3014 2.78
0 3 0.6544 16.94
1 3 0.2301 1.76
2 3 0.3634 3.97
3 3 0.1594 1.02
4 3 0.0153 0.07
5 3 0.0000 0.00
0 2 0.4713 7.00
1 2 0.1994 1.41
2 2 0.2692 2.28
3 2 0.0934 0.51
4 2 -0.0184 -0.07
5 2 0.2218 1.66
0 1 0.5631 10.99
1 1 0.0730 0.38
2 1 0.0216 0.10
3 1 0.1594 1.02
4 1 0.0000 0.00
5 1 0.0000 0.00
0 0 0.2794 2.43
1 0 0.0317 0.15
2 0 0.1777 1.19
3 0 0.0934 0.51
4 0 0.0000 0.00
5 0 0.1008 0.56
};
\end{axis}
\begin{axis}[lcheat, darscale, name=s01, at={(s00.east)}, anchor=west, xshift=3pt, yticklabels={}, ytick=\empty, title={LEACE\strut}, xticklabels={}]
\addplot[lcmatrix] table[meta=c] {
x y c lbl
0 8 0.3394 3.47
1 8 0.8158 35.56
2 8 0.5577 10.71
3 8 0.7303 24.08
4 8 0.5171 8.79
5 8 0.3322 3.33
0 7 0.6156 14.12
1 7 0.6889 19.89
2 7 0.2475 1.98
3 7 0.2104 1.53
4 7 0.0153 0.07
5 7 0.2226 1.67
0 6 0.7105 21.98
1 6 0.4132 5.19
2 6 0.4154 5.25
3 6 0.3949 4.71
4 6 0.0542 0.27
5 6 0.0000 0.00
0 5 0.6058 13.48
1 5 0.5889 12.44
2 5 0.2243 1.69
3 5 0.1107 0.63
4 5 0.0153 0.07
5 5 0.1692 1.11
0 4 0.5221 9.01
1 4 0.5028 8.19
2 4 0.3296 3.28
3 4 0.3503 3.69
4 4 0.0153 0.07
5 4 0.3014 2.78
0 3 0.6373 15.64
1 3 0.2113 1.54
2 3 0.3125 2.97
3 3 0.1294 0.77
4 3 0.0000 0.00
5 3 0.0993 0.55
0 2 0.4725 7.04
1 2 0.1682 1.10
2 2 0.2398 1.88
3 2 0.0277 0.13
4 2 0.0000 0.00
5 2 0.2218 1.66
0 1 0.5373 9.70
1 1 0.0629 0.32
2 1 0.0746 0.39
3 1 0.1443 0.89
4 1 0.0000 0.00
5 1 0.0000 0.00
0 0 0.2727 2.33
1 0 0.0317 0.15
2 0 0.0763 0.40
3 0 0.0730 0.38
4 0 0.0000 0.00
5 0 0.0000 0.00
};
\end{axis}
\begin{axis}[lcheat, darscale, name=s02, at={(s01.east)}, anchor=west, xshift=3pt, title={STD\strut}, xticklabels={}, yticklabels={}, ytick=\empty]
\addplot[lcmatrix] table[meta=c] {
x y c lbl
0 8 0.1937 1.35
1 8 0.6193 14.37
2 8 0.5538 10.51
3 8 0.6413 15.93
4 8 -0.6035 -8.17
5 8 -0.6530 -10.00
0 7 0.5779 11.80
1 7 0.4919 7.76
2 7 0.2622 2.18
3 7 -0.6140 -8.53
4 7 -0.1314 -0.62
5 7 -0.6663 -10.55
0 6 0.7054 21.47
1 6 -0.5710 -7.14
2 6 0.4427 6.05
3 6 -0.5834 -7.52
4 6 -0.3168 -2.20
5 6 -0.7551 -15.00
0 5 0.5741 11.59
1 5 0.0375 0.18
2 5 0.1560 0.99
3 5 -0.6142 -8.54
4 5 -0.0519 -0.21
5 5 -0.6665 -10.56
0 4 0.5242 9.10
1 4 -0.3396 -2.48
2 4 0.2557 2.09
3 4 -0.4978 -5.22
4 4 -0.1631 -0.82
5 4 -0.5913 -7.77
0 3 0.5417 9.91
1 3 -0.5432 -6.35
2 3 0.3394 3.47
3 3 -0.5789 -7.38
4 3 -0.0672 -0.28
5 3 -0.7733 -16.11
0 2 0.4696 6.94
1 2 -0.5357 -6.15
2 2 0.2157 1.59
3 2 -0.5663 -7.00
4 2 -0.1646 -0.83
5 2 -0.5548 -6.67
0 1 0.4679 6.88
1 1 -0.5376 -6.20
2 1 -0.3762 -2.98
3 1 -0.5487 -6.50
4 1 -0.0184 -0.07
5 1 -0.7985 -17.77
0 0 0.2699 2.29
1 0 -0.3843 -3.10
2 0 -0.1429 -0.69
3 0 -0.5034 -5.35
4 0 0.0000 0.00
5 0 -0.7454 -14.44
};
\end{axis}
\begin{axis}[lcheat, darscale, name=s10, at={(s00.south)}, anchor=north, yshift=-6pt, xticklabels={}, ylabel near ticks, ylabel={$\Delta$\acs{DARx}@1 (pp)}, ylabel style={font=\small}]
\addplot[lcmatrix] table[meta=c] {
x y c lbl
0 8 0.5454 10.09
1 8 0.9623 68.80
2 8 0.5980 12.99
3 8 0.9794 74.26
4 8 0.6493 16.54
5 8 0.4061 5.00
0 7 0.6724 18.42
1 7 0.7106 21.99
2 7 0.3014 2.78
3 7 0.1121 0.64
4 7 -0.0357 -0.14
5 7 0.2651 2.22
0 6 0.8153 35.49
1 6 0.1917 1.33
2 6 0.4490 6.25
3 6 0.0934 0.51
4 6 0.0277 0.13
5 6 0.1008 0.56
0 5 0.6699 18.21
1 5 0.6786 18.96
2 5 0.2953 2.68
3 5 0.1121 0.64
4 5 0.0000 0.00
5 5 0.3598 3.89
0 4 0.6230 14.62
1 4 0.5571 10.68
2 4 0.3588 3.87
3 4 0.1121 0.64
4 4 0.0153 0.07
5 4 0.3322 3.33
0 3 0.7282 23.85
1 3 0.0000 0.00
2 3 0.3634 3.97
3 3 -0.0333 -0.13
4 3 0.0000 0.00
5 3 -0.1194 -0.55
0 2 0.4636 6.73
1 2 0.0153 0.07
2 2 0.2692 2.28
3 2 0.0277 0.13
4 2 0.0000 0.00
5 2 0.0000 0.00
0 1 0.6237 14.67
1 1 -0.0054 -0.02
2 1 0.0216 0.10
3 1 0.0000 0.00
4 1 0.0000 0.00
5 1 0.0000 0.00
0 0 0.2475 1.98
1 0 0.0000 0.00
2 0 0.1777 1.19
3 0 -0.0333 -0.13
4 0 0.0000 0.00
5 0 -0.1211 -0.56
};
\end{axis}
\begin{axis}[lcheat, darscale, name=s11, at={(s10.east)}, anchor=west, xshift=3pt, yticklabels={}, ytick=\empty, xticklabels={}]
\addplot[lcmatrix] table[meta=c] {
x y c lbl
0 8 0.4758 7.16
1 8 0.9588 67.71
2 8 0.5577 10.71
3 8 0.9660 69.93
4 8 0.5563 10.64
5 8 0.4061 5.00
0 7 0.6757 18.71
1 7 0.7123 22.16
2 7 0.2475 1.98
3 7 -0.0877 -0.38
4 7 0.0153 0.07
5 7 0.2651 2.22
0 6 0.8001 33.12
1 6 0.1571 1.00
2 6 0.4154 5.25
3 6 0.0730 0.38
4 6 0.0000 0.00
5 6 0.0000 0.00
0 5 0.6611 17.48
1 5 0.6567 17.12
2 5 0.2243 1.69
3 5 -0.2072 -1.14
4 5 0.0000 0.00
5 5 0.3014 2.78
0 4 0.6097 13.73
1 4 0.5634 11.01
2 4 0.3296 3.28
3 4 0.0746 0.39
4 4 0.0000 0.00
5 4 0.4446 6.11
0 3 0.7202 22.98
1 3 0.0045 0.02
2 3 0.3125 2.97
3 3 0.0277 0.13
4 3 0.0000 0.00
5 3 0.0000 0.00
0 2 0.4696 6.94
1 2 0.0195 0.09
2 2 0.2398 1.88
3 2 0.0000 0.00
4 2 0.0000 0.00
5 2 0.0000 0.00
0 1 0.6106 13.79
1 1 0.0000 0.00
2 1 0.0746 0.39
3 1 0.0000 0.00
4 1 0.0000 0.00
5 1 0.0000 0.00
0 0 0.2390 1.87
1 0 0.0000 0.00
2 0 0.0763 0.40
3 0 -0.0333 -0.13
4 0 0.0000 0.00
5 0 0.0000 0.00
};
\end{axis}
\begin{axis}[lcheat, darscale, name=s12, at={(s11.east)}, anchor=west, xshift=3pt, xticklabels={}, yticklabels={}, ytick=\empty]
\addplot[lcmatrix] table[meta=c] {
x y c lbl
0 8 0.3621 3.94
1 8 0.8841 48.43
2 8 0.5538 10.51
3 8 0.9406 62.42
4 8 -0.6987 -12.01
5 8 -0.8498 -21.66
0 7 0.6636 17.68
1 7 0.5979 12.98
2 7 0.2622 2.18
3 7 -0.4557 -4.33
4 7 -0.0651 -0.27
5 7 -0.8630 -22.78
0 6 0.8172 35.79
1 6 -0.2965 -1.97
2 6 0.4427 6.05
3 6 -0.2085 -1.15
4 6 -0.0936 -0.41
5 6 -0.5342 -6.11
0 5 0.6533 16.85
1 5 0.4802 7.32
2 5 0.1560 0.99
3 5 -0.3979 -3.31
4 5 -0.0184 -0.07
5 5 -0.6388 -9.44
0 4 0.6331 15.33
1 4 -0.4220 -3.71
2 4 0.2557 2.09
3 4 -0.2784 -1.78
4 4 -0.0357 -0.14
5 4 -1.0000 -38.33
0 3 0.7018 21.11
1 3 -0.0451 -0.18
2 3 0.3394 3.47
3 3 -0.2085 -1.15
4 3 -0.0184 -0.07
5 3 -0.6241 -8.89
0 2 0.4739 7.09
1 2 -0.1734 -0.89
2 2 0.2157 1.59
3 2 -0.2233 -1.27
4 2 -0.0184 -0.07
5 2 -0.9097 -27.23
0 1 0.6041 13.37
1 1 -0.1031 -0.46
2 1 -0.3762 -2.98
3 1 -0.1915 -1.02
4 1 0.0000 0.00
5 1 -0.5342 -6.11
0 0 0.2475 1.98
1 0 -0.0284 -0.11
2 0 -0.1429 -0.69
3 0 -0.1915 -1.02
4 0 0.0153 0.07
5 0 -0.7144 -12.78
};
\end{axis}
\begin{axis}[lcheat, darscale, name=s20, at={(s10.south)}, anchor=north, yshift=-6pt, ylabel near ticks, ylabel={$\Delta$R@1 (pp)}, ylabel style={font=\small}]
\addplot[lcmatrix] table[meta=c] {
x y c lbl
0 8 0.6742 18.58
1 8 0.1382 0.84
2 8 0.5269 9.22
3 8 0.0934 0.51
4 8 0.6043 13.38
5 8 0.2226 1.67
0 7 0.5933 12.70
1 7 0.1490 0.93
2 7 0.2148 1.58
3 7 0.0506 0.25
4 7 -0.0184 -0.07
5 7 0.2226 1.67
0 6 0.4835 7.44
1 6 -0.0107 -0.04
2 6 0.2068 1.49
3 6 0.0000 0.00
4 6 0.0000 0.00
5 6 0.0993 0.55
0 5 0.6544 16.94
1 5 0.1107 0.63
2 5 0.1571 1.00
3 5 0.0730 0.38
4 5 0.0000 0.00
5 5 0.3014 2.78
0 4 0.5714 11.44
1 4 0.1008 0.56
2 4 0.3634 3.97
3 4 0.0000 0.00
4 4 0.0153 0.07
5 4 -0.2033 -1.11
0 3 0.4274 5.59
1 3 -0.0054 -0.02
2 3 0.2318 1.78
3 3 -0.0608 -0.25
4 3 0.0000 0.00
5 3 -0.1194 -0.55
0 2 0.1307 0.78
1 2 0.0045 0.02
2 2 0.2243 1.69
3 2 0.0277 0.13
4 2 0.0000 0.00
5 2 0.0000 0.00
0 1 0.3625 3.95
1 1 0.0045 0.02
2 1 0.0595 0.30
3 1 0.0000 0.00
4 1 0.0000 0.00
5 1 0.0000 0.00
0 0 0.0469 0.23
1 0 0.0045 0.02
2 0 0.2622 2.18
3 0 -0.0333 -0.13
4 0 0.0000 0.00
5 0 -0.1211 -0.56
};
\end{axis}
\begin{axis}[lcheat, darscale, name=s21, at={(s20.east)}, anchor=west, xshift=3pt, yticklabels={}, ytick=\empty]
\addplot[lcmatrix] table[meta=c] {
x y c lbl
0 8 0.6603 17.41
1 8 0.1332 0.80
2 8 0.4609 6.64
3 8 -0.0629 -0.26
4 8 0.5201 8.92
5 8 -0.1194 -0.55
0 7 0.6205 14.45
1 7 0.1756 1.17
2 7 0.1777 1.19
3 7 -0.0877 -0.38
4 7 0.0000 0.00
5 7 0.1008 0.56
0 6 0.4481 6.22
1 6 -0.0159 -0.06
2 6 0.0216 0.10
3 6 0.0277 0.13
4 6 -0.0184 -0.07
5 6 0.0000 0.00
0 5 0.6600 17.39
1 5 0.0919 0.50
2 5 0.1332 0.80
3 5 -0.1734 -0.89
4 5 -0.0184 -0.07
5 5 0.3014 2.78
0 4 0.5683 11.27
1 4 0.1079 0.61
2 4 0.3493 3.67
3 4 0.0000 0.00
4 4 0.0000 0.00
5 4 0.2226 1.67
0 3 0.4301 5.67
1 3 -0.0054 -0.02
2 3 0.1443 0.89
3 3 0.0000 0.00
4 3 0.0000 0.00
5 3 0.0000 0.00
0 2 0.1382 0.84
1 2 0.0067 0.03
2 2 0.1571 1.00
3 2 0.0000 0.00
4 2 0.0000 0.00
5 2 0.0000 0.00
0 1 0.3513 3.71
1 1 0.0045 0.02
2 1 0.0000 0.00
3 1 0.0000 0.00
4 1 0.0000 0.00
5 1 0.0000 0.00
0 0 0.0394 0.19
1 0 0.0000 0.00
2 0 0.1189 0.69
3 0 -0.0333 -0.13
4 0 0.0000 0.00
5 0 0.0000 0.00
};
\end{axis}
\begin{axis}[lcheat, darscale, name=s22, at={(s21.east)}, anchor=west, xshift=3pt, yticklabels={}, ytick=\empty]
\addplot[lcmatrix] table[meta=c] {
x y c lbl
0 8 0.7547 26.93
1 8 0.1514 0.95
2 8 0.5538 10.51
3 8 -0.0629 -0.26
4 8 -0.6055 -8.24
5 8 0.1008 0.56
0 7 0.6747 18.62
1 7 0.2131 1.56
2 7 0.3067 2.87
3 7 -0.1555 -0.77
4 7 0.0413 0.20
5 7 0.1692 1.11
0 6 0.5389 9.78
1 6 -0.0054 -0.02
2 6 0.3394 3.47
3 6 0.0277 0.13
4 6 -0.0184 -0.07
5 6 0.2218 1.66
0 5 0.7024 21.17
1 5 0.1888 1.30
2 5 0.1777 1.19
3 5 0.0730 0.38
4 5 0.0000 0.00
5 5 0.4264 5.56
0 4 0.6633 17.66
1 4 0.1616 1.04
2 4 0.3680 4.07
3 4 -0.0309 -0.12
4 4 0.0153 0.07
5 4 0.1703 1.12
0 3 0.4832 7.43
1 3 0.0000 0.00
2 3 0.2761 2.38
3 3 -0.0877 -0.38
4 3 0.0000 0.00
5 3 -0.1194 -0.55
0 2 0.2059 1.48
1 2 0.0195 0.09
2 2 0.2157 1.59
3 2 0.0277 0.13
4 2 0.0153 0.07
5 2 0.2226 1.67
0 1 0.3806 4.36
1 1 0.0153 0.07
2 1 0.1320 0.79
3 1 -0.0629 -0.26
4 1 0.0000 0.00
5 1 -0.1211 -0.56
0 0 0.0629 0.32
1 0 -0.0080 -0.03
2 0 0.2235 1.68
3 0 -0.1122 -0.51
4 0 0.0000 0.00
5 0 -0.1211 -0.56
};
\end{axis}
\end{tikzpicture}
\endgroup
\vspace{-1mm}

\caption{\ac{DAR}@1, \acs{DARx}@1, and \ac{R1} gains (pp) over the
unerased baseline; higher is better. Transforms use a 10\% fitting
and 90\% held-out split of the query/distractor pool and database.
Rows and columns follow mean \ac{DAR}@1 gain.
Raw values are provided in the supplementary material (Sec.~S4).}
\label{fig:linear_classifier_gain_heatmap}
\vspace{-3mm}
\end{figure}
 \subsection{Comparing INLP, LEACE, and STD for Improving Condition Blindness}
\label{subsec:condition_suppression}

\Cref{fig:linear_classifier_gain_heatmap} compares concept
erasure methods fitted on 10\% of the query region and
evaluated on the held-out 90\%. INLP and LEACE improve \ac{DAR}@1
in more pairs than \ac{STD}. Nevertheless,
\ac{STD} improves both \ac{DAR}@1 and \ac{DARx}@1 on Nordland,
which provides the largest fitting set at the same fitting
fraction. This is consistent with the results in
\Cref{fig:concept_erasure_curve}, which show that \ac{STD}
benefits from additional fitting data and requires more than
INLP and LEACE to improve average \ac{DAR}@1 and \ac{DARx}@1
on unseen places. The transforms improve or preserve
\ac{R1} in most pairs, although some incur losses.
These results suggest that INLP and LEACE generalize better
to unseen places than \ac{STD} when the amount of fitting data
is small, while all three methods can improve condition
blindness with sufficient fitting data.

\begin{figure}[t!]
\centering
\begingroup
\definecolor{mINLP} {HTML}{1f77b4}
\definecolor{mLEACE}{HTML}{d62728}
\definecolor{mSTD}  {HTML}{2ca02c}
\pgfplotsset{
  erasurecurve/.style={
    width=0.43\columnwidth, height=0.5\columnwidth,
    xmin=0, xmax=0.5, xtick={0,0.1,0.2,0.3,0.4,0.5}, xticklabels={0,,0.2,,0.4,},
    tick label style={font=\scriptsize},
    title style={font=\footnotesize, yshift=-1.0ex},
    label style={font=\scriptsize},
    grid=both,
    major grid style={line width=.1pt,draw=gray!25},
    minor grid style={line width=.1pt,draw=gray!10},
    axis on top,
  },
}
\hspace*{-5mm}
\begin{tikzpicture}
\begin{groupplot}[
  group style={group size=3 by 1, horizontal sep=0.25cm},
  erasurecurve,
]
\nextgroupplot[title={$\Delta$DAR@1\strut},
  ymin=-2, ymax=8, ytick={-2,0,2,4,6,8},
  legend columns=3, legend cell align=left,
  legend style={font=\scriptsize, draw=none, fill=none, line width=0pt,
                /tikz/every even column/.append style={column sep=10pt},
                at={(1.62,-0.30)}, anchor=north}]
\draw[dashed, gray!60] (axis cs:0,0) -- (axis cs:0.5,0);
\addplot[mINLP, mark=*, mark size=1.6pt] coordinates {
  (0.00,0.000) (0.10,5.382) (0.20,6.094) (0.30,6.325) (0.40,6.510) (0.50,6.787)};
\addlegendentry{INLP}
\addplot[mLEACE, mark=square*, mark size=1.6pt] coordinates {
  (0.00,0.000) (0.10,4.683) (0.20,5.476) (0.30,5.698) (0.40,5.913) (0.50,5.998)};
\addlegendentry{LEACE}
\addplot[mSTD, mark=triangle*, mark size=2pt] coordinates {
  (0.00,0.000) (0.10,-0.901) (0.20,2.508) (0.30,3.254) (0.40,3.917) (0.50,4.492)};
\addlegendentry{STD}
\nextgroupplot[title={$\Delta$\acs{DARx}@1\strut},
  ymin=-2, ymax=8, ytick={-2,0,2,4,6,8}, yticklabels={}]
\draw[dashed, gray!60] (axis cs:0,0) -- (axis cs:0.5,0);
\addplot[mINLP, mark=*, mark size=1.6pt] coordinates {
  (0.00,0.000) (0.10,6.818) (0.20,7.258) (0.30,7.548) (0.40,7.686) (0.50,7.811)};
\addplot[mLEACE, mark=square*, mark size=1.6pt] coordinates {
  (0.00,0.000) (0.10,6.471) (0.20,6.881) (0.30,6.946) (0.40,7.145) (0.50,7.261)};
\addplot[mSTD, mark=triangle*, mark size=2pt] coordinates {
  (0.00,0.000) (0.10,1.531) (0.20,4.615) (0.30,5.332) (0.40,6.175) (0.50,6.823)};
\nextgroupplot[title={$\Delta$R@1\strut},
  ymin=-2, ymax=8, ytick={-2,0,2,4,6,8}, yticklabels={}]
\draw[dashed, gray!60] (axis cs:0,0) -- (axis cs:0.5,0);
\addplot[mINLP, mark=*, mark size=1.6pt] coordinates {
  (0.00,0.000) (0.10,1.947) (0.20,1.854) (0.30,2.024) (0.40,1.943) (0.50,1.971)};
\addplot[mLEACE, mark=square*, mark size=1.6pt] coordinates {
  (0.00,0.000) (0.10,1.650) (0.20,1.743) (0.30,1.714) (0.40,1.761) (0.50,1.716)};
\addplot[mSTD, mark=triangle*, mark size=2pt] coordinates {
  (0.00,0.000) (0.10,2.037) (0.20,2.274) (0.30,2.298) (0.40,2.359) (0.50,2.405)};
\end{groupplot}
\node[anchor=north, font=\footnotesize] at
  ($(group c2r1.south)+(0,-0.35cm)$)
  {Fitting fraction};
\end{tikzpicture}
\endgroup
\caption{Concept erasure \ac{DAR}@1, \ac{DARx}@1, and
\ac{R1} gains (pp) versus fitting fraction,
averaged over 54 method--dataset pairs with a fixed 50\% evaluation set.
At fraction $0$, no erasure is applied.}
\label{fig:concept_erasure_curve}

\vspace{-3mm}
\end{figure}
 \subsection{Demonstrating INLP in a Multi-Traversal Scenario}

Consider a real-world scenario where a robot builds a database
of a large area from two traversals under different conditions,
with a small spatial overlap, and later queries it during another
traversal. We fit INLP on database images from the overlap and
use it to pre-process the database descriptors, leaving query
descriptors unchanged. We demonstrate this scenario on RobotCar
and Nordland.

\providecommand{\suppressionarrow}{%
  \begin{tikzpicture}[inner sep=0pt, outer sep=0pt]
    \draw[-{Triangle[length=3mm, width=3mm]}, line width=3pt] (0,0) -- (0.5,0);
  \end{tikzpicture}%
}
\begin{figure}[t!]
\centering
\setlength{\tabcolsep}{2pt}
\renewcommand{\arraystretch}{1.0}
\setlength{\fboxsep}{0pt}
\begin{tabular}{@{}m{0.5cm} m{0.4\columnwidth} c m{0.4\columnwidth}@{}}
 & \centering\small Original & & \centering\small Erased \tabularnewline
\centering\rotatebox{90}{\small AnyLoc} &
\rcornerlabel{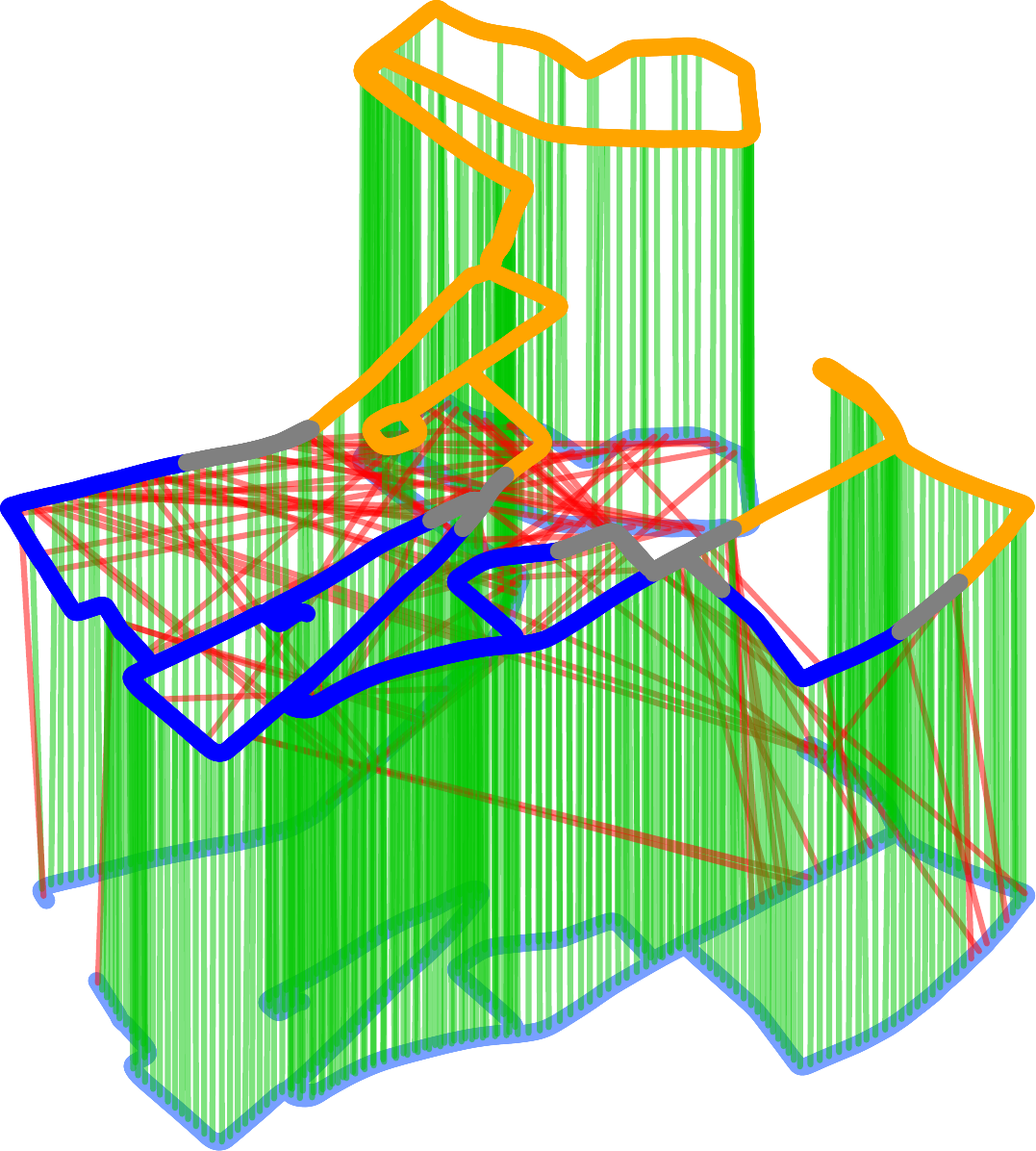}{81.6} &
\suppressionarrow &
\rcornerlabel{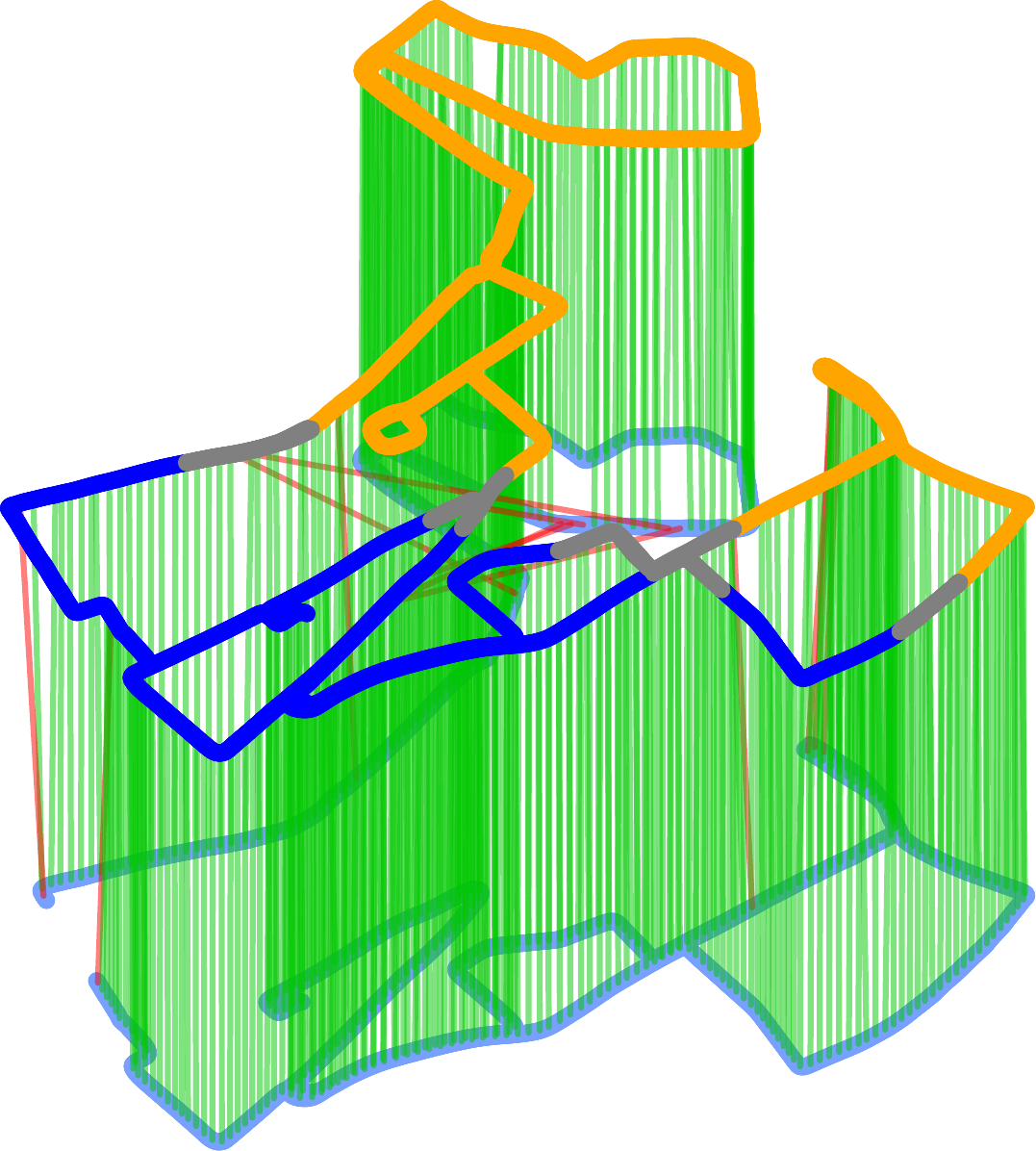}{96.4} \tabularnewline
\centering\rotatebox{90}{\small EigenPlaces} &
\rcornerlabel{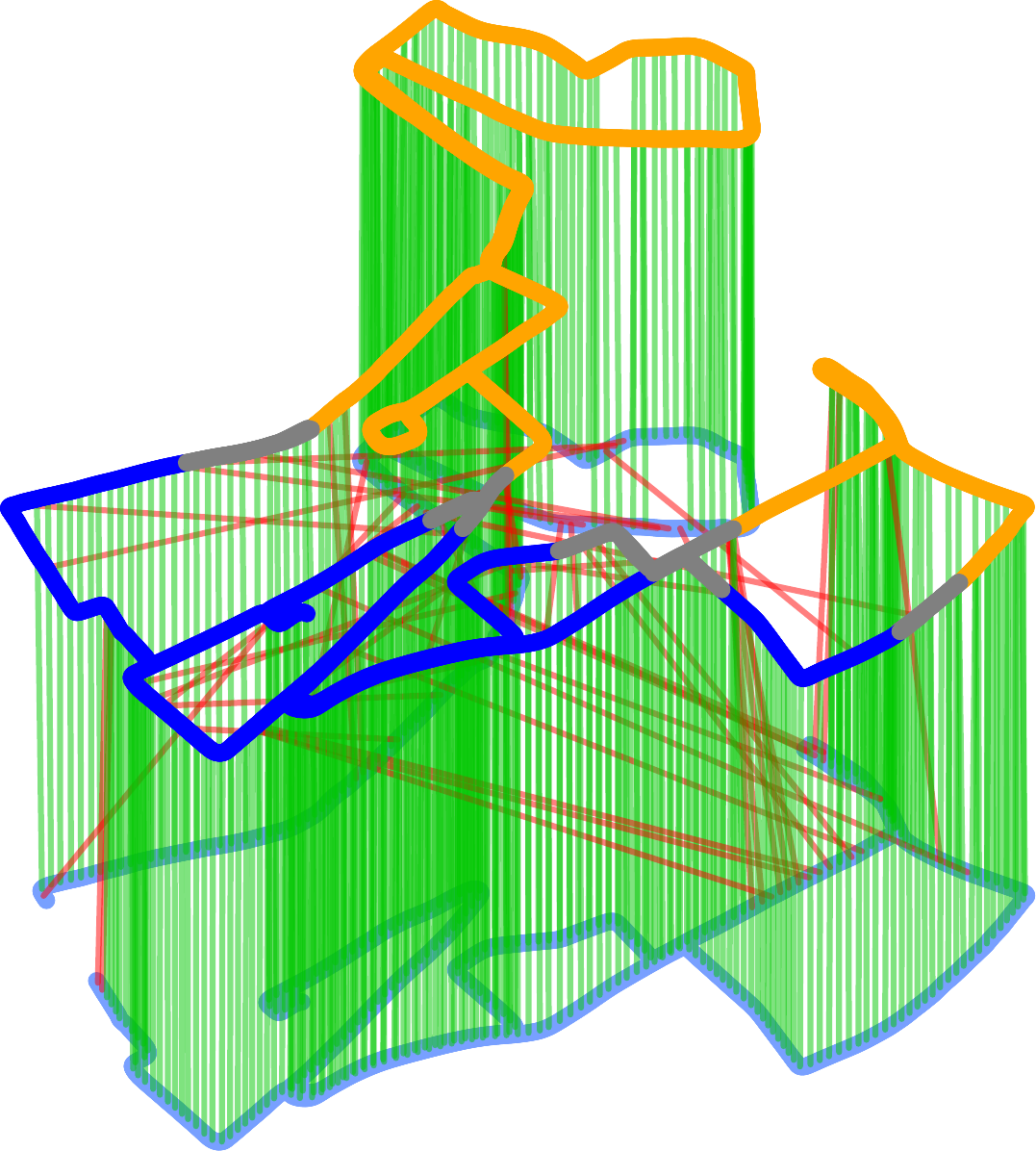}{89.4} &
\suppressionarrow &
\rcornerlabel{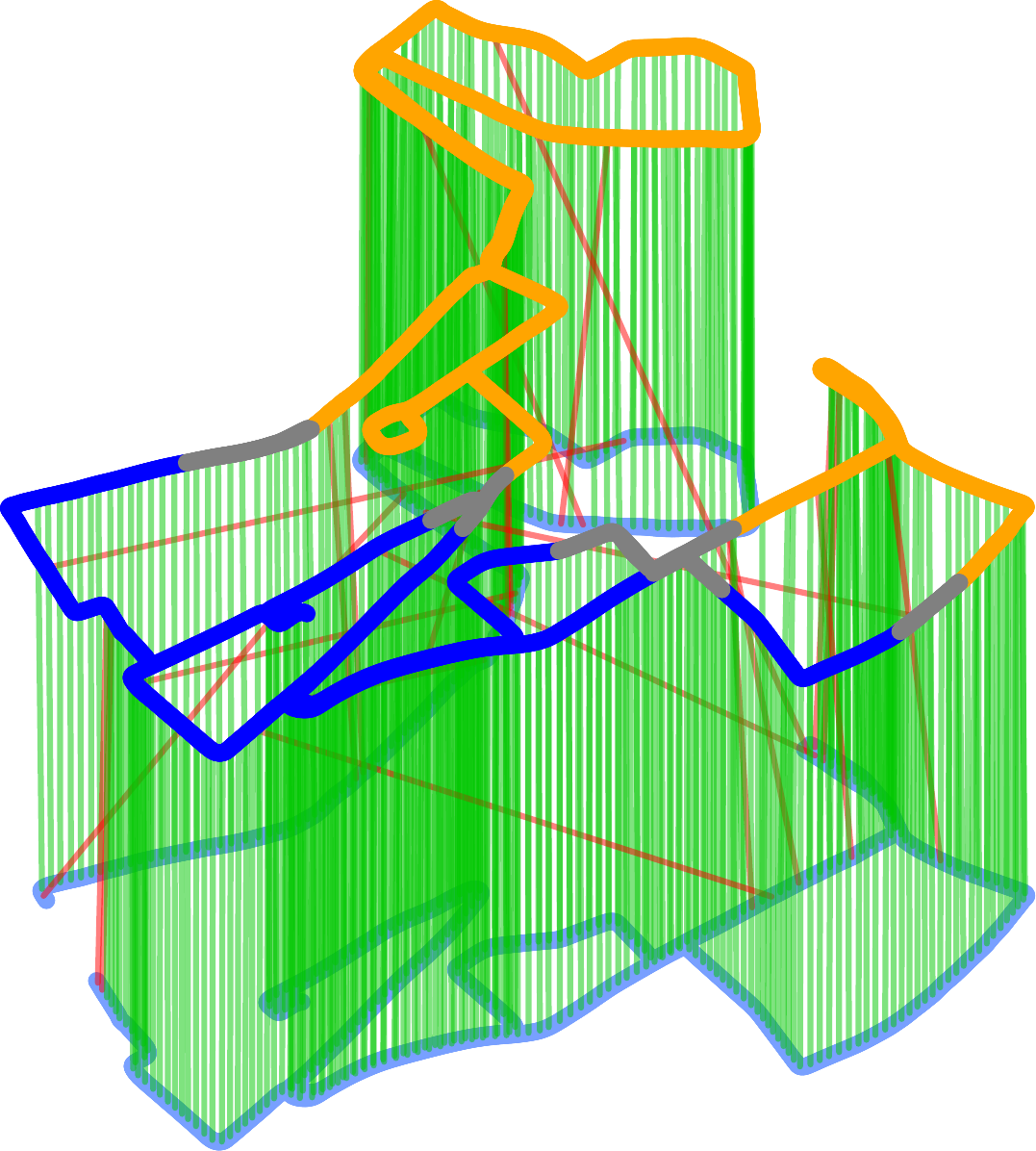}{95.1} \tabularnewline
\centering\rotatebox{90}{\small CosPlace} &
\rcornerlabel{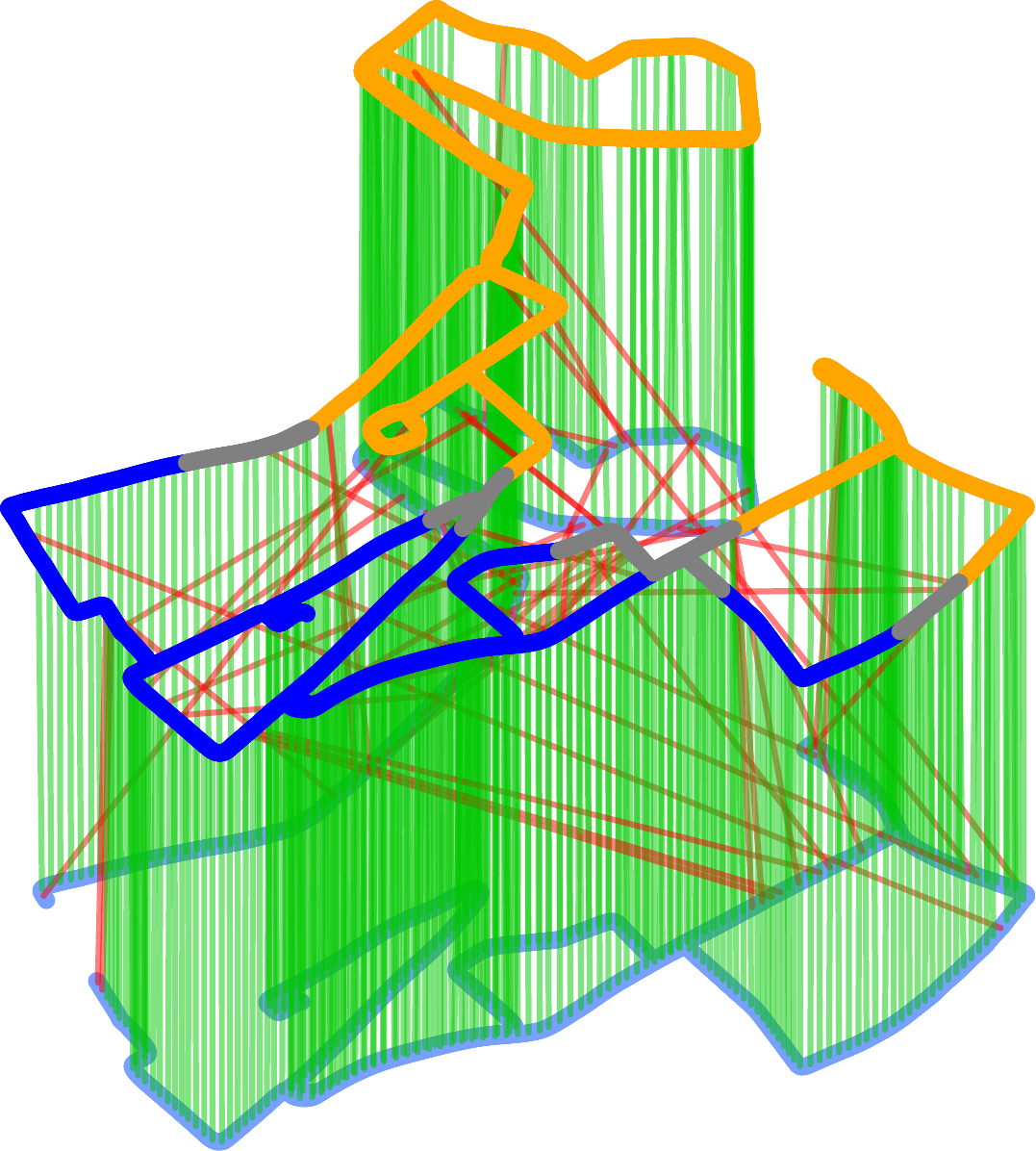}{89.9} &
\suppressionarrow &
\rcornerlabel{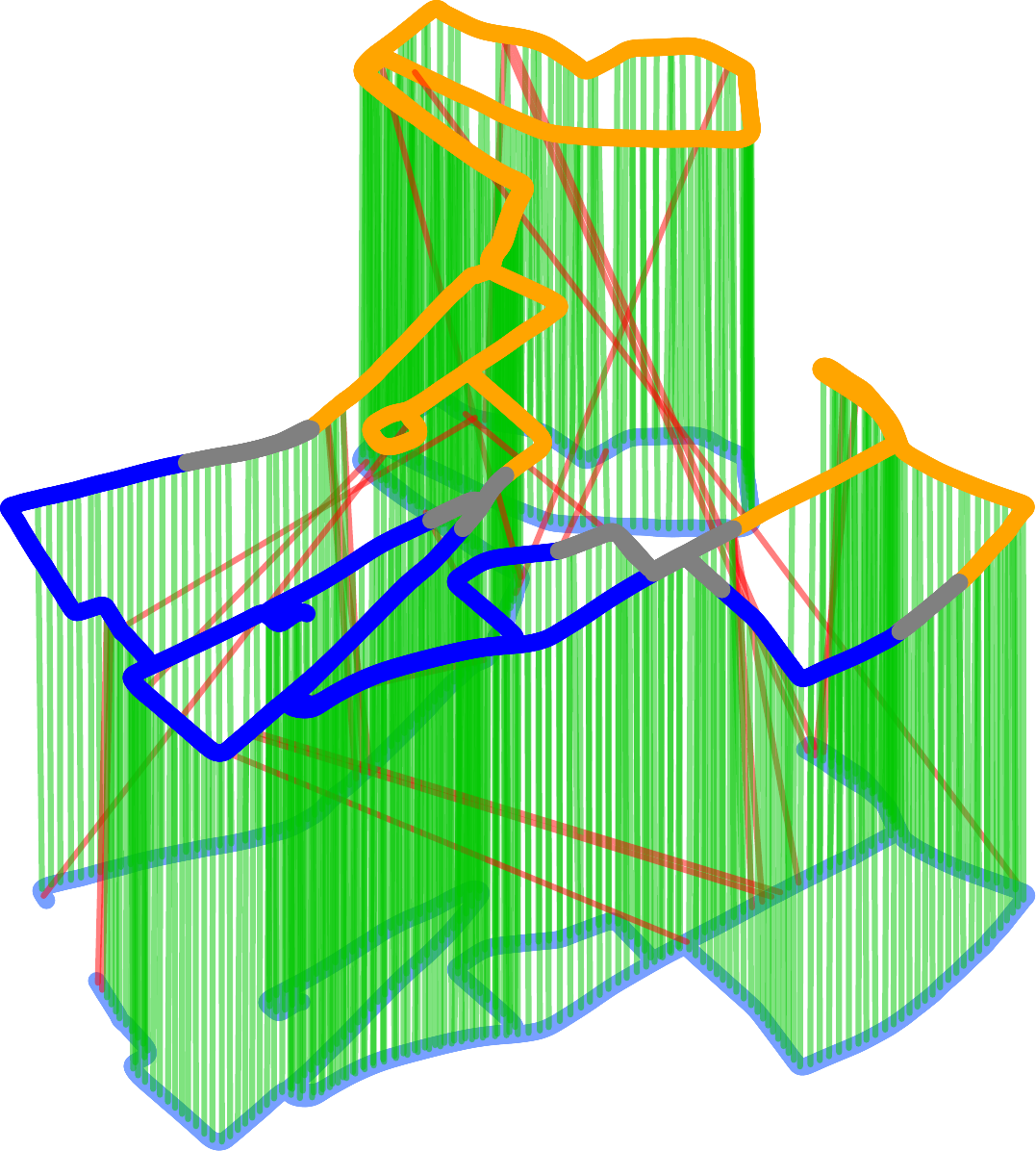}{93.7} \tabularnewline
\centering\rotatebox{90}{\small MixVPR} &
\rcornerlabel{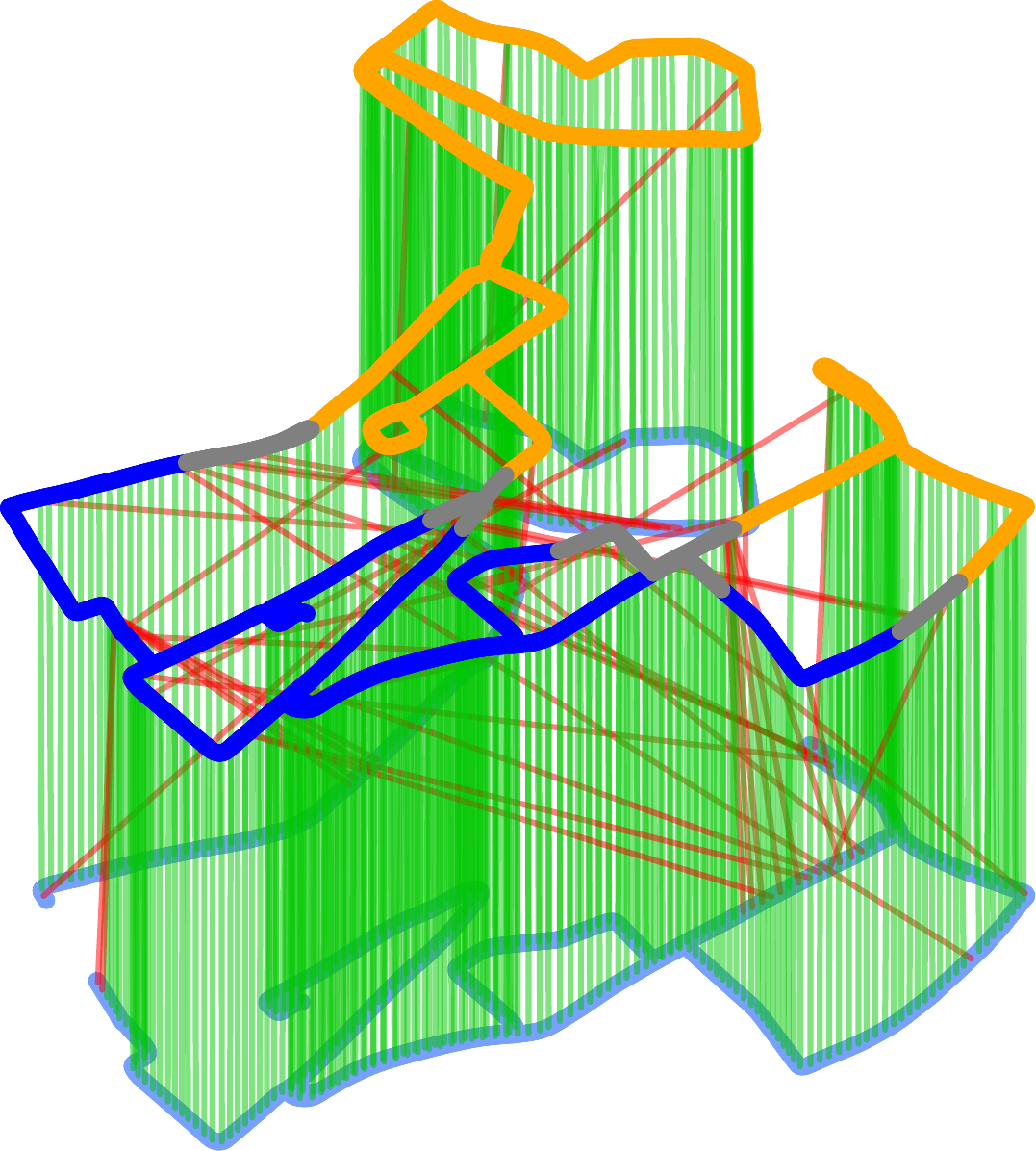}{90.5} &
\suppressionarrow &
\rcornerlabel{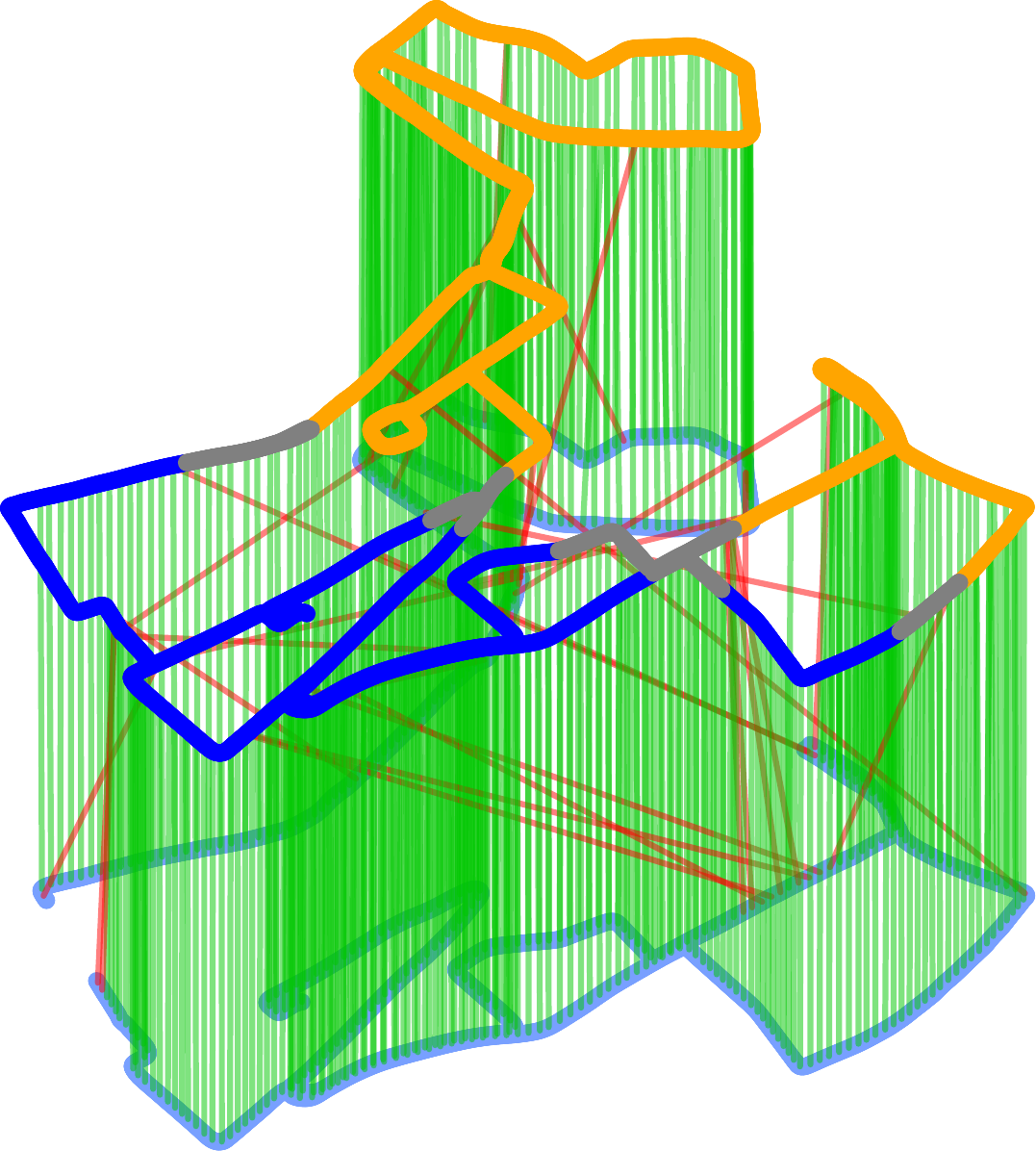}{92.7} \tabularnewline
\end{tabular}

\vspace{2pt}
\begingroup
\definecolor{traincolor}{RGB}{128,128,128}
\definecolor{querycolor}{RGB}{120,160,255}
\definecolor{samedbcolor}{RGB}{0,0,255}
\definecolor{crossdbcolor}{RGB}{255,165,0}
\small
\fcolorbox{black}{traincolor}{\rule{0pt}{6pt}\rule{6pt}{0pt}}~Fitting (overlap)\quad
\fcolorbox{black}{querycolor}{\rule{0pt}{6pt}\rule{6pt}{0pt}}~Day query\quad
\fcolorbox{black}{crossdbcolor}{\rule{0pt}{6pt}\rule{6pt}{0pt}}~Night DB\quad
\fcolorbox{black}{samedbcolor}{\rule{0pt}{6pt}\rule{6pt}{0pt}}~Day DB
\endgroup

\caption{Top-1 retrieval on RobotCar before (left) and after INLP (right).
INLP is fit on the database traversals' 10\% overlap.
Lines show 500 top-1 retrievals: green within 25\,m, red otherwise.}
\label{fig:distractor_inlp_grid}

\vspace{-1mm}
\end{figure}
 On RobotCar (\Cref{fig:distractor_inlp_grid}), daytime and
nighttime database traversals cover opposite halves of the route,
with an overlap covering about 10\% of the database route.
Queries come from another daytime traversal. We show methods
with noticeable residual gaps between \ac{R1} and \ac{DARx}@1 in
\Cref{fig:dar_decomposition}. INLP improves \ac{R1} for all
methods shown, with larger gains for methods with larger gaps.

\providecommand{\suppressionarrow}{%
  \begin{tikzpicture}[inner sep=0pt, outer sep=0pt]
    \draw[-{Triangle[length=3mm, width=3mm]}, line width=3pt] (0,0) -- (0.5,0);
  \end{tikzpicture}%
}
\providecommand{\rstriplabel}[3][]{%
  \begin{tikzpicture}[inner sep=0pt, outer sep=0pt]
    \node[draw, line width=\fboxrule, inner sep=1pt] (img) {%
      \includegraphics[width=\dimexpr\linewidth-2pt-2\fboxrule\relax]{#2}};
    \node[anchor=south west, inner sep=1pt, font=\small]
      at ([yshift=1pt]img.north west) {#1};
    \node[anchor=south east, inner sep=1pt, font=\scriptsize]
      at ([yshift=1pt]img.north east) {R@1: #3\%};
  \end{tikzpicture}%
}
\begin{figure}[t!]
\centering
\setlength{\tabcolsep}{2pt}
\renewcommand{\arraystretch}{1.3}
\setlength{\fboxsep}{0pt}
\begin{tabular}{@{}m{0.44\columnwidth} c m{0.44\columnwidth}@{}}
\centering\small Original & & \centering\small Erased \tabularnewline
\rstriplabel[CricaVPR]{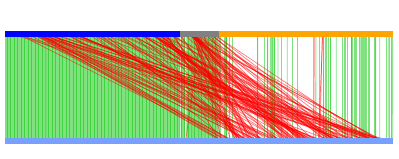}{68.6} &
\suppressionarrow &
\rstriplabel{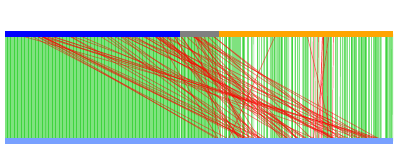}{86.4} \tabularnewline
\rstriplabel[BoQ]{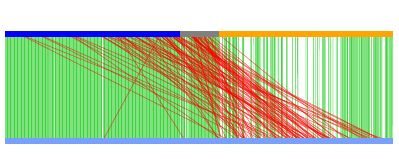}{82.0} &
\suppressionarrow &
\rstriplabel{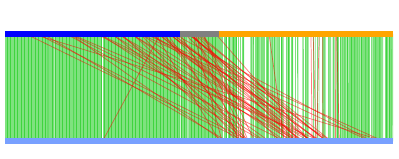}{89.8} \tabularnewline
\rstriplabel[SALAD]{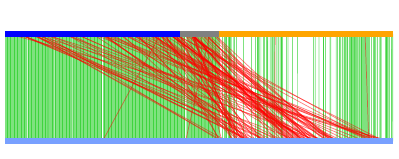}{76.5} &
\suppressionarrow &
\rstriplabel{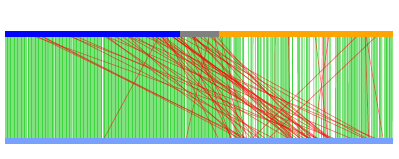}{88.6} \tabularnewline
\rstriplabel[MegaLoc]{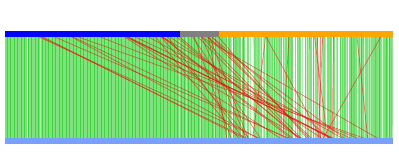}{90.2} &
\suppressionarrow &
\rstriplabel{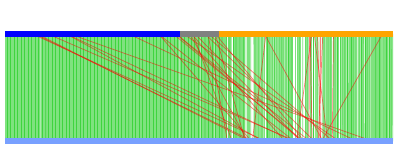}{93.6} \tabularnewline
\rstriplabel[SALAD-CM]{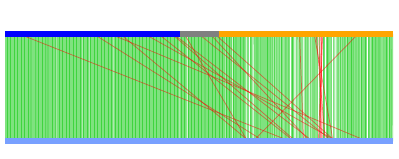}{95.4} &
\suppressionarrow &
\rstriplabel{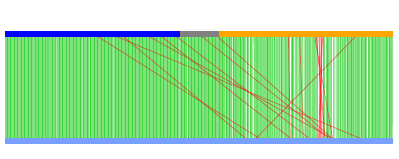}{96.3} \tabularnewline
\end{tabular}

\vspace{2pt}
\begingroup
\definecolor{traincolor}{RGB}{128,128,128}
\definecolor{querycolor}{RGB}{120,160,255}
\definecolor{samedbcolor}{RGB}{0,0,255}
\definecolor{crossdbcolor}{RGB}{255,165,0}
\small
\mbox{\fcolorbox{black}{traincolor}{\rule{0pt}{6pt}\rule{6pt}{0pt}}~Fitting (overlap)}\enspace
\mbox{\fcolorbox{black}{querycolor}{\rule{0pt}{6pt}\rule{6pt}{0pt}}~Summer query}\enspace
\mbox{\fcolorbox{black}{crossdbcolor}{\rule{0pt}{6pt}\rule{6pt}{0pt}}~Winter DB}\enspace
\mbox{\fcolorbox{black}{samedbcolor}{\rule{0pt}{6pt}\rule{6pt}{0pt}}~Fall DB}
\endgroup

\caption{Top-1 retrieval on Nordland before (left) and after INLP (right).
Queries (bottom) and database images (top) follow the frame index.
INLP is fit on the database traversals' 10\% overlap.
Lines show top-1 retrievals: green within 10 frames,
red otherwise.}
\label{fig:nordland_inlp_grid}

\vspace{-1mm}
\end{figure}
 Even methods with small residual gaps on RobotCar exhibit
larger gaps on Nordland (\Cref{fig:dar_decomposition}).
We therefore demonstrate INLP on these methods in
\Cref{fig:nordland_inlp_grid}, using fall and winter
database traversals whose overlap contains 10\% of each
traversal's frames, with queries from a summer traversal.
INLP again improves \ac{R1} for all methods shown. Results for
the remaining methods on both datasets are provided in the
supplementary material (Sec.~S5).
\section{Conclusion and Limitations}
\label{sec:conclusion}
We used \ac{DAR} to evaluate condition blindness through
controlled distractor sets. Although high \ac{R1} generally
transfers to high \ac{DAR}@1, \ac{DAR} reveals how individual
methods respond to distractors. We compared INLP, LEACE,
and \ac{STD} and demonstrated INLP on multi-traversal databases
constructed from RobotCar and Nordland.

Several questions remain. Extending this demonstration
to \ac{STD} or LEACE is nontrivial. Since INLP yields
an orthogonal projection, transforming only the database
descriptors gives the same cosine-similarity retrieval
ranking as transforming both query and database descriptors.
However, \ac{STD} and LEACE do not guarantee unchanged retrieval
rankings when the query transform is omitted and may therefore
require additional query processing. Extending to more than two
database traversals also raises questions about how to apply
multiple transforms and whether they remain effective.

Additionally, the main results use the query set as the
distractor pool to isolate the query--database condition shift.
The supplementary material (Sec.~S2) extends the distractor condition
ablation to all datasets, but it still remains limited
to the conditions available in each dataset.
More diverse distractor pools could provide
a more comprehensive assessment of condition blindness.
\section*{LLM Usage Statement}
\label{sec:llm_statement}
We used Claude Code to review the entire manuscript for grammar
and clarity and to assist with code generation for all experiments and visualizations.

{
  \small
  \nocite{keetha2023anyloc}

}

\clearpage
\setcounter{section}{0}
\setcounter{figure}{0}
\setcounter{table}{0}
\renewcommand{\thesection}{S\arabic{section}}
\renewcommand{\thefigure}{S\arabic{figure}}
\renewcommand{\thetable}{S\arabic{table}}
\renewcommand{\theHsection}{supp.\arabic{section}}
\renewcommand{\theHfigure}{supp.\arabic{figure}}
\renewcommand{\theHtable}{supp.\arabic{table}}
\maketitlesupplementary
\section{Gap Decomposition on \acs{StLuciaMToD} and SVOX}
\label{sec:supp_dar_decomposition}
Fig.~5 of the main paper decomposes the gap between
\ac{R1} and \ac{DAR}@1 on Gardens Point, RobotCar, and Nordland.
\Cref{fig:supp_dar_decomposition} shows the same decomposition on \ac{StLuciaMToD}
and SVOX.
\section{Distractor Sets on All Datasets}
\label{sec:supp_distractor_sets}
The main paper varies the distractor set on RobotCar only. Here we repeat
this ablation on Gardens Point, Nordland, \ac{StLuciaMToD}, and SVOX
(\cref{fig:supp_distractor_sets}). AmsterTime has
no available image set for distractor ablation.
\section{Replacing Database Images with Distractors}
\label{sec:supp_dar_replace}
\ac{DAR} adds the distractors to the database, so each query is ranked against
more images than for \ac{R1}. To check that \ac{DAR}@1 does not drop merely
because of this larger retrieval pool, we keep its size fixed.
For each query, we retain all positives and replace negatives
with distractors one by one, selecting both uniformly at random
from their remaining sets until either set runs out.
The resulting retrieval pool has the same size as for \ac{R1}
but contains distractors in place of some or all negatives.

\Cref{tab:supp_dar_replace} compares replacement with the main
\ac{DAR}@1 results. The absolute difference is at most
2.0~\ac{pp} and averages 0.20~\ac{pp} across 66 method--dataset
pairs, with 36 unchanged. These results suggest that the drop
from \ac{R1} to \ac{DAR}@1 primarily reflects distractor
influence rather than the larger retrieval pool.
\section{Concept Erasure Details}
\label{sec:supp_condition_suppression}

\paragraph{Center Fitting Region.}
Fig.~\ref*{fig:linear_classifier_gain_heatmap} of the main paper
fits concept erasure on a spatially contiguous region around the map center.
\Cref{fig:supp_center_split} shows this region for each dataset. It holds 10\%
of the queries, and all database frames inside it are also used for fitting.

For Fig.~\ref*{fig:concept_erasure_curve}, we fix the evaluation
set to frames outside the region containing 50\% of the queries.
Each fitting fraction uses frames within a region containing
the corresponding fraction of queries. All regions share the
same center, so the fitting regions are nested and remain
disjoint from the evaluation set.

\paragraph{Raw Values.}
\Cref{tab:supp_suppression_raw_split} lists the
\ac{DAR}@1, \ac{DARx}@1, and \ac{R1} of each \ac{VPR} method
before and after concept erasure, from which the gains in
Fig.~\ref*{fig:linear_classifier_gain_heatmap} of the main paper
are computed.

\paragraph{INLP Configuration.}
At each INLP iteration, we fit a linear classifier using
\texttt{SGDClassifier} from scikit-learn v1.8.0.
We use its default settings except \texttt{max\_iter}=2500
and balanced class weights. %
\section{INLP in Multi-Traversal Scenarios}
\label{sec:supp_multi_traversal_inlp}
\paragraph{Nordland.}
\Cref{fig:supp_nordland_inlp_grid} repeats the
demonstration in Fig.~\ref*{fig:nordland_inlp_grid} of the main
paper for the remaining methods. We use the central 10\% of
the traversal frames as the overlap region. INLP improves
\ac{R1} for all methods shown.

\paragraph{RobotCar.}
\Cref{fig:supp_robotcar_inlp_grid} repeats the
demonstration in Fig.~\ref*{fig:distractor_inlp_grid} of the main
paper for the remaining methods. We divide the map
into two halves along a line perpendicular
to the principal axis of the image positions. The overlap band
is centered on this line and expanded symmetrically
until it contains 10\% of the frames from the two
full traversals provided by the database sets. This corresponds to
about 10\% of the route. Their \ac{R1} is already
at least 98.1\%, and INLP changes it by at most 0.1~\ac{pp}.
\section{Dataset Details}
\label{sec:supp_dataset_preparation}
We use RobotCar \texttt{stereo\_centre} images and the SVOX
test set, subsample RobotCar to 1\,m spacing, keep one
\ac{StLuciaMToD} frame per GPS fix (about 11\,m apart),
use the SVOX query and gallery sets as provided, and
exclude queries without a ground-truth database match.
AmsterTime contains byte-identical images shared by several
pairs; we keep one pair per duplicate group, excluding
111 of the 1,231 pairs, and report all AmsterTime
results on the remaining 1,120.

\paragraph{RobotCar Traversals.}
\Cref{tab:supp_robotcar_traversals} lists the full RobotCar
traversal identifiers and their roles in the main paper.
In Fig.~\ref*{fig:umap_grid}, \textit{Day} denotes the
\texttt{sun,clouds} query traversal and \textit{Night} the
nighttime database traversal.
\section{VPR Method Configurations}
\label{sec:supp_methods_config}
\Cref{tab:methods_config} lists the configurations of the eleven
\ac{VPR} methods evaluated in the main paper.

\newlength{\suppcolumnwidth}
\setlength{\suppcolumnwidth}{\columnwidth}
\onecolumn
\ifdefined\nolinenumbers\nolinenumbers\fi

\noindent\begin{minipage}{\textwidth}
\vspace{0pt}
\noindent\begin{minipage}[t]{\suppcolumnwidth}
\vspace{0pt}
\begin{minipage}[t]{\linewidth}
\captionsetup{type=figure}
\centering
\begingroup
\definecolor{nbpart}{HTML}{2A78D6}
\definecolor{cdpart}{HTML}{EB6834}
\pgfplotsset{
  dumbbell/.style={
    scale only axis, width=0.76\columnwidth, height=104pt,
    xmin=0, xmax=101, ymin=-12.4, ymax=0.6,
    xtick={0,50,100}, tick label style={font=\scriptsize},
    title style={font=\scriptsize, yshift=-0.5ex},
    xmajorgrids, major grid style={line width=.1pt,draw=gray!25},
    ytick style={draw=none}, axis x line*=bottom, axis y line*=left, clip=false,
  },
}
\begin{tikzpicture}
\begin{groupplot}[
  group style={group size=1 by 2, vertical sep=1cm, yticklabels at=edge left},
  dumbbell,
  ytick={0,-1,-2,-3,-4.6,-5.6,-7.2,-8.2,-9.8,-10.8,-11.8},
  yticklabels={{AnyLoc},{CosPlace},{EigenPlaces},{MixVPR},{CricaVPR},{SelaVPR},{BoQ},{SALAD},{MegaLoc},{SALAD-CM},{SelaVPR++}},
]
\nextgroupplot[title={St.~Lucia\strut}]
  \draw[gray!60, dashed, line width=0.3pt] (axis cs:0,-3.8) -- (axis cs:101,-3.8);
  \draw[gray!60, dashed, line width=0.3pt] (axis cs:0,-6.4) -- (axis cs:101,-6.4);
  \draw[gray!60, dashed, line width=0.3pt] (axis cs:0,-9) -- (axis cs:101,-9);
  \draw[nbpart, line width=1pt, line cap=round] (axis cs:44.2,0) -- (axis cs:60.6,0); \draw[cdpart, line width=1pt, line cap=round] (axis cs:60.6,0) -- (axis cs:79.9,0); %
  \draw[nbpart, line width=1pt, line cap=round] (axis cs:98.9,-1) -- (axis cs:99.6,-1); \draw[cdpart, line width=1pt, line cap=round] (axis cs:99.6,-1) -- (axis cs:99.6,-1); %
  \draw[nbpart, line width=1pt, line cap=round] (axis cs:98.4,-2) -- (axis cs:99.0,-2); \draw[cdpart, line width=1pt, line cap=round] (axis cs:99.0,-2) -- (axis cs:99.3,-2); %
  \draw[nbpart, line width=1pt, line cap=round] (axis cs:97.7,-3) -- (axis cs:99.2,-3); \draw[cdpart, line width=1pt, line cap=round] (axis cs:99.2,-3) -- (axis cs:99.5,-3); %
  \draw[nbpart, line width=1pt, line cap=round] (axis cs:97.7,-4.6) -- (axis cs:99.7,-4.6); \draw[cdpart, line width=1pt, line cap=round] (axis cs:99.7,-4.6) -- (axis cs:100,-4.6); %
  \draw[nbpart, line width=1pt, line cap=round] (axis cs:98.2,-5.6) -- (axis cs:99.4,-5.6); \draw[cdpart, line width=1pt, line cap=round] (axis cs:99.4,-5.6) -- (axis cs:99.5,-5.6); %
  \draw[nbpart, line width=1pt, line cap=round] (axis cs:99.6,-7.2) -- (axis cs:99.9,-7.2); \draw[cdpart, line width=1pt, line cap=round] (axis cs:99.9,-7.2) -- (axis cs:99.9,-7.2); %
  \draw[nbpart, line width=1pt, line cap=round] (axis cs:99.1,-8.2) -- (axis cs:99.9,-8.2); \draw[cdpart, line width=1pt, line cap=round] (axis cs:99.9,-8.2) -- (axis cs:99.9,-8.2); %
  \draw[nbpart, line width=1pt, line cap=round] (axis cs:99.6,-9.8) -- (axis cs:99.9,-9.8); \draw[cdpart, line width=1pt, line cap=round] (axis cs:99.9,-9.8) -- (axis cs:99.9,-9.8); %
  \draw[nbpart, line width=1pt, line cap=round] (axis cs:99.9,-10.8) -- (axis cs:99.9,-10.8); \draw[cdpart, line width=1pt, line cap=round] (axis cs:99.9,-10.8) -- (axis cs:100,-10.8); %
  \draw[nbpart, line width=1pt, line cap=round] (axis cs:99.9,-11.8) -- (axis cs:100,-11.8); \draw[cdpart, line width=1pt, line cap=round] (axis cs:100,-11.8) -- (axis cs:100,-11.8); %
  \addplot[only marks, mark=o, mark size=1.1pt, mark options={black, fill=white, line width=0.4pt}] coordinates {(44.2,0) (98.9,-1) (98.4,-2) (97.7,-3) (97.7,-4.6) (98.2,-5.6) (99.6,-7.2) (99.1,-8.2) (99.6,-9.8) (99.9,-10.8) (99.9,-11.8)};
  \addplot[only marks, mark=diamond*, mark size=1.3pt, black] coordinates {(60.6,0) (99.6,-1) (99.0,-2) (99.2,-3) (99.7,-4.6) (99.4,-5.6) (99.9,-7.2) (99.9,-8.2) (99.9,-9.8) (99.9,-10.8) (100,-11.8)};
  \addplot[only marks, mark=|, mark size=1.8pt, black, line width=0.7pt] coordinates {(79.9,0) (99.6,-1) (99.3,-2) (99.5,-3) (100,-4.6) (99.5,-5.6) (99.9,-7.2) (99.9,-8.2) (99.9,-9.8) (100,-10.8) (100,-11.8)};
\nextgroupplot[title={SVOX\strut}]
  \draw[gray!60, dashed, line width=0.3pt] (axis cs:0,-3.8) -- (axis cs:101,-3.8);
  \draw[gray!60, dashed, line width=0.3pt] (axis cs:0,-6.4) -- (axis cs:101,-6.4);
  \draw[gray!60, dashed, line width=0.3pt] (axis cs:0,-9) -- (axis cs:101,-9);
  \draw[nbpart, line width=1pt, line cap=round] (axis cs:1.38,0) -- (axis cs:5.62,0); \draw[cdpart, line width=1pt, line cap=round] (axis cs:5.62,0) -- (axis cs:89.3,0); %
  \draw[nbpart, line width=1pt, line cap=round] (axis cs:76.6,-1) -- (axis cs:87.4,-1); \draw[cdpart, line width=1pt, line cap=round] (axis cs:87.4,-1) -- (axis cs:89.7,-1); %
  \draw[nbpart, line width=1pt, line cap=round] (axis cs:77.1,-2) -- (axis cs:88.3,-2); \draw[cdpart, line width=1pt, line cap=round] (axis cs:88.3,-2) -- (axis cs:91.6,-2); %
  \draw[nbpart, line width=1pt, line cap=round] (axis cs:71.2,-3) -- (axis cs:91.9,-3); \draw[cdpart, line width=1pt, line cap=round] (axis cs:91.9,-3) -- (axis cs:96.1,-3); %
  \draw[nbpart, line width=1pt, line cap=round] (axis cs:56.2,-4.6) -- (axis cs:90.2,-4.6); \draw[cdpart, line width=1pt, line cap=round] (axis cs:90.2,-4.6) -- (axis cs:97.0,-4.6); %
  \draw[nbpart, line width=1pt, line cap=round] (axis cs:71.9,-5.6) -- (axis cs:91.6,-5.6); \draw[cdpart, line width=1pt, line cap=round] (axis cs:91.6,-5.6) -- (axis cs:95.8,-5.6); %
  \draw[nbpart, line width=1pt, line cap=round] (axis cs:86.9,-7.2) -- (axis cs:97.1,-7.2); \draw[cdpart, line width=1pt, line cap=round] (axis cs:97.1,-7.2) -- (axis cs:98.4,-7.2); %
  \draw[nbpart, line width=1pt, line cap=round] (axis cs:83.1,-8.2) -- (axis cs:97.7,-8.2); \draw[cdpart, line width=1pt, line cap=round] (axis cs:97.7,-8.2) -- (axis cs:98.3,-8.2); %
  \draw[nbpart, line width=1pt, line cap=round] (axis cs:90.1,-9.8) -- (axis cs:97.6,-9.8); \draw[cdpart, line width=1pt, line cap=round] (axis cs:97.6,-9.8) -- (axis cs:97.7,-9.8); %
  \draw[nbpart, line width=1pt, line cap=round] (axis cs:91.5,-10.8) -- (axis cs:98.0,-10.8); \draw[cdpart, line width=1pt, line cap=round] (axis cs:98.0,-10.8) -- (axis cs:98.5,-10.8); %
  \draw[nbpart, line width=1pt, line cap=round] (axis cs:92.3,-11.8) -- (axis cs:98.5,-11.8); \draw[cdpart, line width=1pt, line cap=round] (axis cs:98.5,-11.8) -- (axis cs:98.5,-11.8); %
  \addplot[only marks, mark=o, mark size=1.1pt, mark options={black, fill=white, line width=0.4pt}] coordinates {(1.38,0) (76.6,-1) (77.1,-2) (71.2,-3) (56.2,-4.6) (71.9,-5.6) (86.9,-7.2) (83.1,-8.2) (90.1,-9.8) (91.5,-10.8) (92.3,-11.8)};
  \addplot[only marks, mark=diamond*, mark size=1.3pt, black] coordinates {(5.62,0) (87.4,-1) (88.3,-2) (91.9,-3) (90.2,-4.6) (91.6,-5.6) (97.1,-7.2) (97.7,-8.2) (97.6,-9.8) (98.0,-10.8) (98.5,-11.8)};
  \addplot[only marks, mark=|, mark size=1.8pt, black, line width=0.7pt] coordinates {(89.3,0) (89.7,-1) (91.6,-2) (96.1,-3) (97.0,-4.6) (95.8,-5.6) (98.4,-7.2) (98.3,-8.2) (97.7,-9.8) (98.5,-10.8) (98.5,-11.8)};
\end{groupplot}
\node[font=\scriptsize, anchor=north] at ($(group c1r2.south)+(0,-0.45cm)$) {Recall (\%)};
\end{tikzpicture}\\[-1mm]
\newcommand{\suppdarlegenditem}[2]{%
  \mbox{\tikz[baseline=-0.5ex]{%
    \path[use as bounding box] (0,-2pt) rectangle (10pt,2pt);
    #1
  }\,#2}%
}
{\scriptsize
\begin{tabular}{@{}c@{}}
\suppdarlegenditem{\draw[black, fill=white, line width=0.4pt] (5pt,0) circle[radius=1.1pt];}{DAR@1}\quad
\suppdarlegenditem{\draw[black] plot[only marks, mark=diamond*, mark size=1.3pt] coordinates {(5pt,0)};}{\acs{DARx}@1}\quad
\suppdarlegenditem{\draw[black, line width=0.7pt] (5pt,-1.8pt) -- (5pt,1.8pt);}{R@1}\\
\suppdarlegenditem{\draw[nbpart, line width=1pt] (0,0) -- (10pt,0);}{Closed by exclusion}\quad
\suppdarlegenditem{\draw[cdpart, line width=1pt] (0,0) -- (10pt,0);}{Remaining gap}
\end{tabular}}
\endgroup
\caption{Decomposition of the gap between \ac{R1} and \ac{DAR}@1 on
\acs{StLuciaMToD} (top) and SVOX (bottom). The blue segment is the part of the gap closed by
excluding nearby distractors (from \ac{DAR}@1 to \acs{DARx}@1), and the orange
segment is the gap that remains.}
\label{fig:supp_dar_decomposition}
\end{minipage}
 \end{minipage}\hfill
\begin{minipage}[t]{\suppcolumnwidth}
\vspace{0pt}
\providecommand{\suppressionarrow}{%
  \begin{tikzpicture}[inner sep=0pt, outer sep=0pt]
    \draw[-{Triangle[length=3mm, width=3mm]}, line width=3pt] (0,0) -- (0.5,0);
  \end{tikzpicture}%
}
\providecommand{\rstriplabel}[3][]{%
  \begin{tikzpicture}[inner sep=0pt, outer sep=0pt]
    \node[draw, line width=\fboxrule, inner sep=1pt] (img) {%
      \includegraphics[width=\dimexpr\linewidth-2pt-2\fboxrule\relax]{#2}};
    \node[anchor=south west, inner sep=1pt, font=\small]
      at ([yshift=1pt]img.north west) {#1};
    \node[anchor=south east, inner sep=1pt, font=\scriptsize]
      at ([yshift=1pt]img.north east) {R@1: #3\%};
  \end{tikzpicture}%
}
\newcommand{\nlpanel}[1]{figures/nordland_inlp/viz_nordland_overlap_concept_erasure_#1.pdf}
\begin{minipage}[t]{\linewidth}
\captionsetup{type=figure}
\centering
\setlength{\tabcolsep}{2pt}
\renewcommand{\arraystretch}{1.3}
\setlength{\fboxsep}{0pt}
\begin{tabular}{@{}m{0.44\columnwidth} c m{0.44\columnwidth}@{}}
\centering\small Original & & \centering\small Erased \tabularnewline
\rstriplabel[AnyLoc]{\nlpanel{AnyLoc_q-summer_db-fall-n15176_winter-n15176_original_R_at_1-49.38}}{49.4} &
\suppressionarrow &
\rstriplabel{\nlpanel{AnyLoc_q-summer_db-fall-n15176_winter-n15176_inlp_R_at_1-56.51}}{56.5} \tabularnewline
\rstriplabel[CosPlace]{\nlpanel{CosPlace_q-summer_db-fall-n15176_winter-n15176_original_R_at_1-59.93}}{59.9} &
\suppressionarrow &
\rstriplabel{\nlpanel{CosPlace_q-summer_db-fall-n15176_winter-n15176_inlp_R_at_1-68.70}}{68.7} \tabularnewline
\rstriplabel[EigenPlaces]{\nlpanel{EigenPlaces_q-summer_db-fall-n15176_winter-n15176_original_R_at_1-59.61}}{59.6} &
\suppressionarrow &
\rstriplabel{\nlpanel{EigenPlaces_q-summer_db-fall-n15176_winter-n15176_inlp_R_at_1-68.45}}{68.5} \tabularnewline
\rstriplabel[MixVPR]{\nlpanel{MixVPR_q-summer_db-fall-n15176_winter-n15176_original_R_at_1-58.10}}{58.1} &
\suppressionarrow &
\rstriplabel{\nlpanel{MixVPR_q-summer_db-fall-n15176_winter-n15176_inlp_R_at_1-66.09}}{66.1} \tabularnewline
\end{tabular}

\vspace{2pt}
\begingroup
\definecolor{traincolor}{RGB}{128,128,128}
\definecolor{querycolor}{RGB}{120,160,255}
\definecolor{samedbcolor}{RGB}{0,0,255}
\definecolor{crossdbcolor}{RGB}{255,165,0}
\small
\mbox{\fcolorbox{black}{traincolor}{\rule{0pt}{6pt}\rule{6pt}{0pt}}~Fitting (overlap)}\enspace
\mbox{\fcolorbox{black}{querycolor}{\rule{0pt}{6pt}\rule{6pt}{0pt}}~Summer query}\enspace
\mbox{\fcolorbox{black}{crossdbcolor}{\rule{0pt}{6pt}\rule{6pt}{0pt}}~Winter DB}\enspace
\mbox{\fcolorbox{black}{samedbcolor}{\rule{0pt}{6pt}\rule{6pt}{0pt}}~Fall DB}
\endgroup

\caption{Top-1 retrieval on Nordland for methods not shown in
Fig.~\ref*{fig:nordland_inlp_grid} of the main paper.}
\label{fig:supp_nordland_inlp_grid}
\end{minipage}
 \par\nointerlineskip\vspace{11pt}
\begin{minipage}[t]{\linewidth}
\captionsetup{type=table}
\centering
\footnotesize
\setlength{\tabcolsep}{3pt}
\begin{tabular}{@{}lll@{}}
\toprule
Condition & Role & Traversal identifier \\
\midrule
\texttt{sun,clouds} & Query & \traversal{2014-11-18-13-20-12} \\
\texttt{overcast} & Day DB / distractor & \traversal{2014-12-12-10-45-15} \\
\texttt{night} & Night DB & \traversal{2014-12-16-18-44-24} \\
\texttt{rain} & Distractor & \traversal{2015-10-29-12-18-17} \\
\texttt{night} & Distractor & \traversal{2014-11-14-16-34-33} \\
\bottomrule
\end{tabular}
\caption{RobotCar traversal identifiers. The query set also
serves as the distractor pool for the main results.}
\label{tab:supp_robotcar_traversals}
\end{minipage}
 \end{minipage}
\par\nointerlineskip\vspace{17pt}
\begin{minipage}[t]{\linewidth}
\captionsetup{type=table}
\centering
\setlength{\tabcolsep}{3pt}
\renewcommand{\arraystretch}{1.05}
\begin{adjustbox}{max width=\textwidth}
\footnotesize
\begin{tabular}{l cc cc cc cc cc cc}
\toprule
 & \multicolumn{2}{c}{AmsterTime} & \multicolumn{2}{c}{Gardens Point} & \multicolumn{2}{c}{Nordland} & \multicolumn{2}{c}{RobotCar} & \multicolumn{2}{c}{StLuciaMToD} & \multicolumn{2}{c}{SVOX} \\
\cmidrule(lr){2-3} \cmidrule(lr){4-5} \cmidrule(lr){6-7} \cmidrule(lr){8-9} \cmidrule(lr){10-11} \cmidrule(lr){12-13}
\textbf{Method} & DAR@1 & repl. & DAR@1 & repl. & DAR@1 & repl. & DAR@1 & repl. & DAR@1 & repl. & DAR@1 & repl. \\
\midrule
AnyLoc & 10.7 & 10.8 & 15.0 & 15.0 & 2.23 & 2.23 & 4.21 & 4.21 & 44.2 & 44.5 & 1.38 & 1.38 \\
BoQ & 49.4 & 50.5 & 37.5 & 37.5 & 44.4 & 44.4 & 85.3 & 85.3 & 99.6 & 99.6 & 86.9 & 86.9 \\
CosPlace & 40.4 & 42.0 & 18.5 & 18.5 & 15.1 & 15.1 & 38.1 & 38.2 & 98.9 & 99.0 & 76.6 & 76.6 \\
CricaVPR & 41.9 & 42.5 & 29.5 & 29.5 & 19.9 & 19.9 & 51.4 & 51.4 & 97.7 & 97.7 & 56.2 & 56.2 \\
EigenPlaces & 37.2 & 38.2 & 16.0 & 18.0 & 15.3 & 15.3 & 37.0 & 37.0 & 98.4 & 98.4 & 77.1 & 77.1 \\
MegaLoc & 50.3 & 51.3 & 12.5 & 12.5 & 73.1 & 73.1 & 79.7 & 79.7 & 99.6 & 99.6 & 90.1 & 90.1 \\
MixVPR & 25.3 & 25.8 & 9.00 & 9.00 & 9.75 & 9.75 & 32.5 & 32.6 & 97.7 & 97.7 & 71.2 & 71.2 \\
SALAD & 42.2 & 43.0 & 34.0 & 34.0 & 32.1 & 32.1 & 83.1 & 83.1 & 99.1 & 99.2 & 83.1 & 83.1 \\
SALAD-CM & 45.6 & 46.6 & 26.0 & 26.0 & 87.7 & 87.9 & 93.6 & 93.7 & 99.9 & 99.9 & 91.5 & 91.5 \\
SelaVPR & 38.5 & 39.5 & 32.0 & 32.0 & 18.0 & 18.0 & 71.3 & 71.3 & 98.2 & 98.2 & 71.9 & 71.9 \\
SelaVPR++ & 45.7 & 46.5 & 30.5 & 30.5 & 70.4 & 70.3 & 92.0 & 92.0 & 99.9 & 99.9 & 92.3 & 92.3 \\
\bottomrule
\end{tabular}
\end{adjustbox}
\caption{\ac{DAR}@1 (\%) with the distractors added to the database (\ac{DAR}@1, as in the main results) and with each distractor replacing a database image (\textit{repl.}).}
\label{tab:supp_dar_replace}
\end{minipage}
 \par\end{minipage}

\clearpage
\noindent\begin{minipage}{\textwidth}
\vspace{0pt}
\begin{minipage}[t]{\linewidth}
\captionsetup{type=table}
\centering
\setlength{\tabcolsep}{2pt}
\renewcommand{\arraystretch}{1.0}
\resizebox{\textwidth}{!}{\footnotesize%
\begin{tabular}{l cccc cccc cccc cccc cccc cccc}
\toprule
 & \multicolumn{4}{c}{AmsterTime} & \multicolumn{4}{c}{Gardens Point} & \multicolumn{4}{c}{Nordland} & \multicolumn{4}{c}{RobotCar} & \multicolumn{4}{c}{StLuciaMToD} & \multicolumn{4}{c}{SVOX} \\
\cmidrule(lr){2-5} \cmidrule(lr){6-9} \cmidrule(lr){10-13} \cmidrule(lr){14-17} \cmidrule(lr){18-21} \cmidrule(lr){22-25}
\textbf{Method} & Orig. & INLP & LEACE & STD & Orig. & INLP & LEACE & STD & Orig. & INLP & LEACE & STD & Orig. & INLP & LEACE & STD & Orig. & INLP & LEACE & STD & Orig. & INLP & LEACE & STD \\
\midrule
\multicolumn{25}{l}{\textit{\ac{DAR}@1}} \\
AnyLoc & 11.5 & 24.5 & 22.2 & 22.0 & 15.0 & 17.2 & 18.3 & 5.00 & 2.46 & 7.70 & 5.93 & 3.81 & 4.05 & 40.7 & 39.6 & 18.4 & 43.2 & 59.6 & 52.0 & 35.0 & 1.27 & 29.4 & 25.4 & 17.2 \\
BoQ & 50.4 & 50.5 & 50.8 & 47.4 & 34.4 & 34.4 & 34.4 & 16.7 & 47.8 & 58.8 & 57.5 & 54.7 & 84.9 & 85.3 & 85.2 & 78.7 & 99.6 & 99.6 & 99.6 & 99.5 & 87.8 & 88.8 & 88.7 & 81.3 \\
CosPlace & 41.0 & 43.2 & 42.7 & 42.0 & 17.8 & 20.0 & 18.9 & 7.22 & 16.4 & 30.6 & 29.9 & 28.0 & 38.1 & 53.0 & 50.5 & 38.2 & 99.2 & 99.2 & 99.2 & 99.0 & 78.2 & 80.5 & 78.8 & 69.7 \\
CricaVPR & 43.1 & 49.3 & 48.3 & 49.1 & 28.9 & 28.3 & 28.9 & 13.9 & 21.8 & 45.7 & 43.8 & 43.3 & 50.1 & 56.1 & 55.2 & 42.9 & 97.7 & 98.1 & 97.9 & 95.5 & 56.8 & 61.9 & 61.5 & 49.3 \\
EigenPlaces & 37.7 & 40.5 & 39.7 & 39.9 & 16.1 & 18.3 & 17.8 & 5.56 & 16.6 & 30.7 & 30.8 & 28.4 & 36.3 & 54.9 & 56.2 & 44.1 & 98.6 & 98.4 & 98.6 & 97.9 & 79.0 & 81.4 & 80.5 & 70.5 \\
MegaLoc & 51.5 & 53.8 & 53.4 & 53.1 & 10.6 & 12.2 & 12.2 & 3.89 & 74.3 & 81.3 & 81.3 & 81.2 & 79.3 & 80.7 & 80.4 & 73.2 & 99.6 & 99.5 & 99.6 & 98.8 & 90.6 & 91.1 & 90.7 & 83.6 \\
MixVPR & 25.7 & 29.6 & 29.0 & 27.8 & 9.44 & 12.2 & 12.2 & 1.67 & 10.7 & 20.3 & 19.7 & 19.8 & 32.4 & 40.7 & 40.6 & 29.9 & 97.9 & 98.0 & 98.0 & 97.1 & 72.2 & 76.8 & 75.9 & 67.0 \\
SALAD & 44.0 & 47.9 & 46.9 & 47.4 & 32.8 & 32.8 & 33.3 & 16.7 & 34.7 & 51.7 & 50.4 & 44.6 & 82.2 & 84.0 & 83.8 & 75.9 & 99.1 & 99.2 & 99.1 & 98.8 & 83.7 & 84.7 & 84.5 & 76.3 \\
SALAD-CM & 47.2 & 48.4 & 47.6 & 46.5 & 23.3 & 23.9 & 23.3 & 8.89 & 88.2 & 90.7 & 90.6 & 90.5 & 93.3 & 93.4 & 93.4 & 90.2 & 99.9 & 99.9 & 99.9 & 99.9 & 92.1 & 92.6 & 92.5 & 86.8 \\
\midrule
\multicolumn{25}{l}{\textit{\acs{DARx}@1}} \\
AnyLoc & 11.5 & 24.5 & 22.2 & 22.0 & 79.4 & 84.4 & 84.4 & 57.8 & 3.90 & 14.0 & 11.1 & 7.84 & 20.7 & 89.5 & 88.4 & 69.1 & 59.9 & 76.5 & 70.6 & 47.9 & 5.99 & 80.2 & 75.9 & 68.4 \\
BoQ & 50.4 & 50.5 & 50.8 & 47.4 & 100 & 100 & 100 & 93.9 & 58.7 & 73.4 & 72.5 & 72.1 & 99.3 & 99.3 & 99.3 & 98.8 & 99.9 & 99.9 & 99.9 & 99.9 & 98.2 & 98.2 & 98.2 & 97.2 \\
CosPlace & 41.0 & 43.6 & 42.7 & 42.0 & 64.4 & 68.3 & 67.2 & 55.0 & 20.4 & 38.6 & 37.9 & 37.2 & 61.4 & 80.4 & 78.5 & 68.8 & 99.6 & 99.6 & 99.6 & 99.5 & 91.5 & 92.1 & 90.3 & 88.2 \\
CricaVPR & 43.1 & 49.3 & 48.3 & 49.1 & 96.1 & 96.7 & 96.1 & 90.0 & 30.8 & 66.3 & 64.0 & 66.6 & 94.1 & 95.4 & 95.1 & 92.1 & 99.7 & 99.8 & 99.7 & 99.2 & 93.6 & 94.1 & 94.0 & 92.5 \\
EigenPlaces & 37.7 & 40.5 & 39.7 & 39.9 & 72.2 & 74.4 & 74.4 & 49.4 & 20.6 & 39.0 & 39.3 & 38.3 & 59.3 & 81.3 & 81.5 & 72.3 & 99.0 & 98.8 & 99.0 & 98.7 & 93.0 & 93.6 & 92.6 & 88.7 \\
MegaLoc & 51.5 & 53.8 & 53.4 & 53.1 & 90.6 & 90.6 & 90.6 & 63.3 & 81.0 & 87.7 & 87.9 & 88.0 & 98.2 & 98.3 & 98.3 & 97.3 & 99.9 & 99.9 & 99.9 & 99.9 & 98.0 & 98.1 & 98.0 & 96.7 \\
MixVPR & 25.7 & 29.6 & 29.0 & 27.8 & 66.1 & 69.4 & 72.2 & 27.8 & 13.6 & 28.2 & 27.3 & 28.9 & 65.3 & 76.0 & 76.3 & 61.6 & 99.5 & 99.5 & 99.5 & 99.3 & 94.9 & 95.5 & 95.3 & 93.1 \\
SALAD & 44.0 & 47.9 & 46.9 & 47.4 & 97.2 & 96.7 & 97.2 & 88.3 & 46.8 & 70.7 & 69.8 & 67.9 & 99.1 & 99.1 & 99.1 & 98.9 & 99.9 & 99.9 & 99.9 & 99.9 & 98.1 & 98.0 & 98.2 & 96.9 \\
SALAD-CM & 47.2 & 48.4 & 47.6 & 46.5 & 93.9 & 93.3 & 93.9 & 81.1 & 91.8 & 93.7 & 93.6 & 93.7 & 99.3 & 99.3 & 99.3 & 99.2 & 99.9 & 99.9 & 99.9 & 100 & 98.7 & 98.6 & 98.6 & 97.7 \\
\midrule
\multicolumn{25}{l}{\textit{\ac{R1}}} \\
AnyLoc & 45.4 & 54.7 & 52.1 & 56.0 & 93.3 & 95.0 & 92.8 & 93.9 & 24.6 & 43.2 & 42.0 & 51.5 & 96.0 & 96.9 & 96.8 & 97.0 & 79.1 & 92.5 & 88.1 & 70.9 & 91.3 & 91.8 & 91.1 & 91.1 \\
BoQ & 68.7 & 69.0 & 68.7 & 69.4 & 100 & 100 & 100 & 99.4 & 90.4 & 94.4 & 94.1 & 94.8 & 99.4 & 99.4 & 99.4 & 99.5 & 99.9 & 99.9 & 99.9 & 99.9 & 98.2 & 98.2 & 98.2 & 98.0 \\
CosPlace & 55.9 & 56.9 & 56.6 & 57.0 & 77.2 & 80.0 & 80.0 & 82.8 & 50.1 & 67.0 & 67.5 & 71.3 & 92.3 & 93.0 & 92.8 & 93.6 & 99.7 & 99.7 & 99.6 & 99.7 & 92.6 & 93.0 & 91.7 & 93.0 \\
CricaVPR & 68.5 & 70.0 & 68.7 & 72.0 & 96.7 & 97.2 & 96.7 & 98.3 & 84.9 & 92.3 & 91.1 & 94.7 & 98.8 & 98.7 & 98.7 & 98.7 & 100 & 100 & 99.9 & 99.9 & 97.5 & 97.5 & 97.6 & 97.6 \\
EigenPlaces & 54.4 & 56.0 & 55.6 & 57.2 & 78.3 & 80.0 & 78.9 & 79.4 & 55.5 & 68.2 & 69.9 & 74.1 & 92.5 & 93.4 & 93.7 & 94.0 & 99.2 & 99.2 & 99.2 & 99.5 & 93.8 & 94.0 & 93.4 & 93.0 \\
MegaLoc & 69.4 & 71.1 & 70.4 & 71.0 & 93.3 & 93.3 & 93.3 & 95.0 & 93.1 & 93.9 & 94.0 & 94.6 & 98.7 & 98.8 & 98.8 & 98.8 & 99.9 & 99.9 & 99.9 & 100 & 98.0 & 98.1 & 98.0 & 98.1 \\
MixVPR & 44.6 & 48.6 & 48.3 & 48.7 & 84.4 & 83.3 & 86.1 & 85.6 & 61.7 & 73.1 & 72.9 & 79.3 & 90.9 & 91.4 & 91.5 & 91.9 & 99.6 & 99.7 & 99.6 & 99.7 & 96.4 & 96.4 & 96.4 & 96.3 \\
SALAD & 64.3 & 66.1 & 65.2 & 66.7 & 98.3 & 97.8 & 98.3 & 97.8 & 86.2 & 91.8 & 91.9 & 93.7 & 99.1 & 99.1 & 99.1 & 99.1 & 99.9 & 99.9 & 99.9 & 99.9 & 98.3 & 98.1 & 98.3 & 98.0 \\
SALAD-CM & 63.0 & 65.2 & 63.7 & 64.7 & 93.9 & 93.3 & 93.9 & 93.3 & 96.1 & 96.3 & 96.3 & 96.4 & 99.3 & 99.3 & 99.3 & 99.3 & 100 & 100 & 100 & 100 & 98.7 & 98.6 & 98.6 & 98.2 \\
\bottomrule
\end{tabular}%
}
\caption{Raw values behind Fig.~\ref*{fig:linear_classifier_gain_heatmap} of the main paper.}
\label{tab:supp_suppression_raw_split}
\end{minipage}
 \par\nointerlineskip\vspace{11pt}
\providecommand{\suppressionarrow}{%
  \begin{tikzpicture}[inner sep=0pt, outer sep=0pt]
    \draw[-{Triangle[length=3mm, width=3mm]}, line width=3pt] (0,0) -- (0.5,0);
  \end{tikzpicture}%
}
\providecommand{\rcornerlabel}[2]{%
  \begin{tikzpicture}[inner sep=0pt, outer sep=0pt]
    \node[draw, line width=\fboxrule, minimum width=\linewidth,
          minimum height=\linewidth, inner sep=0pt] (img) {%
      \adjincludegraphics[width=\dimexpr 0.95\linewidth-2\fboxrule\relax,
                          height=\dimexpr 0.95\linewidth-2\fboxrule\relax,
                          keepaspectratio]{#1}};
    \node[anchor=north east, fill=white, fill opacity=0.85, text opacity=1,
          inner sep=1.5pt, font=\scriptsize]
      at ([xshift=-2pt,yshift=-2pt]img.north east) {R@1: #2\%};
  \end{tikzpicture}%
}
\providecommand{\ovpanel}[2]{\rcornerlabel{figures/distractor_inlp/viz_overlap_ce_#1_q-2014-11-18-13-20-12_db-2014-12-12-10-45-15-n3947_2014-12-16-18-44-24-n4294_overlap0.1_#2.pdf}}
\begin{minipage}[t]{\linewidth}
\captionsetup{type=figure}
\centering
\setlength{\tabcolsep}{2pt}
\renewcommand{\arraystretch}{1.0}
\setlength{\fboxsep}{0pt}
\begin{tabular}{@{}m{0.4cm} m{0.19\textwidth} c m{0.19\textwidth} @{\hspace{0.6cm}} m{0.4cm} m{0.19\textwidth} c m{0.19\textwidth}@{}}
 & \centering\small Original & & \centering\small Erased &
 & \centering\small Original & & \centering\small Erased \tabularnewline
\centering\rotatebox{90}{\small BoQ} &
\ovpanel{BoQ}{original_R_at_1-99.11}{99.1} & \suppressionarrow & \ovpanel{BoQ}{inlp_R_at_1-99.11}{99.1} &
\centering\rotatebox{90}{\small CricaVPR} &
\ovpanel{CricaVPR}{original_R_at_1-98.06}{98.1} & \suppressionarrow & \ovpanel{CricaVPR}{inlp_R_at_1-98.16}{98.2} \tabularnewline
\centering\rotatebox{90}{\small MegaLoc} &
\ovpanel{MegaLoc}{original_R_at_1-98.50}{98.5} & \suppressionarrow & \ovpanel{MegaLoc}{inlp_R_at_1-98.52}{98.5} &
\centering\rotatebox{90}{\small SALAD} &
\ovpanel{SALAD}{original_R_at_1-98.98}{99.0} & \suppressionarrow & \ovpanel{SALAD}{inlp_R_at_1-98.99}{99.0} \tabularnewline
\centering\rotatebox{90}{\small SALAD-CM} &
\ovpanel{SALAD_CM}{original_R_at_1-98.99}{99.0} & \suppressionarrow & \ovpanel{SALAD_CM}{inlp_R_at_1-98.99}{99.0} &
 & & & \tabularnewline
\end{tabular}

\vspace{2pt}
\begingroup
\definecolor{traincolor}{RGB}{128,128,128}
\definecolor{querycolor}{RGB}{120,160,255}
\definecolor{samedbcolor}{RGB}{0,0,255}
\definecolor{crossdbcolor}{RGB}{255,165,0}
\small
\fcolorbox{black}{traincolor}{\rule{0pt}{6pt}\rule{6pt}{0pt}}~Fitting (overlap)\quad
\fcolorbox{black}{querycolor}{\rule{0pt}{6pt}\rule{6pt}{0pt}}~Day query\quad
\fcolorbox{black}{crossdbcolor}{\rule{0pt}{6pt}\rule{6pt}{0pt}}~Night DB\quad
\fcolorbox{black}{samedbcolor}{\rule{0pt}{6pt}\rule{6pt}{0pt}}~Day DB
\endgroup

\caption{
Top-1 retrieval on RobotCar for methods not shown in
Fig.~\ref*{fig:distractor_inlp_grid} of the main paper.
}
\label{fig:supp_robotcar_inlp_grid}
\end{minipage}
 \par\end{minipage}

\clearpage
\noindent\begin{minipage}{\textwidth}
\vspace{0pt}
\begin{minipage}[t]{\linewidth}
\captionsetup{type=figure}
\centering
\setlength{\tabcolsep}{6pt}
\begin{tabular}{ccc}
\small RobotCar & \small St.~Lucia & \small SVOX \\
\includegraphics[height=3.6cm]{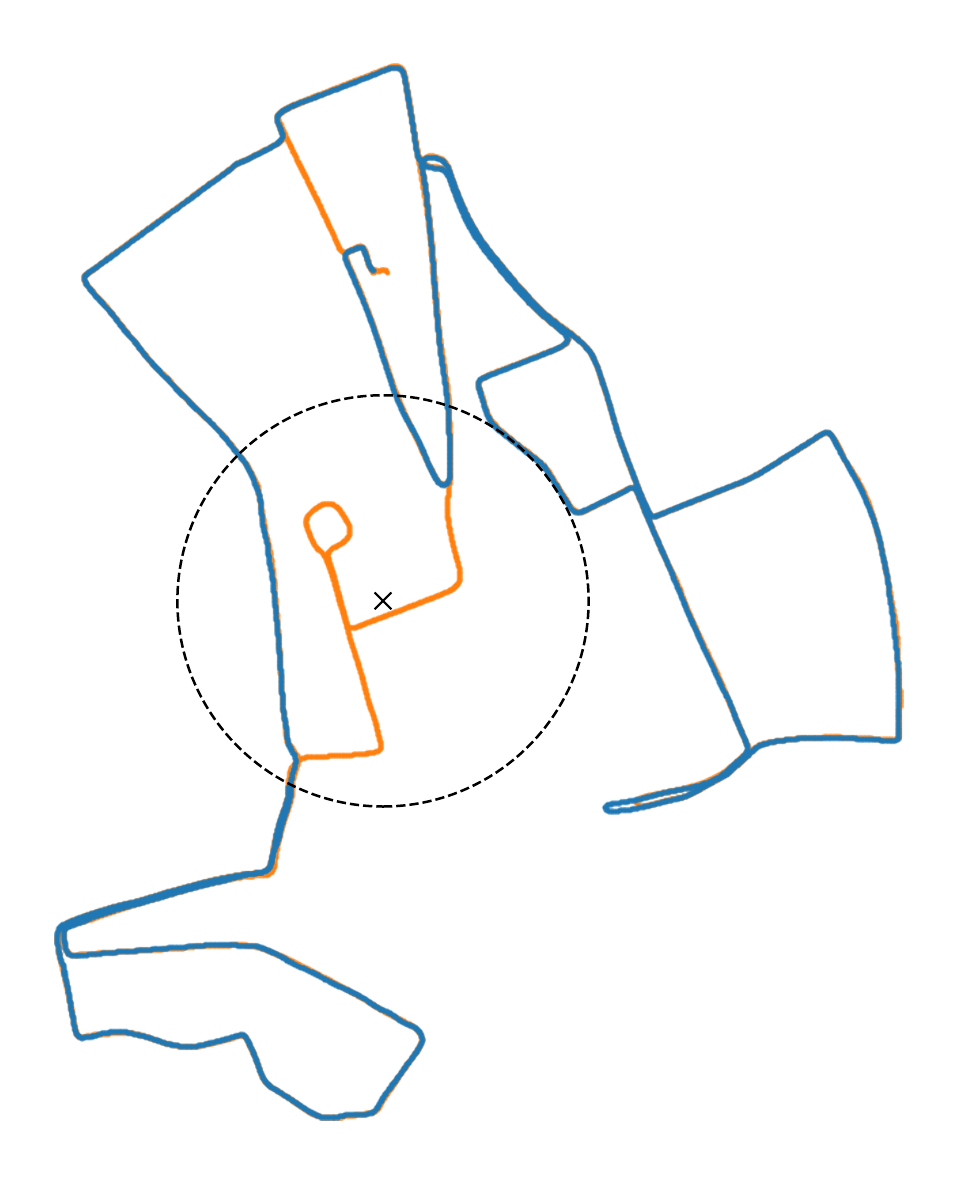} &
\includegraphics[height=3.6cm]{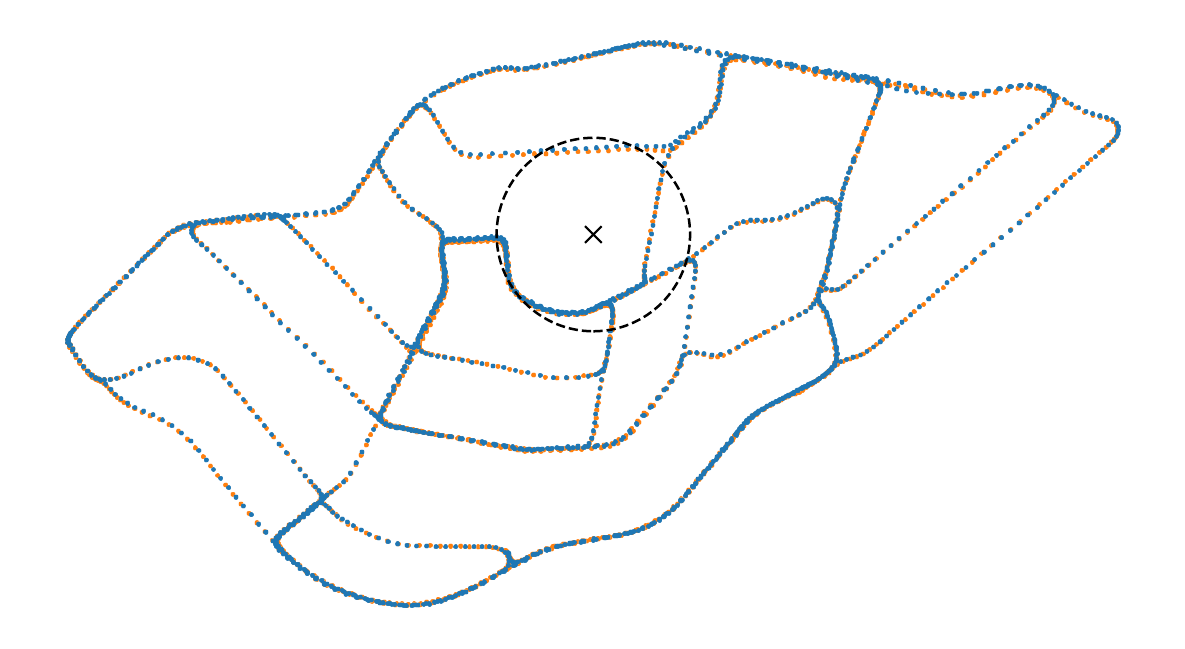} &
\includegraphics[height=3.6cm]{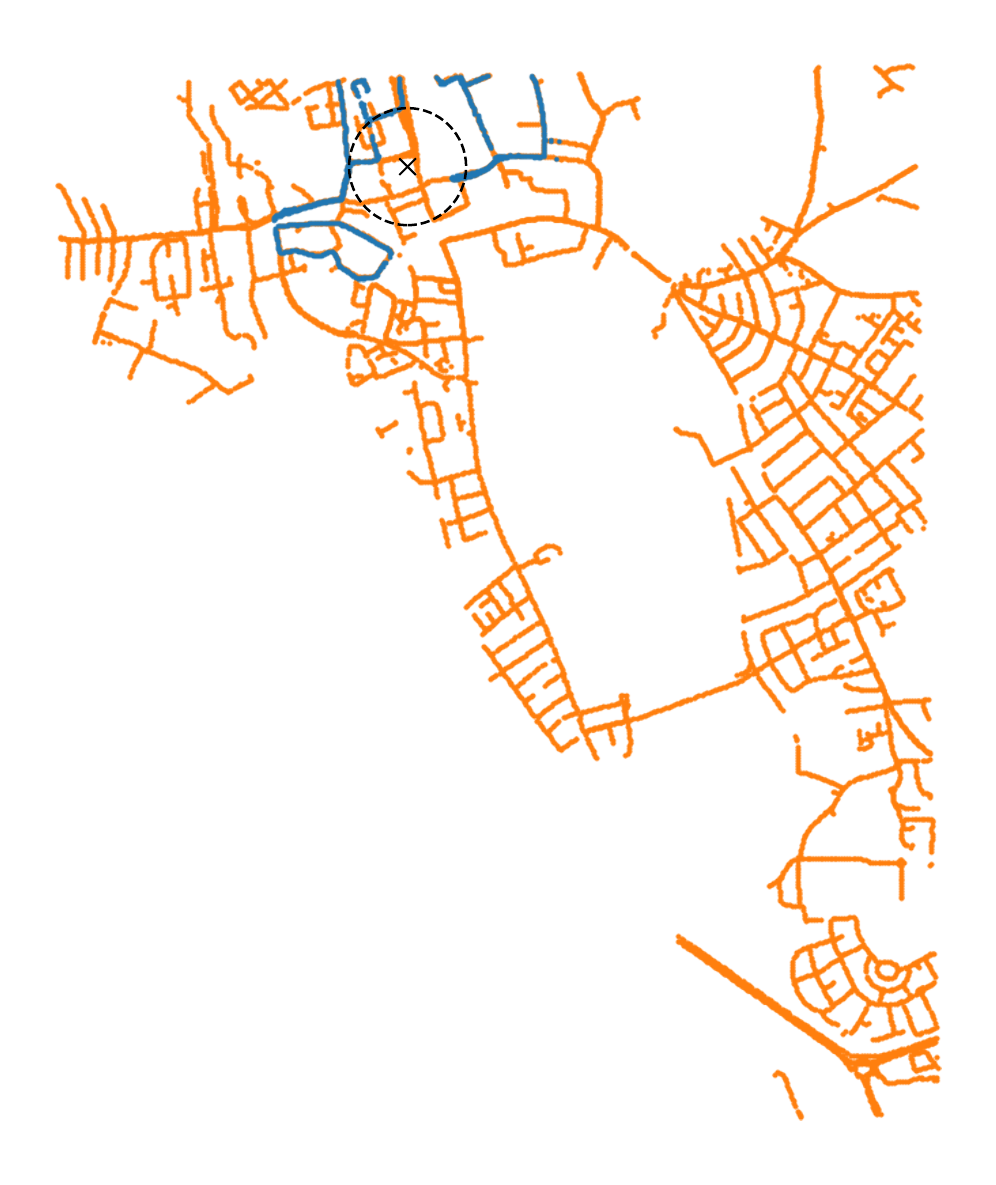} \\
\end{tabular}

\vspace{0pt}
\setlength{\tabcolsep}{2pt}
\begin{tabular}{ccc}
\small AmsterTime & \small Gardens Point & \small Nordland \\
\includegraphics[width=0.325\textwidth]{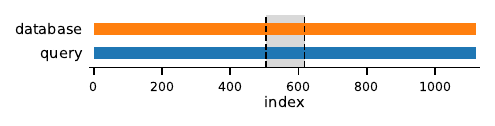} &
\includegraphics[width=0.325\textwidth]{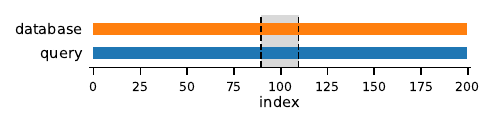} &
\includegraphics[width=0.325\textwidth]{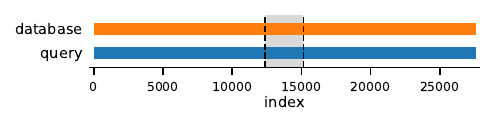} \\
\end{tabular}

\vspace{0pt}
\begingroup
\definecolor{csquery}{HTML}{1F77B4}
\definecolor{csdb}{HTML}{FF7F0E}
\small
\mbox{\textcolor{csquery}{\rule[0.5ex]{6pt}{2pt}}~Query}\quad
\mbox{\textcolor{csdb}{\rule[0.5ex]{6pt}{2pt}}~Database}\quad
\mbox{\textbf{-\,-\,-}~Fitting region boundary ($\times$: center)}\quad
\mbox{\fcolorbox{black}{black!15}{\rule{0pt}{6pt}\rule{6pt}{0pt}}~Fitting interval}
\endgroup

\caption{Center fitting region used for concept erasure, holding 10\% of
the queries.}
\label{fig:supp_center_split}
\end{minipage}
 \par\nointerlineskip\vspace{8pt}
\begin{minipage}[t]{\linewidth}
\captionsetup{type=figure}
\centering
\begingroup
\definecolor{mAnyLoc}{HTML}{1f77b4}
\definecolor{mBoQ}{HTML}{ff7f0e}
\definecolor{mCosPlace}{HTML}{2ca02c}
\definecolor{mCricaVPR}{HTML}{d62728}
\definecolor{mEigenPlaces}{HTML}{9467bd}
\definecolor{mMegaLoc}{HTML}{8c564b}
\definecolor{mMixVPR}{HTML}{e377c2}
\definecolor{mSALAD}{HTML}{7f7f7f}
\definecolor{mSALADCM}{HTML}{bcbd22}
\definecolor{mSelaVPR}{HTML}{17becf}
\definecolor{mSelaVPRpp}{HTML}{000000}
\begin{tikzpicture}
\begin{groupplot}[group style={group size=4 by 2, horizontal sep=6mm, vertical sep=4mm},
  width=0.28\textwidth, height=0.23\textwidth, ymin=0, ymax=100, ytick={0,25,50,75,100},
  tick label style={font=\scriptsize}, label style={font=\scriptsize},
  title style={font=\small, yshift=-1mm},
  xticklabel style={font=\tiny, align=center, anchor=north, inner ysep=2pt},
  enlarge x limits=0.08,
  grid=both, major grid style={line width=.1pt,draw=gray!25}, axis on top]
\nextgroupplot[xtick={0,1,2}, title={Gardens Point}, xticklabels={}, ylabel={DAR@1 (\%)}]
\addplot[mAnyLoc, mark=*, mark size=1.6pt] coordinates {(0,15.00) (1,18.00) (2,93.50)};
\addplot[mBoQ, mark=square*, mark size=1.6pt] coordinates {(0,37.50) (1,51.00) (2,99.00)};
\addplot[mCosPlace, mark=triangle*, mark size=2pt] coordinates {(0,18.50) (1,19.00) (2,77.00)};
\addplot[mCricaVPR, mark=diamond*, mark size=2pt] coordinates {(0,29.50) (1,43.00) (2,95.50)};
\addplot[mEigenPlaces, mark=pentagon*, mark size=2pt] coordinates {(0,16.00) (1,23.00) (2,80.00)};
\addplot[mMegaLoc, mark=*, mark size=1.6pt] coordinates {(0,12.50) (1,13.50) (2,93.00)};
\addplot[mMixVPR, mark=square*, mark size=1.6pt] coordinates {(0,9.00) (1,12.00) (2,85.50)};
\addplot[mSALAD, mark=triangle*, mark size=2pt] coordinates {(0,34.00) (1,39.00) (2,98.00)};
\addplot[mSALADCM, mark=diamond*, mark size=2pt] coordinates {(0,26.00) (1,32.50) (2,93.00)};
\addplot[mSelaVPR, mark=pentagon*, mark size=2pt] coordinates {(0,32.00) (1,43.50) (2,95.50)};
\addplot[mSelaVPRpp, mark=*, mark size=1.6pt] coordinates {(0,30.50) (1,40.00) (2,97.00)};
\nextgroupplot[xtick={0,1,2,3}, title={Nordland}, xticklabels={}]
\addplot[mAnyLoc, mark=*, mark size=1.6pt] coordinates {(0,2.23) (1,3.19) (2,7.81) (3,22.74)};
\addplot[mBoQ, mark=square*, mark size=1.6pt] coordinates {(0,44.41) (1,50.17) (2,59.73) (3,88.49)};
\addplot[mCosPlace, mark=triangle*, mark size=2pt] coordinates {(0,15.07) (1,15.95) (2,20.21) (3,47.80)};
\addplot[mCricaVPR, mark=diamond*, mark size=2pt] coordinates {(0,19.90) (1,24.65) (2,35.37) (3,83.41)};
\addplot[mEigenPlaces, mark=pentagon*, mark size=2pt] coordinates {(0,15.28) (1,16.16) (2,20.86) (3,53.06)};
\addplot[mMegaLoc, mark=*, mark size=1.6pt] coordinates {(0,73.06) (1,76.72) (2,84.85) (3,92.90)};
\addplot[mMixVPR, mark=square*, mark size=1.6pt] coordinates {(0,9.75) (1,11.79) (2,15.45) (3,59.93)};
\addplot[mSALAD, mark=triangle*, mark size=2pt] coordinates {(0,32.14) (1,37.62) (2,47.66) (3,84.14)};
\addplot[mSALADCM, mark=diamond*, mark size=2pt] coordinates {(0,87.74) (1,88.74) (2,91.95) (3,95.97)};
\addplot[mSelaVPR, mark=pentagon*, mark size=2pt] coordinates {(0,18.01) (1,21.31) (2,31.00) (3,77.54)};
\addplot[mSelaVPRpp, mark=*, mark size=1.6pt] coordinates {(0,70.36) (1,74.18) (2,83.47) (3,96.00)};
\nextgroupplot[xtick={0,1,2,3,4}, title={St.~Lucia}, xticklabels={}]
\addplot[mAnyLoc, mark=*, mark size=1.6pt] coordinates {(0,44.16) (1,57.32) (2,65.66) (3,76.41) (4,79.93)};
\addplot[mBoQ, mark=square*, mark size=1.6pt] coordinates {(0,99.63) (1,99.75) (2,99.75) (3,99.88) (4,99.94)};
\addplot[mCosPlace, mark=triangle*, mark size=2pt] coordinates {(0,98.89) (1,99.26) (2,99.20) (3,99.26) (4,99.63)};
\addplot[mCricaVPR, mark=diamond*, mark size=2pt] coordinates {(0,97.65) (1,99.01) (2,99.14) (3,99.63) (4,100.00)};
\addplot[mEigenPlaces, mark=pentagon*, mark size=2pt] coordinates {(0,98.39) (1,98.39) (2,98.89) (3,99.07) (4,99.26)};
\addplot[mMegaLoc, mark=*, mark size=1.6pt] coordinates {(0,99.63) (1,99.88) (2,99.81) (3,99.94) (4,99.94)};
\addplot[mMixVPR, mark=square*, mark size=1.6pt] coordinates {(0,97.65) (1,98.21) (2,98.64) (3,99.20) (4,99.51)};
\addplot[mSALAD, mark=triangle*, mark size=2pt] coordinates {(0,99.14) (1,99.44) (2,99.57) (3,99.57) (4,99.88)};
\addplot[mSALADCM, mark=diamond*, mark size=2pt] coordinates {(0,99.88) (1,99.88) (2,100.00) (3,100.00) (4,100.00)};
\addplot[mSelaVPR, mark=pentagon*, mark size=2pt] coordinates {(0,98.15) (1,98.33) (2,98.64) (3,99.07) (4,99.51)};
\addplot[mSelaVPRpp, mark=*, mark size=1.6pt] coordinates {(0,99.88) (1,100.00) (2,99.94) (3,99.94) (4,100.00)};
\nextgroupplot[xtick={0,1,2,3,4,5}, title={SVOX}, xticklabels={}]
\addplot[mAnyLoc, mark=*, mark size=1.6pt] coordinates {(0,1.38) (1,4.36) (2,7.91) (3,9.86) (4,55.62) (5,89.33)};
\addplot[mBoQ, mark=square*, mark size=1.6pt] coordinates {(0,86.93) (1,93.58) (2,93.92) (3,97.02) (4,98.39) (5,98.39)};
\addplot[mCosPlace, mark=triangle*, mark size=2pt] coordinates {(0,76.61) (1,80.05) (2,81.31) (3,84.63) (4,89.45) (5,89.68)};
\addplot[mCricaVPR, mark=diamond*, mark size=2pt] coordinates {(0,56.19) (1,70.87) (2,75.69) (3,77.98) (4,90.14) (5,97.02)};
\addplot[mEigenPlaces, mark=pentagon*, mark size=2pt] coordinates {(0,77.06) (1,80.28) (2,82.80) (3,86.70) (4,91.28) (5,91.63)};
\addplot[mMegaLoc, mark=*, mark size=1.6pt] coordinates {(0,90.14) (1,94.61) (2,93.81) (3,96.44) (4,97.48) (5,97.71)};
\addplot[mMixVPR, mark=square*, mark size=1.6pt] coordinates {(0,71.22) (1,81.31) (2,86.93) (3,91.51) (4,95.99) (5,96.10)};
\addplot[mSALAD, mark=triangle*, mark size=2pt] coordinates {(0,83.14) (1,90.02) (2,92.43) (3,92.89) (4,97.48) (5,98.28)};
\addplot[mSALADCM, mark=diamond*, mark size=2pt] coordinates {(0,91.51) (1,94.27) (2,94.50) (3,97.02) (4,97.48) (5,98.51)};
\addplot[mSelaVPR, mark=pentagon*, mark size=2pt] coordinates {(0,71.90) (1,81.54) (2,86.24) (3,89.22) (4,93.35) (5,95.76)};
\addplot[mSelaVPRpp, mark=*, mark size=1.6pt] coordinates {(0,92.32) (1,96.33) (2,96.44) (3,98.28) (4,98.39) (5,98.51)};
\nextgroupplot[xtick={0,1,2}, xticklabels={{\strut day\_right\\\strut (query)},{\strut day\_left},{\strut none}}, ylabel={\acs{DARx}@1 (\%)}]
\addplot[mAnyLoc, mark=*, mark size=1.6pt] coordinates {(0,81.00) (1,82.00) (2,93.50)};
\addplot[mBoQ, mark=square*, mark size=1.6pt] coordinates {(0,99.00) (1,99.00) (2,99.00)};
\addplot[mCosPlace, mark=triangle*, mark size=2pt] coordinates {(0,63.50) (1,67.50) (2,77.00)};
\addplot[mCricaVPR, mark=diamond*, mark size=2pt] coordinates {(0,95.00) (1,95.50) (2,95.50)};
\addplot[mEigenPlaces, mark=pentagon*, mark size=2pt] coordinates {(0,71.50) (1,72.50) (2,80.00)};
\addplot[mMegaLoc, mark=*, mark size=1.6pt] coordinates {(0,90.50) (1,91.50) (2,93.00)};
\addplot[mMixVPR, mark=square*, mark size=1.6pt] coordinates {(0,67.00) (1,73.00) (2,85.50)};
\addplot[mSALAD, mark=triangle*, mark size=2pt] coordinates {(0,97.00) (1,98.00) (2,98.00)};
\addplot[mSALADCM, mark=diamond*, mark size=2pt] coordinates {(0,93.00) (1,93.00) (2,93.00)};
\addplot[mSelaVPR, mark=pentagon*, mark size=2pt] coordinates {(0,94.00) (1,95.50) (2,95.50)};
\addplot[mSelaVPRpp, mark=*, mark size=1.6pt] coordinates {(0,97.00) (1,97.00) (2,97.00)};
\nextgroupplot[xtick={0,1,2,3}, xticklabels={{\strut summer\\\strut (query)},{\strut fall},{\strut spring},{\strut none}}]
\addplot[mAnyLoc, mark=*, mark size=1.6pt] coordinates {(0,3.50) (1,3.94) (2,8.68) (3,22.74)};
\addplot[mBoQ, mark=square*, mark size=1.6pt] coordinates {(0,54.43) (1,58.05) (2,65.84) (3,88.49)};
\addplot[mCosPlace, mark=triangle*, mark size=2pt] coordinates {(0,18.68) (1,18.99) (2,23.64) (3,47.80)};
\addplot[mCricaVPR, mark=diamond*, mark size=2pt] coordinates {(0,28.12) (1,32.01) (2,42.72) (3,83.41)};
\addplot[mEigenPlaces, mark=pentagon*, mark size=2pt] coordinates {(0,18.74) (1,19.03) (2,24.11) (3,53.06)};
\addplot[mMegaLoc, mark=*, mark size=1.6pt] coordinates {(0,79.54) (1,81.77) (2,87.30) (3,92.90)};
\addplot[mMixVPR, mark=square*, mark size=1.6pt] coordinates {(0,12.38) (1,13.91) (2,18.05) (3,59.93)};
\addplot[mSALAD, mark=triangle*, mark size=2pt] coordinates {(0,43.25) (1,46.54) (2,55.44) (3,84.14)};
\addplot[mSALADCM, mark=diamond*, mark size=2pt] coordinates {(0,91.12) (1,91.40) (2,93.38) (3,95.97)};
\addplot[mSelaVPR, mark=pentagon*, mark size=2pt] coordinates {(0,26.92) (1,28.72) (2,38.19) (3,77.54)};
\addplot[mSelaVPRpp, mark=*, mark size=1.6pt] coordinates {(0,80.38) (1,81.80) (2,87.87) (3,96.00)};
\nextgroupplot[xtick={0,1,2,3,4}, xticklabels={{\strut 08:45\\\strut (query)},{\strut 08:45\\\strut (19 Aug)},{\strut 12:10\\\strut (10 Sep)},{\strut 15:45\\\strut (11 Sep)},{\strut none}}]
\addplot[mAnyLoc, mark=*, mark size=1.6pt] coordinates {(0,60.65) (1,69.67) (2,72.45) (3,76.78) (4,79.93)};
\addplot[mBoQ, mark=square*, mark size=1.6pt] coordinates {(0,99.94) (1,99.94) (2,99.94) (3,99.94) (4,99.94)};
\addplot[mCosPlace, mark=triangle*, mark size=2pt] coordinates {(0,99.57) (1,99.51) (2,99.57) (3,99.51) (4,99.63)};
\addplot[mCricaVPR, mark=diamond*, mark size=2pt] coordinates {(0,99.69) (1,99.75) (2,99.81) (3,99.94) (4,100.00)};
\addplot[mEigenPlaces, mark=pentagon*, mark size=2pt] coordinates {(0,99.01) (1,99.07) (2,99.20) (3,99.14) (4,99.26)};
\addplot[mMegaLoc, mark=*, mark size=1.6pt] coordinates {(0,99.94) (1,99.94) (2,99.94) (3,99.94) (4,99.94)};
\addplot[mMixVPR, mark=square*, mark size=1.6pt] coordinates {(0,99.20) (1,99.26) (2,99.26) (3,99.32) (4,99.51)};
\addplot[mSALAD, mark=triangle*, mark size=2pt] coordinates {(0,99.88) (1,99.88) (2,99.88) (3,99.75) (4,99.88)};
\addplot[mSALADCM, mark=diamond*, mark size=2pt] coordinates {(0,99.94) (1,100.00) (2,100.00) (3,100.00) (4,100.00)};
\addplot[mSelaVPR, mark=pentagon*, mark size=2pt] coordinates {(0,99.44) (1,99.44) (2,99.26) (3,99.38) (4,99.51)};
\addplot[mSelaVPRpp, mark=*, mark size=1.6pt] coordinates {(0,100.00) (1,100.00) (2,100.00) (3,100.00) (4,100.00)};
\nextgroupplot[xtick={0,1,2,3,4,5}, xticklabels={{\strut overcast\\\strut (query)},{\strut snow},{\strut rain},{\strut sun},{\strut night},{\strut none}}, legend to name=leg:supp_distractor_sets, legend columns=6, legend style={font=\scriptsize, draw=none, /tikz/every even column/.append style={column sep=6pt}}]
\addplot[mAnyLoc, mark=*, mark size=1.6pt] coordinates {(0,5.62) (1,15.71) (2,26.61) (3,28.10) (4,75.92) (5,89.33)};
\addlegendentry{AnyLoc}
\addplot[mBoQ, mark=square*, mark size=1.6pt] coordinates {(0,97.13) (1,97.82) (2,97.48) (3,98.17) (4,98.39) (5,98.39)};
\addlegendentry{BoQ}
\addplot[mCosPlace, mark=triangle*, mark size=2pt] coordinates {(0,87.39) (1,87.96) (2,88.07) (3,89.33) (4,89.56) (5,89.68)};
\addlegendentry{CosPlace}
\addplot[mCricaVPR, mark=diamond*, mark size=2pt] coordinates {(0,90.25) (1,91.28) (2,92.32) (3,93.00) (4,94.61) (5,97.02)};
\addlegendentry{CricaVPR}
\addplot[mEigenPlaces, mark=pentagon*, mark size=2pt] coordinates {(0,88.30) (1,88.30) (2,88.76) (3,91.17) (4,91.51) (5,91.63)};
\addlegendentry{EigenPlaces}
\addplot[mMegaLoc, mark=*, mark size=1.6pt] coordinates {(0,97.59) (1,97.71) (2,97.71) (3,97.71) (4,97.71) (5,97.71)};
\addlegendentry{MegaLoc}
\addplot[mMixVPR, mark=square*, mark size=1.6pt] coordinates {(0,91.86) (1,93.12) (2,93.81) (3,95.41) (4,96.10) (5,96.10)};
\addlegendentry{MixVPR}
\addplot[mSALAD, mark=triangle*, mark size=2pt] coordinates {(0,97.71) (1,98.05) (2,98.17) (3,98.28) (4,98.28) (5,98.28)};
\addlegendentry{SALAD}
\addplot[mSALADCM, mark=diamond*, mark size=2pt] coordinates {(0,98.05) (1,98.28) (2,98.51) (3,98.51) (4,98.39) (5,98.51)};
\addlegendentry{SALAD-CM}
\addplot[mSelaVPR, mark=pentagon*, mark size=2pt] coordinates {(0,91.63) (1,92.89) (2,93.58) (3,94.50) (4,95.64) (5,95.76)};
\addlegendentry{SelaVPR}
\addplot[mSelaVPRpp, mark=*, mark size=1.6pt] coordinates {(0,98.51) (1,98.51) (2,98.51) (3,98.51) (4,98.51) (5,98.51)};
\addlegendentry{SelaVPR++}
\end{groupplot}
\end{tikzpicture}\\[1mm]
\pgfplotslegendfromname{leg:supp_distractor_sets}
\endgroup
\caption{\ac{DAR}@1 (top) and \acs{DARx}@1 with distractors drawn from different sets.
On each dataset the query set (left, the main-results setting) comes first, the other
sets are ordered by mean \ac{DAR}@1 across methods, and \textit{none} (right) is the
no-distractor reference, equal to \ac{R1}.}
\label{fig:supp_distractor_sets}
\end{minipage}
 \par\nointerlineskip\vspace{8pt}
\begin{minipage}[t]{\linewidth}
\captionsetup{type=table}
\centering
\begingroup
\setlength{\tabcolsep}{3pt}
\renewcommand{\arraystretch}{1.05}
\caption{VPR method configurations. \textbf{Dim}=global
descriptor dimensionality; \textbf{Re-rank}=number of
top-ranked candidates re-ranked. $^{\ddagger}$CricaVPR
uses a cross-image encoder, so descriptors depend on the
inference batch; following the original paper, we use a
batch size of 16.}
\label{tab:methods_config}
{\footnotesize%
\begin{tabular}{l l l r l l}
\toprule
Method                                      & Backbone          & Aggregation                   & Dim          & Train data          & Re-rank          \\
\midrule
AnyLoc~\cite{keetha2023anyloc}              & DINOv2 ViT-G/14   & VLAD (Urban, 32 clusters)     & 49152        & --                  & --               \\
BoQ~\cite{ali2024boq}                       & DINOv2 ViT-B/14   & BoQ                           & 12288        & GSV-Cities          & --               \\
CosPlace~\cite{berton2022rethinking}        & ResNet-101        & GeM + FC                      & 1024         & SF-XL               & --               \\
CricaVPR$^{\ddagger}$~\cite{lu2024cricavpr} & DINOv2 ViT-B/14   & multi-scale GeM + cross-image & 10752        & GSV-Cities          & --               \\
EigenPlaces~\cite{berton2023eigenplaces}    & ResNet-50         & GeM + FC                      & 2048         & SF-XL               & --               \\
MegaLoc~\cite{berton2025megaloc}            & DINOv2 ViT-B/14   & SALAD + projection            & 8448         & 5-set union         & --               \\
MixVPR~\cite{ali2023mixvpr}                 & ResNet-50         & Feature-Mixer                 & 4096         & GSV-Cities          & --               \\
SALAD~\cite{izquierdo2024optimal}           & DINOv2 ViT-B/14   & SALAD                         & 8448         & GSV-Cities          & --               \\
SALAD-CM~\cite{izquierdo2024close}          & DINOv2 ViT-B/14   & SALAD                         & 8448         & GSV-Cities + MSLS   & --               \\
SelaVPR~\cite{lu2024towards}                & DINOv2 ViT-L/14   & GeM                           & 1024         & MSLS + Pitts30k     & top 100          \\
SelaVPR++~\cite{lu2025selavpr++}            & DINOv2 ViT-L/14   & GeM                           & 4096         & GSV+MSLS+SF-XL+P30k & top 100          \\
\bottomrule
\end{tabular}%
}
\endgroup
\end{minipage}
 \par\end{minipage}
 
\end{document}